\documentclass[11pt]{article}
\usepackage{fullpage}

\newcommand{\blue}{\color{blue}}

\usepackage[english]{babel}
\usepackage{xspace}
\usepackage{algorithm,algorithmicx}
\usepackage{tcolorbox}
\usepackage{tikz}

\usepackage[colorlinks=true,
linkcolor=blue,
urlcolor=blue,
citecolor=blue]{hyperref}

\newcommand{\aname}{{\textsf{CAC}}}
\newcommand{\bname}{{\textsf{CAC-NPS}}}

\usepackage[utf8]{inputenc} % allow utf-8 input
\usepackage[T1]{fontenc}    % use 8-bit T1 fonts
\usepackage{hyperref}       % hyperlinks
\usepackage{url}            % simple URL typesetting
\usepackage{booktabs}       % professional-quality tables
\usepackage{amsfonts}  
\usepackage{amsmath}
\usepackage{nicefrac}       % compact symbols for 1/2, etc.
\usepackage{microtype}      % microtypography
\usepackage{xcolor, colortbl}        % colors
\usepackage{wrapfig}
\usepackage{caption}
\usepackage{subcaption}
\usepackage{footnote}
\makesavenoteenv{tabular}
\makesavenoteenv{table}
\usepackage{multirow}
\usepackage{bbm}
\usepackage{amsthm}
\usepackage[makeroom]{cancel}
\usepackage{stmaryrd}
\usepackage{amsmath,bm}
\usepackage{amssymb}
\usepackage{mathtools, cuted}
\usepackage{kantlipsum,setspace}
\usepackage{enumitem}
\usepackage{algorithm}
\usepackage{algpseudocode}
 
\definecolor{Gray}{gray}{0.85}

\newcommand{\EE}{\mathbb{E}}

\newcommand{\tb}{\textbf}

\newcommand{\bomega}{\boldsymbol{\omega}}
\newcommand{\blambda}{\boldsymbol{\lambda}}
\newcommand{\btheta}{\boldsymbol{\theta}}
\newcommand{\bone}{\mathbf{1}}
\newcommand{\ba}{\boldsymbol{a}}

\newtheorem{proposition}{Proposition}
\newtheorem{lemma}{Lemma}

\newtheorem{theorem}{Theorem}

\newtheorem{assumption}{Assumption}
\newtheorem*{setting}{Setting}
\newtheorem{Corollary}{Corollary}

\title{Learning to Coordinate in Multi-Agent Systems: \\
A Coordinated Actor-Critic Algorithm and Finite-Time Guarantees}

\author{\large Siliang Zeng$^\dagger$, Tianyi Chen$^\ast$, Alfredo Garcia$^{\ddagger}$, Mingyi Hong$^\dagger$\\[.5cm]
	\small $^{\dagger}$Department  of Electrical and Computer Engineering, \\
	\small University of Minnesota, MN, USA\\
	\small $^{\ast}$Department of Electrical, Computer, and Systems Engineering,\\
	\small Rensselaer Polytechnic Institute, NY, USA\\
	\small $^{\ddagger}$ Department of Industrial and Systems Engineering,\\
	\small Texas A\&M University, TX, USA\\
	\small Email: \texttt{\{zeng0176, mhong\}@umn.edu}, \texttt{chent18@rpi.edu},\\
	\small \texttt{alfredo.garcia@tamu.edu}} 

\begin{document}

\maketitle

\maketitle

\begin{abstract}%
 Multi-agent reinforcement learning (MARL) has attracted much research attention recently. However, unlike its single-agent counterpart, many theoretical and algorithmic aspects of MARL have not been well-understood. 	In this paper, we study the emergence of coordinated behavior by autonomous agents using an actor-critic (AC) algorithm. Specifically, we propose and analyze a class of coordinated actor-critic  (\aname) algorithms in which individually parametrized policies have a {\it  shared} part (which is jointly optimized among all agents) and a {\it personalized} part (which is only locally optimized). Such a kind of {\it partially personalized} policy allows agents to coordinate by leveraging peers' experience and adapt to individual tasks. The flexibility in our design allows the proposed \aname ~algorithm to be used in a {\it fully decentralized} setting, where the agents can only communicate with their neighbors, as well as in a {\it federated} setting, where the agents occasionally communicate with a server while optimizing their (partially personalized) local models. %It is worth mentioning that in the latter case, our algorithm reduce
		Theoretically, we show that under some standard regularity assumptions, the proposed \aname ~algorithm requires $\mathcal{O}(\epsilon^{-\frac{5}{2}})$ samples to achieve an $\epsilon$-stationary solution (defined as the solution whose squared norm  of the gradient of the objective function  is less than $\epsilon$). %Further, as a special case, we show that when the policy parameters are {\it completely personalized}, the sample complexity improves to $\mathcal{O}(\epsilon^{-\frac{5}{2}})$. 
		To the best of our knowledge, this work provides the first finite-sample guarantee for decentralized AC algorithm with partially personalized policies.
\end{abstract}

\section{Introduction} \label{sec: intro}
	%Reinforcement learning (RL) \cite{sutton2018reinforcement} evolves from the Markov Decision Process (MDP), where the agent
    %solves a sequential decision-making problem through interacting with the environment to search for a optimal policy. 
    We consider the multi-agent reinforcement learning (MARL) problem, in which a common environment is influenced by the joint actions of multiple autonomous agents, each aiming to optimize their own individual objective. 
    % This problem belongs to the class of  cooperative MARL and has important applications including production planning and manufacturing scheduling (\cite{gabel2007,dittrich2020}) and vehicle routing (\cite{silva2020,Zhang2020,zeng2020multi}). 
    The MARL \cite{zhang2019multi,lee2020optimization} has received significant attention recently due to their outstanding performance in
many practical applications including robotics \cite{stone2000multiagent}, autonomous driving \cite{shalev2016safe} and video games \cite{tampuu2017multiagent}. Many efficient algorithms have been proposed  \cite{lowe2017multi,espeholt2018impala,rashid2018qmix}, but unlike its single-agent counterpart, the theoretical understanding of MARL is still very limited, especially in settings 
where there is no central controller to coordinate different agents, so that the information sharing is limited \cite{zhang2019multi}.   %strongly motivates a rigorous theoretical understanding  %Despite realizing the potentials of MARL, there are many open questions waiting to be answered in order to obtain a deeper understanding on MARL. 

%       Recently, %In recent years, by leveraging the advances of deep learning, 
%    
%    %Despite the fact that MARL has shown promising empirical performance in a number of applications, most of them still treat MARL from the perspective of centralized control. 
%    However, the theoretical understanding about MARL is still limited,  especially in settings 
%    %the fact that MARL has shown promising empirical performance in a number of applications, In many practical settings such as autonomous driving \cite{shalev2016safe} and cloud-edge AI systems \cite{stoica2017berkeley}, 
%    where there is not a central controller to coordinate different agents and the information sharing is limited \cite{zhang2019multi}. This is partly due to the difficulties in characterizing how different actions of the agents can affect each other, as well as the challenges in understanding the invariance and the heterogeneity among different agents. 
%    
    
   An important subclass of MARL -- the so-called {\it cooperative} MARL -- has become popular recently due to its wide applications. 
%   as well as its potential of obtaining some interesting theoretical characterization, thanks to its strong connection with the classical decentralized optimization. 
   In the cooperative MARL, the agents aim to collaborate with each other to learn and optimize a joint global objective. To this end, local information exchange and local communication may be used to jointly optimize a system-level performance measure \cite{zhang2018fully,grosnit2021decentralized,lu2021decentralized}. Next, we provide a brief survey about related works in cooperative MARL, and discuss their settings as well as theoretical guarantees. %A class of MARL coordinated actor-critic (CAC) algorithms will be proposed and analyzed, which allows the agents to jointly estimate the system value functions, as well as to coordinately search for their individual actions.   %to design a cla, and provide theoretical analysis for them. 

    {\bf Related Works.} 
    The systematic study of the cooperative MARL can be traced back to \cite{claus1998dynamics,wolpert1999general}, which extended Q-learning algorithm \cite{watkins1992q} or its variants to the multi-agent setting.
    More recently,  there are a number of works that characterize the theoretical performance of cooperative MARL algorithms in a fully observable, decentralized setting (\cite{kar2012qd,zhang2018fully,doan2019finite}). In such a setting, the agents are connected by a time-varying graph, and they can only communicate with their immediate neighbors.
    The goal of the agents is to cooperatively maximize certain global reward, by communicating local information with their neighbors. %To clarify the common objective for all agents in the networked system, at each iteration step the global reward of the whole system is defined as the {\em average} of the local reward, which is locally received by each agent. %{\red[this is not the only cooperative RL setting, but a special case.]}
    %Thus in the above cooperative MARL setting, partially sharing policies agents with other agents with similar local rewards may improve overall performance {\red[not cler why this is true?]}. 
    %{\red[it is a bit not clear what you want to say in the above setting. I think the main theme should be to introduce recent MRAL algorithms with theoretical performance, and a typical/general settings that most works are based upon. Is the above setting a common one? ]}
	%{\red[for the above setting, typically what are the dencentralized algorithm do?]}
	Under the above cooperative MARL setting, there are several lines of works which studied different problem formulations, proposed new algorithms and analyzed their theoretical performance. 
	
	The first line of works about the coorperative and fully observable MARL has focused on developing and analyzing policy evaluation algorithms, where the agents  jointly estimate the global value function for a given policy. In \cite{wai2018multi}, a decentralized double averaging
    primal-dual optimization algorithm was proposed to solve the mean squared projected Bellman error minimization problem. It is shown that the proposed algorithm converges to the optimal solution at a global geometric rate. In \cite{doan2019finite}, the authors obtained a finite-sample analysis for decentralized TD(0) method. Their analysis is closely related to the theoretical results of decentralized stochastic gradient descent method on convex optimization problems \cite{nedic2010constrained}.
    
%    It is worth noting that, most of the above mentioned works have utilized the connection between MARL policy evaluation problem and the classical decentralized (convex) optimization algorithms. 
However, the problem becomes much more challenging when the agents are allowed to optimize their policies. %and in this case, it is no longer straightforward to leverage the techniques from the  decentralized optimization literature.  
A recent line of works  has focused on applying and analyzing various policy optimization methods in the MARL setting. 
    	In \cite{zhang2018fully}, the authors extended the actor-critic (AC) algorithm \cite{konda2000actor} to the cooperative MARL setting. The algorithm allows each agent to perform its local policy improvement step while approximating the global value function.
    	A few more recent works have extended \cite{zhang2018fully} in different directions. For example in \cite{grosnit2021decentralized}, the authors considered the continuous action spaces and obtained the asymptotic convergence guarantee under both off-policy and on-policy settings.
    	Moreover, \cite{zhang2021marl} considered a new decentralized formulation where all agents cooperate to maximize general utilities in the cooperative MARL system, it developed AC-type algorithms to fit this setting but still suffering from high sampling cost in estimating the occupancy measure for all states and the nested loop of optimization steps. A concurrent work \cite{chen2021sample} adopts large-batch updates in decentralized (natural) AC methods to improve sample and communication efficiency, whose convergence rate matches the analysis results of the corresponding centralized versions \cite{xu2020improving}. However, the proposed algorithms in \cite{chen2021sample} needs to generate $\mathcal{O}(\epsilon^{-1} \ln{ \epsilon^{-1}})$ samples to update critic parameter before performing each actor update. %which is infeasible for practical reinforcement learning scenario.}
       It is worth noting, that all the above mentioned works do not allow the agents to share their local policies.

       {\bf Our Contributions.} Although there have been a growing literature on analyzing theoretical aspects of cooperative MARL, many challenges still remain, even under the basic fully observed setting. For example, most of the cooperative policy optimization algorithms, assume relatively simple collaboration mechanism, where the agents collaborate by jointly estimating the global value function, while {\it independently} optimizing their local policies.  Such a form of collaboration decouples the agents' policy optimization process, and it is relatively easy to analyze. However, it fails to capture some intrinsic aspects of cooperative MARL, in the sense that when the agents' local tasks are similar (a.k.a. the {\it homogeneous} setting), the agent's policy should also be closely related to each other. Such an intuition has been verified in  MARL systems  \cite{gupta2017cooperative,terry2020revisiting}, multi-task RL systems \cite{omidshafiei2017deep,zeng2020decentralized, yu2020meta}, Markov games \cite{vadori2020calibration} and mean-field multi-agent reinforcement learning \cite{liu2020decentralized, li2021permutation}, where parameter sharing scheme results in more stable convergence due to the benefit of learning homogeneity among different agents. %However, it is not clear how to design and analyze more sophisticated collaboration schemes which enable the agents to (partially) share their local policies to help them leverage each other's experience and build better behavior strategies. 

       In this work, we aim at providing  better theoretical and practical understandings about the cooperative MARL problem. In particular, we consider the setting where the agents are connected by a time-varying network, and they can access the common observations while having different reward functions.  We propose and analyze a Coordinated Actor-Critic  (\aname) algorithm, which allows each agent to (partially) share its policy parameters with the neighbors for learning the homogeneity / common knowledge in the multi-agent system. To our knowledge, we provide the first non-asymptotic convergence result for two-timescale multi-agent AC methods. Moreover, we conduct extensive numerical experiments to demonstrate the effectiveness of the proposed algorithm.

	\section{Preliminaries} \label{sec: background}
	In this section, we introduce the background and formulation of the cooperative, fully observable MARL in a decentralized system.
	%Suppose there are  multiple agents  aiming to independently learn and optimize a common global objective, and each agent can communicate with its neighbors in a network with time-varying topology. The  common environment is observable by all the agents, and it is influenced by their joint actions. 
	To model the communication pattern among the agents, let us define the time-varying graph $ \mathcal{G}_t = ( \mathcal{N}, \mathcal{E}_t ) $ consisting of a set of $ \mathcal{N} $ nodes and a set of $ \mathcal{E}_t $ edges, with $|\mathcal{N}|=N$ and $|\mathcal{E}| = E$. Each node $i \in \mathcal{N}$ represents an agent and $\mathcal{E}_t$ represents the set of communication links at time $t$ so that the agents are connected according to the links $ \mathcal{E}_t $.
	
	Consider the MARL problem, formulated as a discrete-time Markov Decision Process (MDP)  $M : = \langle \mathcal{S}, \mathcal{A}, \mathcal{P}, \eta, \mathcal{R}, \gamma \rangle $, where $ \mathcal{S} $ is the finite space for global state $s$ and $ \mathcal{A} $ is the finite space for joint action $ \ba = \{a_{i}\}_{i=1}^N $; 
	$ \eta(s): \mathcal{S} \to [0,1] $ denotes the initial state distribution;  $ \mathcal{P}(s^\prime \mid s, \ba): \mathcal{S} \times \mathcal{A} \times \mathcal{S} \to [0, 1] $  denotes the transition probability; $ r_i(s, \ba): \mathcal{S} \times \mathcal{A} \to \mathcal{R} $ denotes the local reward function of agent $i$; $ \gamma \in (0,1) $ is the discounted factor. Furthermore, suppose the policy of each agent $i$ is parameterized by $ \theta_i $, then $\btheta := \{ \theta_i \}_{i=1}^N$ denotes the collections of all policy parameters in the multi-agent system.
	Then $ \mu_{\btheta}(s) $ denotes the stationary distribution of each state $ s $ under joint policy $ \pi_{\btheta} $, and $ d_{\btheta}(\cdot) $ denotes the discounted visitation measure where $ d_{\btheta}(s) := (1-\gamma)\sum_{t=0}^{\infty} \gamma^t \cdot \mathcal{P}^{\pi_{\btheta}} (s_t = s\mid s_0 \sim \eta)$. Under the joint policy $ \pi_{\btheta} $, the probability for choosing any joint action $ \ba := \{ a_{i} \}_{i=1}^N $ could be expressed as $\pi_{\btheta}( \ba | s) := \Pi_{i=1}^{N} \pi_{i}(a_i| s, \theta_i) $.
	
	Consider the discrete-time MDP under infinite horizon, the policy $ \pi_{\btheta} $ can generate a trajectory $ \tau := (s_0, \ba_0, s_1, \ba_1, \cdots) $ based on the  initial state $ s_0 $ sampled from $ \eta (\cdot)$. In this work, we consider the discounted cumulative reward setting and the global value function is defined as below:
	\begin{align}	V_{\pi_{\btheta}}(s) := \mathbb{E} \bigg[ \sum_{t = 0}^{\infty} \gamma^t \cdot \bar{r}(s_t, \ba_t) \mid s_0 = s \bigg]  \label{def: value func},
	\end{align}
	where we define $ \bar{r}(s_t, \ba_t) := \frac{1}{N}\sum_{i=1}^N r_i(s_t, \ba_t) $ and the expectation is taken over the trajectory $ \tau $ generated from joint policy $ \pi_{\btheta} $. When  $ \pi_{\btheta} $ is fixed, the value function $ V_{\pi_{\btheta}}(s) $ will satisfy the Bellman equation \cite{bertsekas2000dynamic} for all states $ s \in \mathcal{S} $:
	\begin{align}
	V_{\pi_{\btheta}}(s) = \mathbb{E}_{\ba\sim \pi_{\btheta}(\cdot | s), s^\prime \sim \mathcal{P}(\cdot| s, \ba)} \left[ \bar{r}(s, \ba) + \gamma \cdot  V_{\pi_{\btheta}}(s^\prime)  \right]. \label{def: bellman eq}
	\end{align}
	
	\iffalse
	Moreover, the state-action value function (Q-function) is  defined as: 
	\begin{align}
	Q_{\pi_{\theta}}(s,a) := \mathbb{E} \bigg[ \sum_{t = 0}^{\infty} \gamma^t r(s_t, a_t) | s_0 = s, a_0 = a \bigg].  \label{def: Q func}
	\end{align}
	\fi
	
	The objective of RL is to find the optimal policy parameter $ \btheta^* $ which maximizes the expected discounted cumulative reward as below:
	\begin{align}
	 	\max_{\bm \theta} ~ J(\btheta) := \mathbb{E}_{s\sim \eta(\cdot)}[V_{\pi_{\btheta}}(s)] =  \mathbb{E} \bigg[ \sum_{t = 0}^{\infty} \gamma^t \cdot \bar{r}(s_t, \ba_t) \bigg] = \mathbb{E} \bigg[ \sum_{t = 0}^{\infty} \frac{\gamma^t}{N} \sum_{i = 1}^{N} r_{i}(s_t, \ba_t) \bigg] \label{def: marl objective}.
	\end{align}
	
	\iffalse
	The policy parameters $ \theta $ can be optimized along its policy gradient direction \cite{agarwal2019reinforcement} through calculating $ \nabla J(\pi_{\theta}) $ w.r.t. $ \theta $:
	\begin{align}
		\nabla J(\pi_{\theta}) := \frac{1}{1 - \gamma} \mathbb{E}_{s \sim d^{\pi_{\theta}}} \mathbb{E}_{a\sim \pi_{\theta}(\cdot | s)} \bigg[ A^{\pi_{\theta}}(s,a) \nabla \log \pi_{\theta}(a|s) \bigg]  \label{def: pg theorem}
	\end{align}
	where $ A^{\pi_{\theta}}(s,a) := Q^{\pi_{\theta}}(s,a) - V^\pi(s,a) $ denotes the advantage function and $ d^{\pi_{\theta}} $ denotes the discounted state visitation measure where $ d^{\pi_{\theta}}(s) := \mathbb{E}_{s_0 \sim \eta} \left[ (1 - \gamma) \sum_{t = 0}^{\infty} \gamma^t \mathcal{P}(s_t = s| s_0) \right] $.
	\fi
	
	In order to optimize $J(\btheta)$, the policy gradient \cite{sutton2000policy}, could be expressed as
 	\begin{align}
 	    \nabla J(\btheta) := \frac{1}{1- \gamma} \EE_{s \sim d_{\btheta}(\cdot),a \sim \pi_{\btheta}(\cdot|s), s^\prime \sim \mathcal{P}(\cdot|s, \ba)} \big[ \big( \bar{r}(s, \ba) + \gamma V_{\pi_{\btheta}}(s^\prime) \big) \nabla_{\btheta} \log \pi_{\btheta}(\ba|s) \big]. \label{eq:pg_theorem}
 	\end{align}

 	\section{The Proposed Coordinated Actor-Critic Algorithm} \label{sec: alg}
	
%	Assumptions \ref{ass:problem} - \ref{ass: reward_bound} define the setting for the decentralized MARL problem. Our proposed algorithm is based on the decentralized setting to solve the cooperative tasks for each RL agent.

		\subsection{The Proposed Formulation}
	In this section, we describe our MARL formulation. 
	Our proposed formulation is based upon \eqref{def: marl objective}, but with the key difference that we no longer require the agents to have independent policy parameters $\theta_i$. Specifically, we assume that the agents can (partially) share their policy parameters with their neighbors. Hence, each agent will decompose its policy into $ \theta_{i} := \{ \theta_{i}^{s}, \theta_{i}^{p} \} $, where the shared part $\theta^{s}_i$   has the same dimension across all agents, and the personalized part $\theta^{p}_i$ will be kept locally.

	The above partially personalized policy structure leads to the following MARL formulation: 
	\begin{align} 
	\max_{\btheta} \quad & J(\btheta) := \EE_{s \sim \eta(\cdot)} \bigg[ V_{\pi_{\btheta}}(s) \bigg] = \EE \bigg[ \sum_{t = 0}^{\infty} \gamma^t\cdot  \bar{r}(s_t,\ba_t) \bigg]  \label{eq: formulation 1} \\
	{\rm s.t}. \quad	& \theta^{s}_i = \theta^{s}_j\; \text{ if } (i, j) \text{ are neighbors}\nonumber %\label{eq: formulation 2}
	\end{align}
	where $ \btheta := \{\theta_i^s, \theta_i^p\}_{i = 1}^N $ is the collections of all local policy parameters $ \theta_{i} := \{ \theta_i^s, \theta_i^p \} $.
	%{\red[need transition to the following problem. need to discuss parameterization, approximation, etc, before jumping to the complete formulation. Now it is too sudden. ]}
%	{\red[how to connect (7) and its bilevel formulation (8)? is (8) an approximation of (7) so that we can do easier decentralized implementation? or is (8) equivalent to (7)? At this point, it is not very clear.]}
To cast problem \eqref{eq: formulation 1} into a more tractable form, we perform the following steps.

First, we approximate the global reward function for any $s \in \mathcal{S}$ and $\ba \in \mathcal{A}$. Specifically, we use the following linear function $ \widehat{r}(s, \ba; \lambda) := \varphi(s, \ba)^T \lambda $ to approximate the global reward $ \bar{r}(s,\ba) := \frac{1}{N} \sum_{i=1}^N r_i(s,\ba) $, where $ \varphi(\cdot, \cdot) : \mathcal{S} \times \mathcal{A} \to \mathbb{R}^L$ is the feature mapping. Then the optimal parameter $\lambda^*(\bm \theta)$ can be found by solving the following problem:
	\begin{subequations} \label{eq:decentralized_reward_evaluation}
	   \begin{align}
	   \lambda^*(\bm \theta) & \in \underset{ \lambda }{\text{arg min}} ~ \mathbb{E}_{s \sim \mu_{\btheta}(\cdot), \ba \sim \pi_{\btheta}(\cdot|s)} \bigg[\big(\frac{1}{N}\sum_{i=1}^{N} r_i (s,\ba) - \varphi(s,\ba)^T\lambda \big)^2 \bigg] \label{eq:average_reward_approx} \\
	   & = \underset{ \lambda }{\text{arg min}} ~ \sum_{i=1}^{N}\mathbb{E}_{s \sim \mu_{\btheta}(\cdot), \ba \sim \pi_{\btheta}(\cdot|s)} \left[\left( r_i (s,\ba) - \varphi(s,\ba)^T\lambda\right)^2\right]. \label{eq:decompose_average_reward_approx}
	\end{align}
	\end{subequations}

Second, we approximate the global value function $V_{\pi_{\btheta}}(s)$ for any $s \in \mathcal{S}$ under a fixed joint policy $\pi_{\btheta}$. Specifically, we use the following linear function 
$\widehat{V}(s;\omega) := \phi(s)^T \omega$ to approximate the global reward function $V_{\pi_{\btheta}}(s)$, where $ \phi(\cdot): \mathcal{S} \to \mathbb{R}^K $ is a given feature mapping. 
Towards achieving the above approximation, we can solve the following mean squared Bellman error (MSBE) minimization problem \cite{tsitsiklis1997analysis}:
%and following the standard scheme in single-agent policy evaluation problem \eqref{obj:single_agent_TD}, it is typical to find $w^*(\btheta)$ under fixed policy parameters $ \btheta := \{ \theta_i \}_{i=1}^N $ such that the following
%mean squared  Bellman error (MSBE) is minimized:
\begin{subequations}
\label{eq:Decentralized_TD_evaluation}
   \begin{align}
    \omega^*(\btheta) &\in \arg\min_{\omega}\mathbb{E}_{s \sim \mu_{\btheta}(\cdot), \ba \sim \pi_{\btheta}(\cdot|s), s^\prime \sim \mathcal{P}(\cdot|s,\ba)} \left[\big(  \frac{1}{N}\sum_{i=1}^{N}r_i(s,\ba) + \gamma  \widehat{V} (s^{\prime};\omega)- \widehat{V} (s;\omega)\big)^2\right] \label{eq:dec_TD_eval} \\
    & \!=\! \arg\min_{\omega}\mathbb\sum_{i=1}^{N}{\mathbb{E}}_{s \sim \mu_{\btheta}(\cdot), \ba \sim \pi_{\btheta}(\cdot|s), s^\prime \sim \mathcal{P}(\cdot|s,\ba)} \left[\big(r_i(s,\ba) + \gamma \widehat{V} (s^{\prime};\omega)-\widehat{V} (s;\omega) \big)^2\right]. \label{eq:decompose_dec_TD_eval}
    \end{align}
\end{subequations}
% Extending the optimal condition in \eqref{eq:single_agent_fixed_critic} from single-agent setting to our multi-agent setting, it holds that there is a unique optimal solution $\omega^*(\btheta)$ in \eqref{eq:Decentralized_TD_evaluation} satisfying

To separate the objective into the sum of $N$ terms (one for each agent),  we introduce local copies of $w$ and $\lambda$ as $\{w_i\}_{i=1}^{N}$, $\{\lambda_i\}_{i=1}^{N}$, and define their vectorized versions $\boldsymbol{\omega} = [\omega_1, \cdots, \omega_N]^T$ and $ \boldsymbol{\lambda} = [\lambda_1, \cdots, \lambda_N]^T $.
	Similarly, we also define $\boldsymbol{\omega}^*(\btheta) := [\omega_1^*(\btheta), \cdots, \omega_N^*(\btheta)]^T$ and $\boldsymbol{\lambda}^*(\btheta) :=  [\lambda_1^*(\btheta), \cdots, \lambda_N^*(\btheta)]^T$. %as the optimal parameters to approximate global value function and global reward function under the fixed policy $\pi_{\btheta} := \cup_{i=1}^N \pi_{\theta_i}$. %{\blue Moreover, under linear approximations, the uniqueness property of optimal solutions in \eqref{eq:decentralized_reward_evaluation} and \eqref{eq:Decentralized_TD_evaluation} leads that $\boldsymbol{\omega}^*(\btheta)$ and $\boldsymbol{\lambda}^*(\btheta)$ are in consensus, which means $\omega^*(\btheta) := \omega_1^*(\btheta) = \cdots = \omega_N^*(\btheta)$ and $ \lambda^*(\btheta) := \lambda_{1}^*(\btheta) = \cdots = \lambda_{N}^*(\btheta) $. }
	\iffalse
Then the objective function of problem \eqref{eq: formulation 1} can be approximated by: 
\begin{align}
    %\mathbb{E}_{s \sim \eta} \bigg[ \widehat{V}_{\pi_{\theta}}(s;w)] \bigg] %\stackrel{} 
    J(\btheta)
    &\approx %\mathop{\mathbb{E}}\limits_{s\in\eta, a\sim \pi(\cdot | s), s^\prime \sim \mathcal{P}(\cdot| s, a)}
    \EE_{s \sim \eta_{\btheta}(\cdot), \ba \sim \pi_{\btheta}(\cdot|s), s^\prime \sim \mathcal{P}(\cdot|s,\ba)} \left[\frac{1}{N}\sum_{i=1}^{N}\bigg( \widehat{r}(s,\ba; \lambda_i) + \gamma \cdot  \widehat{V} (s^{\prime};\omega_i) \bigg) \right]\nonumber\\
     & = \EE_{s \sim \eta_{\btheta}(\cdot), \ba \sim \pi_{\btheta}(\cdot|s), s^\prime \sim \mathcal{P}(\cdot|s,\ba)} \left[ \frac{1}{N}\sum_{i=1}^{N} \bigg( \psi(s,\ba)^T \lambda_i + \gamma \cdot \phi(s^\prime)^T \omega_i \bigg) \right] 
\end{align}
\fi
%where the function approximators are used due to the lack of availability of global reward function and global value function in each agent $i \in \mathcal{N}$.}
	
	Summarizing the above discussion, problem \eqref{eq: formulation 1} can be approximated using the following bi-level optimization problem: %{\red[note, in the objective, $\hat{r}$ has to be used, because otherwise there is no need to define $\hat{r}$]}: 
%	One challenge here is how to approximate the global reward function $\bar{r}(s,a)$ and $V_{\pi_\theta}(s)$ through only leveraging local reward signal $r_i(s,a)$ for all agent $i \in \mathcal{N}$. To tackle this challenge, under a fixed policy $\pi_\theta$, the approximation problem for global reward function and global value function could be formulated as decentralized optimization problems. First, the approximation to global 
%	and reformulate  problem \eqref{eq: formulation 1} into the following bi-level optimization problem:} {\red[still not clear how (8) comes from (7); if you directly apply the previous two approximation into (7), how (8) is obtained?]}
%	To solve the  MARL formulation \eqref{eq: formulation 1}, in a decentralized manner, we cast the problem into a bi-level optimization framework where a actor-critic method with personalization is proposed to fit this framework. In our formulation, all agents will collaborate with each other to approximate the system-level value function (with linear function or neural networks). In the meanwhile, with the help of approximation to value function, the agents will perform policy improvement steps to optimize their common objective. From this perspective, we understand that the quality of the policy improvement steps highly depends on the accuracy of the value function approximators. Here, we present our complete formulation as below:
{\small\begin{subequations}\label{eq:bi-level}
\begin{align} \label{bi-level: formulation 1}
	\max_{\btheta} \quad & \mathop{\mathbb{E}} \limits_{ \substack{s \sim \eta(\cdot), \ba \sim \pi_{\btheta}(\cdot| s) \\ s^\prime \sim \mathcal{P}(\cdot|s,\ba) }} \bigg[ \frac{1}{N} \sum_{i = 1}^{N} \bigg( \widehat{r}(s,\ba; \lambda^*_i(\btheta))+ \gamma \cdot \widehat{V} (s^\prime; \omega^*_i(\btheta)) \bigg) \bigg]   \\
	s.t. \quad	& \bomega^* (\btheta) \in \underset{\omega }{\text{arg min}} ~ \sum_{i=1}^{N} \mathop{\mathbb{E}} \limits_{ \substack{s \sim \mu_{\btheta}(\cdot), \ba \sim \pi_{\btheta}(\cdot| s) \\ s^\prime \sim \mathcal{P}(\cdot|s,\ba) }} \bigg[\big( r_i(s,\ba) + \gamma \cdot \widehat{V} (s^\prime; \omega_i) - \widehat{V} (s; \omega_i) \big)^2 \bigg],  \label{bi-level: formulation 2} \\
 \quad	& \blambda^*(\btheta) \in \underset{\lambda }{\text{arg min}} ~ \sum_{i=1}^{N} \mathbb{E}_{s \sim \mu_{\btheta}(\cdot), \ba \sim \pi_{\btheta}(\cdot| s)} \bigg[\big( r_i(s,\ba)  - \widehat{r}(s, \ba; \lambda_i) \big)^2 \bigg],  \label{bi-level: formulation 3} \\
\quad & \theta_i^s = \theta_j^s,~ \omega^*_i( \btheta) = \omega^*_j( \btheta), ~ \lambda^*_i(\btheta) = \lambda^*_j(\btheta), ~  \text{ if } (i, j) \text{ are neighbors}. \label{bi-level: formulation 4}
\end{align}
\end{subequations}}
%    where at time $ t $ the global reward $ \bar{r}(s,a) $ and the global value function $ V_{\pi_\theta}(s) $ are linearly approximated by parameters $ \lambda_{i,t} \in \mathbb{R}^L $ and $ \omega_{i,t} \in \mathbb{R}^K $ at each agent $ i $. For the linear approximation, it follows that $ \widehat{r}(s_t, a_t; \lambda_{i,t}) := \varphi(s_t, a_t)^T \lambda_{i,t} $ and $ \widehat{V}(s_t ; \omega_{i,t}) := \phi(s_t)^T \omega_{i,t} $ where $ \varphi(\cdot, \cdot) : \mathcal{S} \times \mathcal{A} \to \mathbb{R}^L$ and $ \phi(\cdot): \mathcal{S} \to \mathbb{R}^K $ are feature mappings. %Based on the proposed problem formulation, we make a few assumptions on the function approximations here, which specify the basic properties of the feature mappings $ \phi(s) $ and $ \varphi(s, a) $ for all $s \in \mathcal{S}, a \in \mathcal{A}$.
%In the above, for  each agent $ i $, $ \omega_i(\theta) $ denotes the solution of the lower-level problem under the joint policy $ \pi_{\theta} := \{ \pi_{\theta_i} \}_{i=1}^N $. 

In the subsequent discussion, we will refer to the problem of finding the optimal policy $\btheta$ as the {\it upper-level} problem, while referring to the problem of finding the optimal $ \bomega^* (\btheta) $ and $ \blambda^* (\btheta) $
under a fixed policy parameters as the lower-level problem.

    	\subsection{The Proposed Algorithm} \label{sub: alg}
    	
    	In this subsection, we first present the assumptions related to network connectivity and communication protocols in the multi-agent systems. Then we describe the proposed Coordinated Actor-Critic (\aname) algorithm which is summarized in Algorithm \ref{proposed algorithm}.
    	
    	\begin{assumption}[Network Connectivity]\label{ass:connectivity}
		There exists an integer $ B $ such that the union of the consecutive $ B $ graphs is connected for all positive integers $ \ell $. That is, the following graph is connected:
		$$ \bigg(\mathcal{N}, \mathcal{E}(\ell \cdot B) \cup \mathcal{E}(\ell\cdot B + 1) \cdots \cup \mathcal{E}((\ell + 1) B - 1) \bigg), \;\forall~\ell\ge 1$$
		where $ \mathcal{N} $ denotes the vertice set  and $ \mathcal{E}(t) $ denotes the set of active edges at time $ t $.
	\end{assumption} 
	
	\begin{assumption}[Weight Matrices] \label{ass: mixing_matrix}
		There exists a positive constant $ c $ such that $W_t = [W_t^{ij}] \in \mathcal{R}^{N \times N}$  is doubly stochastic and $W_t^{ii} \geq c \text{   }$ for all $i \in \mathcal{N}$. Moreover, $W_t^{ij} \in [c ,1)$ if $(i,j) \in \mathcal{E}(t)$, otherwise $W_t^{ij} = 0$ for all $i,j \in \mathcal{N}$. %Also $c_2(W_t) < 1$, where $ c_2(W_t) $ denotes the second largest eigenvalue of the weight matrix $W_t$ for all time step $t$. %{\red[I think you need more assumptions than this, for example the null space assumption.]}
	\end{assumption}
	
	Assumption \ref{ass:connectivity} ensures that the graph sequence
    is sufficiently connected for each agent to have repeated influence on other agents. Assumption \ref{ass: mixing_matrix} is standard in developing decentralized algorithms \cite{nedic2009distributed}, which could guarantee consensus results for shared parameter in each agent converging to a common vector.
    	
    \iffalse
    Let us use the subscript $``t"$ to denote the iteration number. 
    First, let us specify the weight matrix to be used by the agents. We consider the networked system as a graph which is time-varying and undirectional. For a given network, we can then define a mixing matrix $ W_t := [W_t^{ij}]_{N \times N}$ at each iteration $ t $. %where $ W_t^{ij}$ is the weight coefficient for messages transmitted from the agent $i$ to the agent $j$ at time $ t $.
    Roughly speaking,  the entries of $W_t$ are non-negative and $ W_t^{ij} = 0 $ if $ (i, j) \notin \mathcal{E}_t $. Later in Assumption \ref{ass: mixing_matrix}, we provide detailed conditions on $W_t$.	
    \fi
    	 
    After presenting the assumptions related to the network topology in the decentralized system, we are able to introduce the proposed \aname ~algorithm. The \aname ~algorithm takes two main steps, the policy optimization step (which optimizes $\btheta$),  and policy evaluation step (which approximately solves the lower-level problem in \eqref{eq:bi-level}), as we describe below. For simplicity, we denote $\bar{r}(s_t, \ba_t)$ as $\bar{r}_t$ and $r_{i}(s_t, \ba_t)$ as $r_{i,t}$.
    %{\red[it appears at certain places you used $W^{i,j}$ for the (i,j)th entry of matrix?]} } %{\red[collect the assumption of graph, as well as the property of W, into an assumption.]}

	\iffalse
	  Although each agent in the networked system faces different tasks and their actions may have different dimensions, they have similar dynamics and can receive a same observation (the global state). Therefore, in order to learn the \textcolor{red}{common knowledge} among different agents for better behavior strategies, we propose a general setting that each agent can (partially) share their policy parameters with their neighbors.
	 % (\textcolor{red}{this is a key observation, worth expanding...})
	\fi

		\begin{algorithm}[t] 
		\caption{{\small {\it Coordinated Actor-Critic (\aname) Algorithm}}}% $(\bW^{(0)},L_{\max},L,\rho,\beta,\delta,\Delta f)$}
		\begin{algorithmic}[1]
			\State {\bfseries Input:} Parameters $ \{\alpha_t \}_{t = 0}^{T-1} $, $ \{\beta_t \}_{t = 0}^{T-1} $, $ \{\zeta_t \}_{t = 0}^{T-1} $. Initialize $ \theta_{i,0}, \omega_{i,0}, \lambda_{i,0}$ for all $ i \in \mathcal{N} $
			\For{$t=0,1,\ldots, T-1$}
			\State \textbf{Data Sampling}: $ s_t \sim \mu_{\btheta_{t}}(\cdot) $, $ \ba_t := \{a_{i,t} \sim \pi_i(\cdot| s_t, \theta_{i,t})\}_{i=1}^N $, $ s_{t+1} \sim \mathcal{P}(\cdot| s_t, \ba_t) $
			\State \textbf{Consensus Step}: $ \widetilde{\bomega}_t = W_t \cdot \bomega_t$, $ \widetilde{\blambda}_t = W_t \cdot \blambda_t$ and $ {\widetilde{\btheta}}_t^s := W_t \cdot \btheta_t^s $
			\For{$ i \in \mathcal{N} $}
			\State Construct $ {\widetilde{\theta}}_{i,t} = \{ {\widetilde{\theta}}^s_{i,t},  \theta^p_{i,t} \} $
			\State Receive local reward $ r_{i,t} $ and global observation $ (s_t, \ba_t, s_{t+1}) $
			\State Update $ \delta_{i,t} = r_{i,t} + \gamma\cdot \phi(s_{t+1})^T \omega_{i,t} - \phi(s_{t})^T \omega_{i,t} $
			\State $ \omega_{i,t+1} = \Pi_{R_\omega} \big(\widetilde{ \omega}_{i,t} + \beta_t \cdot \delta_{i,t} \cdot \phi(s_t) \big) $
			\State $\lambda_{i, t+1} = \Pi_{R_\lambda}\bigg(\widetilde{ \lambda}_{i,t} + \zeta_t \cdot \big(r_{i,t} - \varphi(s_t,\ba_t)^T \lambda_{i,t} \big) \cdot  \varphi(s_t,\ba_t) \bigg)$
			\State $  \theta_{i,t+1} = {\widetilde{\theta}}_{i,t} + \alpha_t \bigg( \varphi(s_t,\ba_t)^T \lambda_{i,t} + \gamma \cdot  \phi(s_{t+1})^T \omega_{i,t} - \phi(s_{t})^T \omega_{i,t} \bigg)\cdot  \nabla_{ \theta_i} \log \pi_i( a_{i,t}| s_t, \theta_{i,t}) $
			\EndFor
			\EndFor
		\end{algorithmic}
		\label{proposed algorithm}
		%\vspace{-2px}
	\end{algorithm}

	 \noindent{\bf Policy Optimization.} In this step, 
	 the agents optimize their local policy parameters, while trying to make sure that the shared parameters are not too far from their neighbors. 
	 
	 Towards this end, 	each agent $i$ first produces a {\it locally averaged} shared parameter by linearly combining with its neighbors’ current shared parameters. Such an operation can be expressed as 
	 \begin{align}
	     \widetilde{\btheta}^s_t := W_t \cdot \btheta^s_{t} \label{eq:actor_consensus}
	 \end{align}
	 \iffalse
	   \begin{align}
%	   \widetilde{\theta}^s_{i,t} := \sum_{j\in \mathcal{N}_i}W^{ij}_{t}  \widetilde{\theta}^s_{j,t} \cdot \btheta^s_{t}
	  &\widetilde{\btheta}^s_t := W_t \cdot \btheta^s_{t}, \quad \label{alg: actor_consensus}
	  \end{align} 
	  \fi
	  where $ \btheta^s_{t} := [\theta^s_{1,t}, \theta^s_{2,t}, \cdots;\theta^s_{N,t}]^T \in \mathbb{R}^{N \times H} $ is a matrix which stores all parameters $ \{\theta^s_{i,t}\}_{i=1}^N $, and $\widetilde{\btheta}^s_t$ is defined similarly.
	  In the decentralized setting, the global reward $\bar{r}_t$ and the global value function $V_{\pi_{\btheta_t}}(\cdot)$ are not available for each agent $i$. Instead, the agents can locally estimate the global reward and the global value function using some linear approximation, evaluated on their local variables, as described in the previous subsection.
% 	  That is, we can define the local linear approximation for the global value function $ V_{\pi_{\theta}}(s_t) $  and the global rewards $\bar{r}_t$ by:
% 	  \begin{align}\label{eq:value:reward}
% 	  \widehat{V}(s_t; \omega_{i,t}) := \phi(s_t)^T \omega_{i,t},  \quad \widehat{r}_{i,t} = \varphi(s_t, a_t)^T \lambda_{i,t}
% 	  \end{align}
	  	 %$$  $$ 
	  As shown in line $11$ of Algorithm 1, in a decentralized system, we consider the policy optimization step for each agent as below:
	  	  \begin{align}
	  %	  &\widetilde{\theta}^s_t := W_t \theta^s_{t}, \quad \widetilde{\theta}_{i,t} := \{\widetilde{\theta}^s_{i,t}, {\theta}^l_{i,t}\} \label{alg: actor_consensus} \\
	  \theta_{i,t+1} & :=  \widetilde{\theta}_{i,t} + \alpha_t \cdot \widehat{\delta}_{i,t}  \cdot \nabla_{ \theta_i} \log \pi_i( a_{i,t}| s_t, \theta_{i,t}), ~~ \forall i \in \mathcal{N} \label{alg: actor_local updates}\\
	  {\rm where}\quad  \widehat{\delta}_{i,t} & := \widehat{r}(s_t, \ba_t; \lambda_{i,t}) + \gamma\cdot \widehat{V}(s_{t+1}; \omega_{i,t}) - \widehat{V}(s_t ; \omega_{i,t}) \label{eq:local:td}.
	  \end{align}
	  %{\red [I think it is important to clearly denote the subsecript of $V$, as $\btheta_{t+1}$] }
	  
%	  {\red[is $\widehat{r}(s_t, a_t; \lambda_{i,t})$ related to $\hat{r}_{i,j}$ defined in (16)?]}
	  %where $$.  %{\red[where is $\lambda$ coming from?]} For the linear approximation, it follows that $ \widehat{r}(s_t, a_t; \lambda_{i,t}) := \varphi(s_t, a_t)^T \lambda_{i,t} $ and $ \widehat{V}(s_t ; \omega_{i,t}) := \phi(s_t)^T \omega_{i,t} $ where $ \varphi(\cdot, \cdot) : \mathcal{S} \times \mathcal{A} \to \mathbb{R}^L$ and $ \phi(\cdot): \mathcal{S} \to \mathbb{R}^K $ are feature mappings.
	  
%	  Therefore, the actor step (\ref{alg: actor_update}) which performs local gradient update for the policy parameters $\theta_i$ of each agent $i$ can be rewritten as follows:
%	  \begin{align}
%	    \theta_{i,t+1} :=  \widetilde{\theta}_{i,t} + \alpha_t \Delta_{i,t}  \nabla_{ \theta_i} \log \pi_i( a_{i,t}| s_t, \theta_{i,t}) ~~ \forall i \in \mathcal{N}. \label{alg: actor_local updates}
%	  \end{align} 

 \noindent{\bf Policy Evaluation.} Next, we update the local parameters $\lambda_{i,t}$ and $\omega_{i,t}$, which parameterize the global reward function and global value function. 
 %Recall  that the value functions and reward functions have been parameterized, so estimating the value function should be done by updating the parameters. 
 Towards this end, the parameters $\lambda_{i,t}$ and $\omega_{i,t}$ will be updated by first averaging over their neighbors, then performing one stochastic gradient descent step to minimize the  local objectives, which are defined as in \eqref{bi-level: formulation 2} - \eqref{bi-level: formulation 3} and under consensus constraints \eqref{bi-level: formulation 4}.  %{\red[add local objectives.]}
That is, we have the following updates for $\blambda_t$ and $\bomega_t$: % be leveraged. First, both of $\lambda_{i,t}$ and $\omega_{i,t}$ will be linearly combined according its neighbor’s parameter estimates. After performing the local consensus steps, they will be optimized along their gradient directions. Denoting the $i$-th agent's estimation of global reward $\bar{r}_t$ as $\widehat{r}_{i,t} := \widehat{r}(s_t, a_t; \lambda_{i,t}) $, the update steps for the parameters $\lambda_{i,t}$ and $\omega_{i,t}$ are expressed as below:
	\begin{align}
	    &\widetilde{\blambda}_t = W_t \cdot \blambda_{t}, \quad \lambda_{i, t+1} = \Pi_{R_\lambda}\bigg(\widetilde{ \lambda}_{i,t} + \zeta_t \cdot \big(r_{i,t} - \widehat{r}(s_t, \ba_t; \lambda_{i,t})\big) \cdot \nabla_{ \lambda_i} \widehat{r}(s_t, \ba_t; \lambda_{i,t}) \bigg),  \label{alg: reward updates} \\
		&\widetilde{\bomega}_t = W_t \cdot \bomega_{t}, \quad \omega_{i,t+1} = \Pi_{R_\omega} \bigg(\widetilde{ \omega}_{i,t} + \beta_t \cdot \delta_{i,t} \cdot \nabla_{ \omega_i} \widehat{V}(s_t; \omega_{i,t}) \bigg), \quad  \forall i \in \mathcal{N} \label{alg: critic updates}
	\end{align}
%	where we have defined
%Using the above definitions,  {\blue we define another local TD error which could be utilized to perform gradient update in value function approximators}: %{\red[it needs to be better transitioned from above and below.]}
	where we define $ \delta_{i,t} := r_{i,t} + \gamma \cdot \widehat{V}(s_{t+1} ; \omega_{i,t}) - \widehat{V}(s_t; \omega_{i,t}). $
	Moreover, $ \Pi_{R_\omega} \left( \cdot \right) $ and $ \Pi_{R_\lambda} \left( \cdot \right) $ are the projection operators, with $ R_\omega $ and $R_\lambda$ being the predetermined projection radii which are used to stabilize the update process \cite{tsitsiklis1997analysis}. Please see lines $8$-$10$ in Algorithm \ref{proposed algorithm}.

	\section{Theoretical Results}
	In this section,  we first present Assumptions \ref{ass: reward_bound} - \ref{ass:feature_bound} about reward function and linear approximations for policy evaluation. Then we show our theoretical results for the proposed \aname ~algorithm.
	
	    \begin{assumption}[Bounded Reward] \label{ass: reward_bound}
		All the local rewards $ r_i(s,a) $ are uniformly bounded, i.e., there exist constants $ R_{\max} $, for all $ i \in \mathcal{N} $ and $s \in \mathcal{S}$ such that $|r_i(s,a)| \leq R_{\max}$.
	\end{assumption}
	
    	\begin{assumption}[Function Approximation] \label{ass:feature_bound}
	    For each agent $i$, the value function and the global reward function are both parameterized by the class of linear functions, i.e., $\widehat{V}(s; \omega_i) := \phi(s)^T \omega_i$ and $\widehat{r}(s, \ba; \lambda_i) := \varphi(s,\ba)^T \lambda_i$ where we denote $\phi(s): = [\phi_1(s),\cdots ,\phi_K(s)]^T \in \mathbb{R}^K$ and $\varphi(s,\ba) = [\varphi_1(s,\ba),\cdots ,\varphi_L(s,\ba)]^T \in \mathbb{R}^L$ are the feature vector associated with $s$ and $(s,\ba)$, respectively. The feature vectors $\phi(s)$ and $ \varphi(s,\ba) $ are uniformly bounded for any $s \in \mathcal{S}, \ba \in \mathcal{A}$, i.e., $ \| \phi(s) \| \leq 1 $ and $ \| \varphi(s,\ba) \| \leq 1 $. Furthermore, constructing the feature matrix $ \Phi \in \mathbb{R}^{|\mathcal{S}| \times K} $ which has $[\phi_k(s),s \in \mathcal{S}]^T$ as its $k$-th column for any $ k \in K $. Also constructing the feature matrix $ \Psi \in \mathbb{R}^{|\mathcal{S}|\cdot|\mathcal{A}| \times L} $ which has $[\varphi_l(s),s \in \mathcal{S}]^T$ as its $\ell$-th column for any $ \ell \in L $. Then, we further assume both $ \Phi $ and $ \Psi $ have full column ranks.
	\end{assumption}
	
	Assumption \ref{ass: reward_bound} - \ref{ass:feature_bound} are common in analyzing TD with linear function approximation; see e.g., \cite{konda2000actor,bhandari2018finite,wu2020finite}. With global observability, each agent could construct linear function approximations of the global value function and global reward function.
	Under these assumptions, it is guaranteed that there exist unique optimal solutions $\lambda^*(\btheta)$ and $\omega^*(\btheta)$ to approximate the global reward function in \eqref{eq:decentralized_reward_evaluation} and the global value function in \eqref{eq:Decentralized_TD_evaluation} with linear functions. It is crucial to have the properties of unique optimal solutions in $\lambda^*(\btheta)$ and $\omega^*(\btheta)$ for constructing the convergence analysis of policy parameters $\btheta$.

%{\red[looks like assumption 2 and 3 needs to be moved to the previous sections. We have already introduced the linear approximation before, now we re-introduced them. It is redundant.]}
    
	%\begin{assumption}[value function approximation] \label{ass:feature_state_action_bound}
	    %For each agent $i$, the global reward $\bar{r}$ is parameterized by the class of linear functions %{\red[I don't understand this. what does it mean by for each agent, the ``global" reward satisfies... the global rewards are different for each agent?? We shall just state the parameterization for global reward, not for each agent. $\lambda_i$ will be introduced later when discussing algorithms. It has nothing to do with assumptions.]}
	    %, i.e., $\widehat{r}(s, a; \lambda_i) := \varphi(s,a)^T \lambda_i$ where $\varphi(s,\textcolor{red}{a}) = [\varphi_1(s,\textcolor{red}{a}),\cdots ,\varphi_L(s,\textcolor{red}{a})]^T \in \mathbb{R}^L$ is the feature associated with the state $s$ and the joint action $a$. The size of the feature vector $\varphi(s, a)$ is uniformly bounded for any $s \in \mathcal{S} \text{ and } a \in \mathcal{A}$, i.e., $ \| \varphi(s, a) \| \leq 1 $. Furthermore, the feature matrix {\red[define this]} $ \Psi \in \mathbb{R}^{|\mathcal{S}|\cdot|\mathcal{A}| \times K} $ has full column rank. %where the $l$-th column of $\Psi$ is $[\varphi_l(s),s \in \mathcal{S}, a\in \mathcal{A}]^T$ for any $l \in [L]$. {\red[this is a bit redundant with the previous section?]}
	%\end{assumption}
	
	{Due to space limitation, we relegate remaining technical assumptions (i.e., Assumptions \ref{ass: Lipschitz policy} - \ref{ass: Markov chain})} to Appendix \ref{Appendix:Assumptions} and technical lemmas to Appendix \ref{appendix:auxiliary_lemmas}.
	 We first present the convergence speed of the variables $\{\bomega
	_t\}$ and $\{\blambda_t\}$ for the policy evaluation problem defined in \eqref{bi-level: formulation 2} - \eqref{bi-level: formulation 4}.
	Please see Appendix \ref{appendix:proposition_critic} for the detailed proof.
 	
	\begin{proposition}\label{proposition:critic}
		Suppose Assumptions \ref{ass:connectivity} - \ref{ass: Markov chain} hold. For any iteration $t$, by selecting stepsizes $$ \alpha_t = \frac{\alpha_0}{T^{\sigma_1}},\,  \beta_t = \frac{\beta_0}{T^{\sigma_2}},\, \zeta_t = \frac{\zeta_0}{T^{\sigma_2}} $$
	where $ 0 < \sigma_2 < \sigma_1 < 1 $ and $ \alpha_0, \beta_0, \zeta_0 > 0 $ are some fixed constants, the following holds:
		\begin{align}
		&\frac{1}{T} \sum_{t = 0}^{T-1} \sum_{i=1}^N \bigg( \mathbb{E}\bigg[\| \omega_{i,t} - \omega^*(\btheta_t) \|^2\bigg] + \mathbb{E}\bigg[\| \lambda_{i,t} - \lambda^*(\btheta_t) \|^2\bigg] \bigg) \nonumber \\
		=& \mathcal{O}(T^{-1+\sigma_2}) + \mathcal{O}(T^{ - \sigma_2}) + \mathcal{O} \left( T^{\sigma_2 - 2\sigma_1} \right) + \mathcal{O}(T^{-2\sigma_1 + 2\sigma_2}) + \mathcal{O}(T^{-2 + 2\sigma_2}) + \mathcal{O}(T^{-2\sigma_2}) \nonumber
		\end{align}
		 where the expectation is taken over the data sampling procedure as shown in line 3 of Algorithm \ref{proposed algorithm}.
	\end{proposition}
	
 Compared with existing works \cite{wai2018multi,doan2019finite} which established finite-time convergence guarantees for decentralized policy evaluation problems under the fixed policy, our results in Proposition \ref{proposition:critic} are analyzed in a more challenging situation where both policies and critics are updated in an alternating manner. Here, we must set $\sigma_1 > \sigma_2 $ to ensure that the relation above is useful. This is reasonable since the optimal critic parameter $ \omega^*(\btheta_t) $ is constantly drifting as the policy parameters $ \btheta_t $ changes at each iteration, so the actor should update slowly compared with the critic.
	
	Next, we study the convergence rate of policy parameters. We define $Q := I - \frac{1}{N}\bm 1 \bm 1^T$ and define the average gradient of shared policy parameters as $ \overline{\nabla_{\btheta^s} J(\btheta)} := \frac{1}{N} \sum_{i=1}^N \nabla_{\theta^s_i} J(\btheta) $. We will show that after averaging over the iterations, the expected stationarity condition violation for the policy optimization problem defined in \eqref{bi-level: formulation 1} is small. Please see Appendix \ref{appendix:actor_convergence} for the proof.
	
	\iffalse
	Towards this end, we define the critic approximation error and the sampling error, respectively, as follows:  \cite{wu2020finite,hong2020two,qiu2021finite}: %where the approximation error is inevitable due to the use of linear approximation.
	\begin{align}
	    \epsilon_{\text{app}} & := \max_{\btheta} \sqrt{ \mathop{\mathbb{E}} \limits_{ \substack{s \sim \mu_{\btheta}(\cdot) }} \left[ \bigg( V_{\pi_{\btheta}}(s) - \phi(s)^T \omega^*(\btheta) \bigg)^2 \right] } + \sqrt{ \mathop{\mathbb{E}} \limits_{ \substack{s \sim \mu_{\btheta}(\cdot) \\ a \sim \pi_{\btheta}(\cdot) }} \left[ \bigg( \bar{r}(s,\ba) - \varphi(s,\ba)^T\lambda^*(\btheta) \bigg)^2 \right] } \label{error:approximation} \\
	   \epsilon_{\text{sp}} & := 4\cdot R_{\max}\cdot  C_{\psi} \cdot L_v\cdot \left( \log_{\tau} \kappa^{-1} + \frac{1}{1 - \tau} \right) \label{error:sampling}
	\end{align}
	where $\mu_{\btheta}(\cdot)$ is the stationary distribution of state under policy $\pi_{\btheta}$ and the transition kernel $\mathcal{P}(\cdot)$. 
	\fi
	
	\begin{proposition}
	    \label{proposition:actor} %Select the stepsize $ \alpha_t = \frac{\alpha_0}{(1+t)^{\sigma_1}} $, $ \beta_t = \frac{\beta_0}{(1+t)^{\sigma_2}} $ and $ \zeta_t = \frac{\zeta_0}{(1+t)^{\sigma_2}} $ where $ 0 < \sigma_2 < \sigma_1 < 1 $ and $ \alpha_0, \beta_0, \zeta_0 > 0 $. 
		Under the same setting as Proposition \ref{proposition:critic}, there exist two constant error term $\epsilon_{app} > 0 $ and $\epsilon_{sp}>0$. Algorithm \ref{proposed algorithm} generates a sequence of policies $\{\btheta_t\}$, which satisfies the following:
		\begin{align}
		&  \frac{1}{T} \sum_{t = 0}^{T-1} \bigg( \mathbb{E}\bigg[ \|Q\cdot \btheta^s_t \|^2 \bigg] + N \cdot \mathbb{E} \bigg[ \| \overline{\nabla_{\theta^s} J(\btheta_{t})}   \|^2 \bigg] + \sum_{i = 1}^{N} \mathbb{E} \bigg[ \| \nabla_{\theta_i^p} J( \btheta_{t}) \|^2 \bigg] \bigg)  \nonumber \\
		=& \mathcal{O}(T^{- 1 + \sigma_1}) + \mathcal{O}(T^{-\sigma_1}) + \mathcal{O}(T^{-1+\sigma_2}) + \mathcal{O}(T^{ - \sigma_2}) + \mathcal{O} \left( T^{\sigma_2 - 2\sigma_1} \right) 
		+ \mathcal{O}(T^{-2\sigma_1 + 2\sigma_2})   \nonumber \\
		&  + \mathcal{O}(T^{-2 + 2\sigma_2}) + \mathcal{O}(T^{-2\sigma_2}) + \mathcal{O}\left(\epsilon_{app} + \epsilon_{sp} \right). \nonumber
		\end{align}
		%When two sampling procedure are implemented for actor and critic update separately, the last term sampling error is removed as below.
		%\begin{align}
		%&  \frac{1}{T} \sum_{t = 0}^{T-1} \bigg( N \mathbb{E} \bigg[ \| \overline{\nabla_{\theta^s} J(\theta_{t})}   \|^2 \bigg] + \sum_{i = 1}^{N} \mathbb{E} \bigg[ \| \nabla_{\theta_i^l} J( \theta_{t}) \|^2 \bigg] \bigg)  \nonumber \\
		%=& \mathcal{O}\left(\frac{1}{T^{1 - \sigma_1}}\right) + \mathcal{O}\left(\frac{1}{T^{\sigma_1}}\right) + \mathcal{O}\left(\frac{1}{T^{1 - \sigma_2}}\right) + \mathcal{O}\left(\frac{1}{T^{\sigma_2}}\right)  + \mathcal{O}\left(\frac{1}{T^{2\sigma_1 - \sigma_2}}\right) + \mathcal{O}\left(\frac{1}{T^{2(\sigma_1 - \sigma_2)}}\right)   \nonumber \\
		%&  + \mathcal{O}\left(\frac{1}{T^{2\sigma_1 - \sigma_2 -1}}\right) + \mathcal{O}\left(\frac{1}{T^{2\sigma_1 - 2\sigma_2 - 1}}\right)  + \mathcal{O}\left(\epsilon_{app} \right). \label{rate: actor convergence 2}
		%\end{align}
	\end{proposition}
	
	The approximation error $\epsilon_{app}$ and sampling error $\epsilon_{sp}$ are defined in Appendix \ref{Discussion:Results}. A few remarks about the above results follow. First, one challenge in analyzing the convergence of Actor-Critic algorithms is that the actor and critic updates are typically sampled from different distributions (i.e., the distribution mismatch problem). To see this, note that to obtain an unbiased estimator for the policy gradient in
	\eqref{eq:pg_theorem}, one needs to sample from the discounted visitation measure $ d_{\btheta}(\cdot)$, while to obtain an unbiased estimator for the gradient of the MSBE in \eqref{eq:Decentralized_TD_evaluation} (which is utilized to update the critic parameters), one needs to sample from  the  stationary distribution $\mu_{\btheta}(\cdot)$. However, standard implementations for AC methods in practice  only use one sampling procedure for both actor and critic updates \cite{mnih2016asynchronous,shen2020asynchronous}.
Therefore, the mismatch between the two sampling distributions inevitably introduces constant biases, and this is where the error term $\epsilon_{sp}$ comes from.

Second, at each local agent $i$, the value function $V_{\pi_{\btheta}}(s)$ is approximated by $\phi(s)^T \omega_i$ and the global reward function is approximated by $\varphi(s, \ba)^T \lambda_i$. Due to the linear approximation, the approximation error is inevitable in the convergence analysis. Here, we use a constant term $\epsilon_{app}$ to quantify the approximation error due to utilizing linear function for policy evaluation.

By combining previous Propositions, and by properly selecting the stepsize parameters $\sigma_1$ and $\sigma_2$, we show the main result as below. In Appendix \ref{Discussion:Results}, we will present more discussion about a special case where there is no policy parameter sharing.
%(again note, that Assumptions \ref{ass: Lipschitz policy} -- \ref{ass: Markov chain} can be found in Appendix \ref{Appendix:Assumptions}):

% Hence, through selecting the diminishing order of the stepsizes as $ \sigma_1 = \frac{4}{5} $ and $ \sigma_2 = \frac{1}{5} $, we obtain the convergence rate of the \aname ~as below.

\begin{algorithm}[t] 
		\caption{{\small {\it Double Sampling Procedures}}}% $(\bW^{(0)},L_{\max},L,\rho,\beta,\delta,\Delta f)$}
		\begin{algorithmic}	
			\State {\bfseries Input:} Parameters $ \{ \omega_{i,t} \}_{i = 1}^{N} $, $ \{ \lambda_{i,t} \}_{i = 1}^{N} $, $ \{ \theta_{i,t} \}_{i = 1}^{N} $. 
			\State \textbf{Double i.i.d. Sampling}:
			\State 1) Sample $ s_t \sim \mu_{\btheta_{t}}(\cdot) $, $ \ba_t := \{a_{i,t} \sim \pi_i(\cdot| s_t, \theta_{i,t})\}_{i=1}^N $, $ s_{t+1} \sim \mathcal{P}(\cdot| s_t, \ba_t) $
			\State 2) Sample $ \tilde{s}_t \sim d_{\btheta_{t}}(\cdot) $, $ \tilde{\ba}_t := \{\tilde{a}_{i,t} \sim \pi_i(\cdot| \tilde{s}_t, \theta_{i,t})\}_{i=1}^N $, $ \tilde{s}_{t+1} \sim \mathcal{P}(\cdot| \tilde{s}_t, \tilde{\ba}_t) $
		\end{algorithmic}
		\label{alg: double sampling}
		%\vspace{-2px}
	\end{algorithm}

	\begin{theorem}\label{thm:cac}
		(Convergence of the \aname~ Algorithm) Suppose Assumptions \ref{ass:connectivity} - \ref{ass: Markov chain} hold. Consider Algorithm \ref{proposed algorithm} with partially shared policy parameters  $ \btheta := \cup_{i=1}^N \{ \theta^s_i, \theta^p_i \} $. Let $ \sigma_1 = \frac{3}{5} $ and $ \sigma_2 = \frac{2}{5} $, it holds that: 
		\begin{align}
		&\frac{1}{T} \sum_{t = 0}^{T-1} \sum_{i = 1}^{N} \bigg( \mathbb{E} \bigg[ \| \omega_{i,t} - \omega^*(\btheta_t) \|^2 \bigg] + \mathbb{E} \bigg[ \| \lambda_{i,t} - \lambda^*(\btheta_t) \|^2 \bigg] \bigg) = \mathcal{O}(T^{-\frac{2}{5}}), \nonumber \\
		&\frac{1}{T} \sum_{t = 0}^{T-1} \bigg(\mathbb{E} \bigg[ \| Q\cdot \btheta_t^s \|^2 \bigg] + N \cdot \mathbb{E} \bigg[ \| \overline{\nabla_{\btheta^s} J(\btheta_{t})}   \|^2 \bigg] + \sum_{i = 1}^{N} \mathbb{E} \bigg[ \| \nabla_{\theta_i^p} J( \btheta_{t}) \|^2 \bigg] \bigg)  = \mathcal{O}(T^{-\frac{2}{5}}) + \mathcal{O}(\epsilon_{app} + \epsilon_{sp}). \nonumber
		\end{align}
	\end{theorem}

% Moreover, we discuss the possibility of removing the sampling error $\epsilon_{sp}$. 
As mentioned before, the sampling error $\epsilon_{sp}$ arises because there is a mismatch between the way that estimators of the actor's and the critics' updates are obtained. To remove the sampling error, one can  implement separate sampling protocols for the critic and the actor. More specifically, we can use two different i.i.d. samples %for updating actor parameters and critic parameters respectively in Algorithm \ref{proposed algorithm}, the system will collect two samples 
 at each iteration step $t$: 1) $x_t := (s_t, \ba_t, s_{t+1})$ where $s_t \sim \mu_{\btheta}(\cdot)$, $ \ba_t \sim \pi_{\btheta}(\cdot \mid s_t)$ and $ s_{t+1} \sim \mathcal{P}(\cdot \mid s_t, \ba_t) $; 2) $\tilde{x}_t := (\tilde{s}_t, \tilde{\ba}_t, \tilde{s}_{t+1})$ where $\tilde{s}_t \sim d_{\btheta}(\cdot)$, $\tilde{\ba}_t \sim \pi_{\btheta}(\cdot \mid s_t)$ and $ \tilde{s}_{t+1} \sim \mathcal{P}(\cdot \mid \tilde{s}_t, \tilde{\ba}_t) $;  see Algorithm \ref{alg: double sampling}. Then $x_t$ and $\tilde{x}_t$ will be utilized in policy evaluation and policy optimization, respectively. %With two separate sampling procedures, the AC methods could utilize $x_t$ to calculate stochastic gradient estimates in critic step and utilize $\tilde{x}_t$ to perform actor step respectively. Therefore, the sampling error $ \epsilon_{sp} $ could be avoided due to the two different sampling procedures. 
 The corollary below shows the convergence result for the modified \aname~algorithm. Please see Appendix \ref{double sampling analysis} for the proof.
	
		\begin{Corollary}\label{cor:double_sample_cac}
		(Convergence under double sampling) %Considering each agent partially share its policy parameters with neighbors, we can denote $ \theta := \cup_{i=1}^N \{ \theta^s_i, \theta^l_i \} $. When double sampling procedures are implemented as Algorithm \ref{alg: double sampling} and selecting $ \sigma_1 = \frac{4}{5} $ and $ \sigma_2 = \frac{1}{5} $ to optimize the convergence rate
		Under the same setting as Theorem \ref{thm:cac}, consider \aname~with the double sampling procedures in Algorithm \ref{alg: double sampling}. The following result holds: 
		\begin{align}
		&\frac{1}{T} \sum_{t = 0}^{T-1} \sum_{i = 1}^{N} \bigg( \mathbb{E} \bigg[ \| \omega_{i,t} - \omega^*(\btheta_t) \|^2 \bigg] + \mathbb{E} \bigg[ \| \lambda_{i,t} - \lambda^*(\btheta_t) \|^2 \bigg] \bigg) = \mathcal{O}(T^{-\frac{2}{5}}), \nonumber \\
		&\frac{1}{T} \sum_{t = 0}^{T-1} \bigg(\mathbb{E} \bigg[ \| Q\cdot  \btheta_t^s \|^2 \bigg] + N \cdot \mathbb{E} \bigg[ \| \overline{\nabla_{\btheta^s} J(\btheta_{t})}   \|^2 \bigg] + \sum_{i = 1}^{N} \mathbb{E} \bigg[ \| \nabla_{\theta_i^p} J( \btheta_{t}) \|^2 \bigg] \bigg)  = \mathcal{O}(T^{-\frac{2}{5}}) + \mathcal{O}(\epsilon_{app}). \nonumber
		\end{align}
	\end{Corollary}

	\section{Numerical Results}
	 In this section, we present our simulation results on two environments: 1) the coordination game \cite{osborne1994course}; 2) the pursuit-evasion game \cite{gupta2017cooperative}, which is built on the PettingZoo platform \cite{terry2020pettingzoo}. Detailed experiment settings are present in Appendix \ref{appendix:exper:navigation}.
	
\begin{figure}[t]
 \centering
 \includegraphics[width=.23\linewidth]{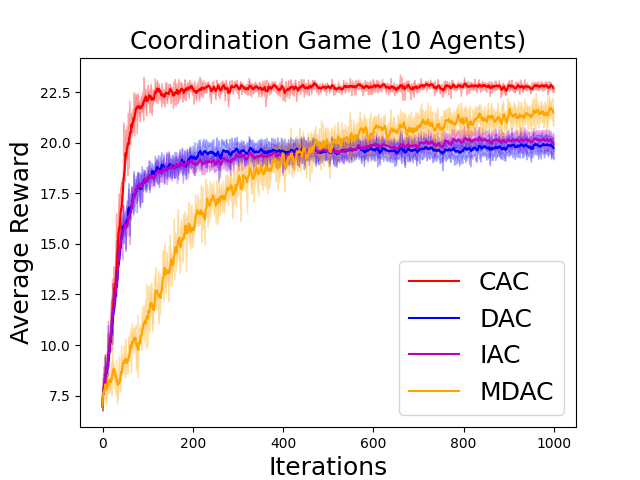} \quad
 \includegraphics[width=.23\linewidth]{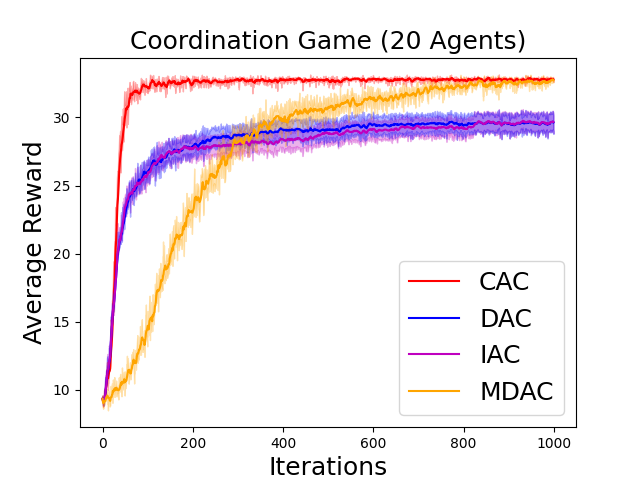} \quad
  \includegraphics[width=.23\linewidth]{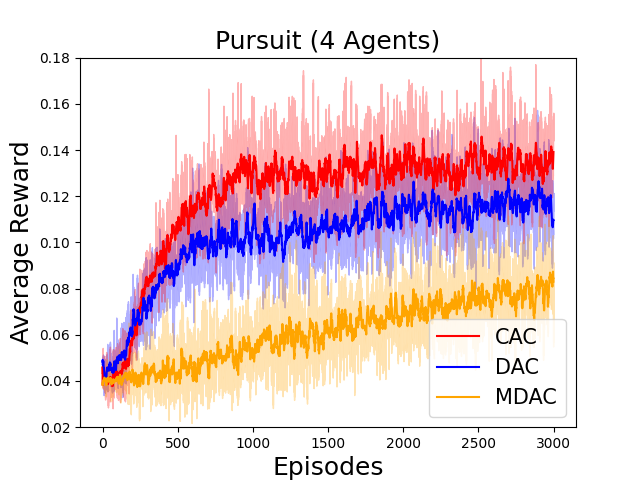}
  \includegraphics[width=.23\linewidth]{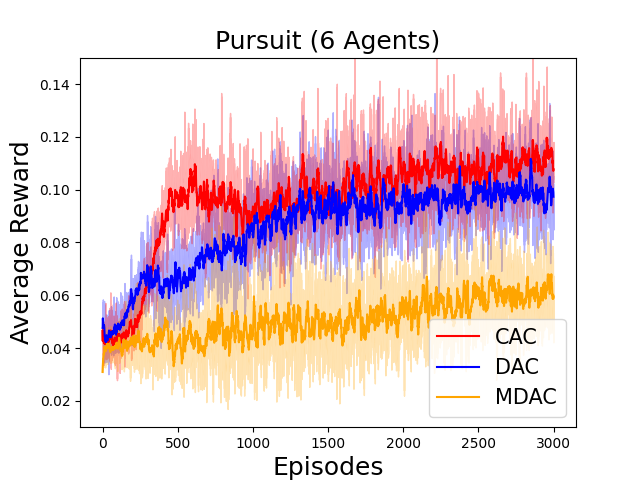}
\caption{ \tb{Simulation Results.} The averaged reward versus the learning process. The performance is averaged over $10$ Monte Carlo runs.}
\label{fig:exper}  
\end{figure}

	\textbf{Coordination Game}: In this setting, there are $N$ agents staying at a static state and they choose their actions simultaneously at each time. %The cardinality of their action space is $8$ and each agent choose their action among the available action set $\{ 0, 1, 2, \cdots, 7 \}$.
	After actions are executed at each time $t$, each agent $i$ receives its reward as: $r_{i,t} = (a_{i,t} - 3.5)^2 + \sum_{j \neq i} I_{\{a_{j,t} = a_{i,t}\}} + \epsilon_{i,t}$ where the action space is $\{ 0, 1, 2, \cdots, 7 \}$, $I_{\{a_{j,t} = a_{i,t}\}}$ is an indicator function and $ \epsilon_{i,t} $ is a random payoff following standard Gumbel distribution. In this coordination game, there are multiple Nash equilibria where two optimal equilibria are that all agents select $a = \{0\}$ or $a = \{ 7 \}$ simultaneously. In order to obtain high rewards and achieve efficient equilibria, it is crucial for agents to coordinate with others while only having limited communications. 
	Here, the communication graph $\mathcal{G}_t$ between the agents is a complete graph every $5$ iterations, and is not connected for the rest of time. We compare the performance of \aname ~with three benchmark algorithms: independent Actor-Critic (IAC); decentralized Actor-Critic (DAC) in \cite{zhang2018fully}; mini-batch decentralized Actor-Critic (MDAC) in \cite{chen2021sample}. For each algorithm, we set the actor stepsize and critic stepsize as $0.05$ and $0.1$. Theoretically, MDAC needs $\mathcal{O}(\epsilon^{-1} \ln{\epsilon^{-1}})$ batch size in its inner loop to update critic parameters before each update in policy parameters, which is inefficient in practice. Here, we set small batch $B = 5$ in the inner loop for MDAC to achieve fast convergence. The simulation results on this coordination game are present in Fig.\ref{fig:exper} (two left figures). According to the simulations, compared with the benchmarks, we see that the \aname ~algorithm converges faster  and has higher probability to achieve efficient equilibria due to the use of policy sharing and coordination.
	
	\textbf{Pursuit-Evasion Game}: there are two groups of nodes, pursuers (agents) and evaders.  
	The pursuers aim to obtain reward through catching evaders. In a two-dimensional environment, an evader is considered caught if two pursuers simultaneously arrive at the evader's location. In order to catch an evader, each pursuer should learn to cooperate with other pursuers to catch the evaders. From this perspective, the pursuers share some similarities with each other since they need to follow similar strategies to achieve their local tasks: simultaneously catching a same evader with other pursuers.  
	In Figure \ref{fig:exper} (two right figures), we compare the numerical performance of the proposed \aname ~algorithm and two benchmarks: decentralized Actor-Critic (DAC) in \cite{zhang2018fully}; mini-batch decentralized Actor-Critic (MDAC) in \cite{chen2021sample}.
	Each agent maintains two convolutional neural networks (CNNs), one for the actor and one for the critic. Please see Figure \ref{Exper:Architecture:NN} in Appendix for the structure diagrams of actor network and critic network being used. In the \aname, two convolutional layers of actor network will be regarded as shared policy parameters, and the output layer is personalized (thus not shared). %For each algorithm, we set the actor stepsize and critic stepsize as $1 \times 10^{-4}$ and $1 \times 10^{-3}$. For algorithm MDAC to achieve quick convergence, we tune its batch size and set small batch $B = 5$ in its inner loop.

The two sets of numerical results suggest that, when local tasks share a certain degree of similarity / homogeneity, \aname ~algorithm with (partial) parameter sharing could  achieve more stable convergence. 

\section{Conclusion}

This paper develops a novel collaboration mechanism for designing robust MARL systems. Further, it 
develops and analyzes a novel multi-agent AC method, where agents are allowed to (partially) share their policy parameters with
the neighbors to learn from different agents. To our knowledge, this is the first
non-asymptotic convergence result for two-timescale multi-agent AC methods.

%\acks{We thank a bunch of people.}

	\bibliography{decentralized_RL}
    \bibliographystyle{IEEEtran}

% Acknowledgments---Will not appear in anonymized version

\newpage
\appendix

\section{Experiment Details}
\label{appendix:exper:navigation}

	\begin{figure}[t]
 \centering
 \includegraphics[width=.48\linewidth]{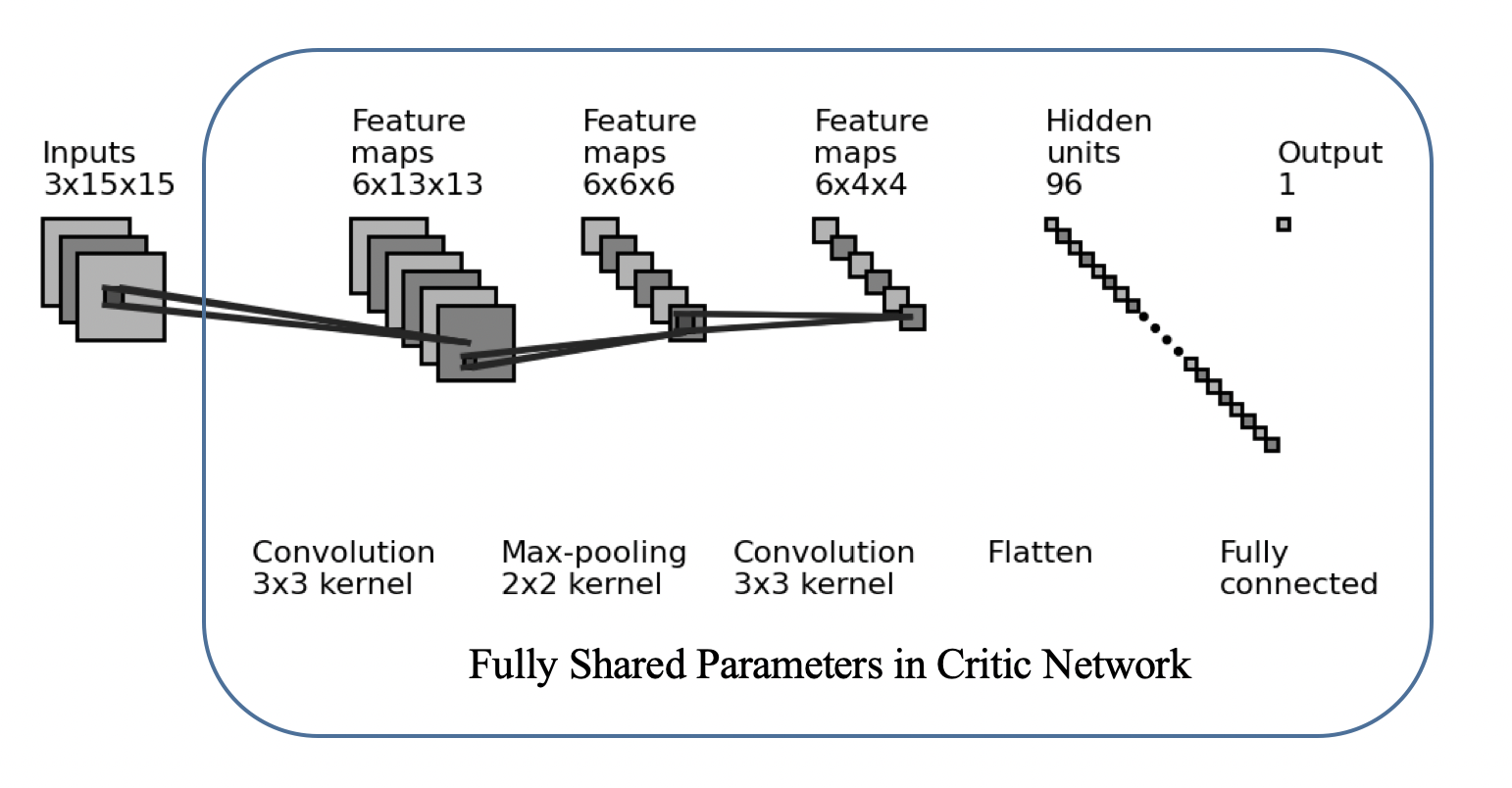} \quad
  \includegraphics[width=.48\linewidth]{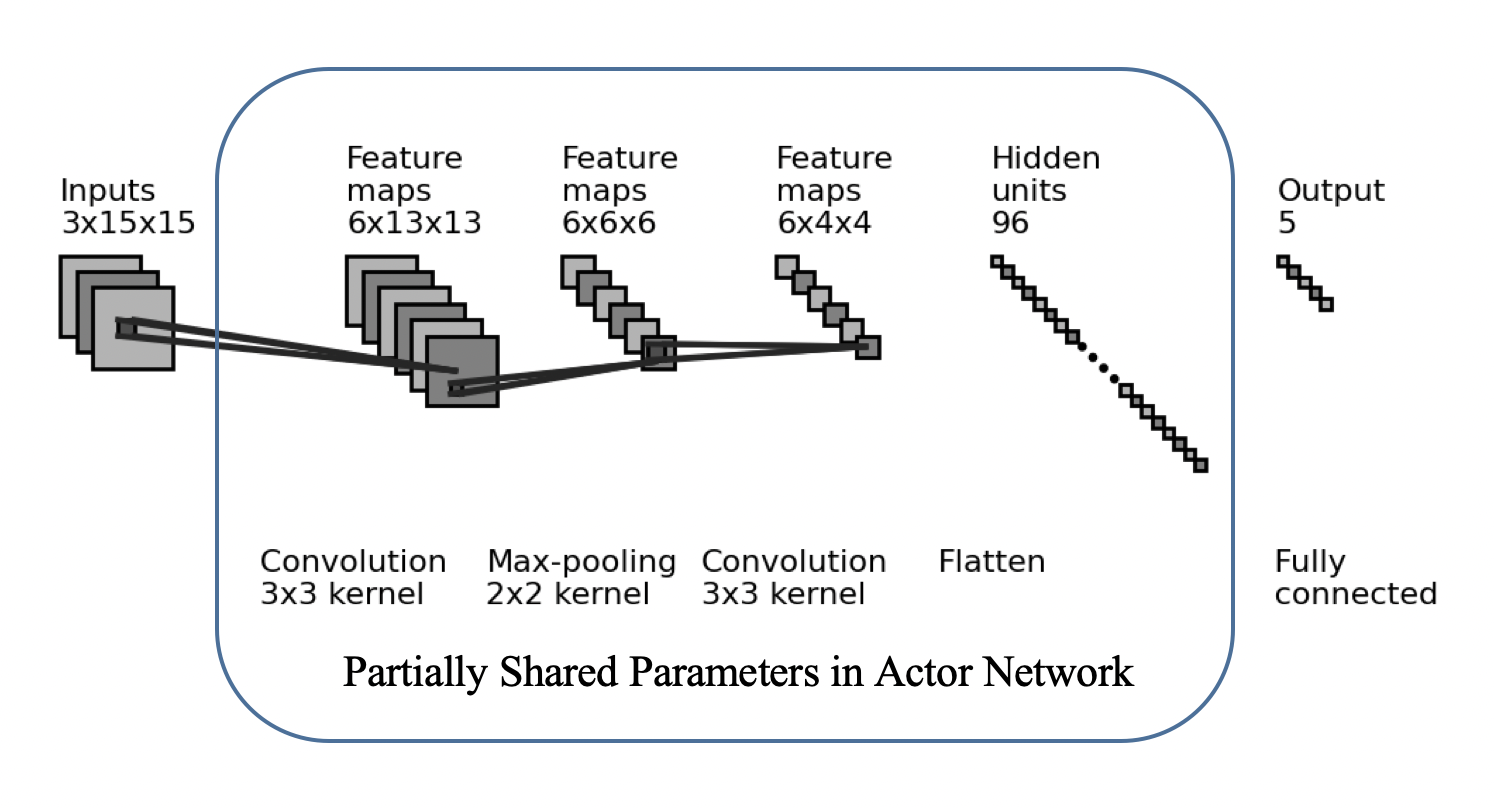}
\caption{\tb{Neural Network Architecture Diagrams for the CAC Algorithm.} The architecture diagrams for actor network and critic network of algorithm \aname ~in the pursuit-evasion Game. (Left) The diagram of critic network. (Right) The diagram of actor network.} \label{Exper:Architecture:NN}  
\end{figure} 

In this section, we will present the experiment details on the pursuit-evasion game.
\subsection{pursuit-evasion Game}
	The `capture' reward for each agent is set to be $5$ when a pursuer successfully catches an evader. Moreover, the pursuer will receive a small reward signal which is set to be $0.1$ when the pursuer encounters an evader at its current location. The environment is set to be a $15 \times 15$ grid and this 2D grid contains obstacles where the agents cannot pass through. Hence, the global state of the pursuit-evasion game consists of three images (binary matrices) of the size of $15 \times 15$. Hence, the dimension of the global state is $3 \times 15 \times 15$. These three images (binary matrices) respectively present the location of the pursuers, evaders and obstacles in the two-dimensional grid. Only given the 3-channel images as the global state, it is difficult for each pursuer (agent) to distinguish itself with other pursuers since the 3-channel images (global state) does not directly show the ID for each pursuer in the pursuit-evasion Game. To tackle this challenging, we center each agent's observation at its own location. With a large observation radius, each agent could observe the global information in the environment. 
	
	Considering the observation of each agent is a 3-channel image, each agent respectively maintains two convolutional neural networks (CNNs) with two convolutional layers, one max-pooling layer and one fully connected layer for the actor and the critic. Please see Figure \ref{Exper:Architecture:NN} for the structure diagrams of actor network and critic network in algorithm \aname. The communication graph $\mathcal{G}_t$ between the agents is a complete graph every $20$ iterations, and is not connected for the rest of time. Hence, for \aname ~algorithm, the
	global averaging step will be performed on the entire critic networks and the two CNN layers of actor networks every $20 $ iterations. The RuLU activation function is utilized in each hidden layer of actor network and critic network. The output of critic network approximates the value function $V_{\pi_{\btheta}}(s)$ for all $s \in \mathcal{S}$ and the dimension of the output layer is $1$. Furthermore, the output dimension of actor network is $5$ which corresponds to the number of possible actions. In each CNN, the raw images (3-channel location matrices), whose dimension is $3 \times 15 \times 15$, are processed by two convolutional layers and one max-pooling layer first and then pass through a fully connected layer as the output layer. We utilize the RMSprop optimizer \cite{ruder2016overview} to train neural networks, which is a common choice in training neural networks for reinforcement learning problems \cite{mnih2013playing}.
	For each algorithm, we set the actor stepsize and critic stepsize as $1 \times 10^{-4}$ and $1 \times 10^{-3}$. For algorithm MDAC to achieve quick convergence, we tune its batch size and set small batch $B = 5$ in its inner loop. The discount factor $\gamma$ is set to be $0.95$ in this simulation.

\section{Discussion: Application Scenarios and Potential Benefits}

Here, we discuss how the proposed \aname ~algorithm can be used in two popular multi-agent settings: 

$\bullet$	 {\bf Fully personalized multi-agent RL.} The \aname ~algorithm can be applied to the special setting where the agents {\it do not} share their policy parameters with neighbors, and no cooperation is considered in generating the local policies. This setting has been proposed and studied in a number of existing works, such as  \cite{zhang2018fully,chen2018communication}. 

$\bullet$	 {\bf Federated RL.} The proposed algorithm can be applied to a general federated (reinforcement) learning setting, where the agents jointly optimize a common objective. 
    To see the connection, let us first describe a standard federated learning (FL) setting \cite{konevcny2016federated,arivazhagan2019federated}: a central controller coordinates a few agents, where the agents 
    continuously optimize their local parameters and perform occasional averaging steps over the parameters. It can be shown that this protocol corresponds to a dynamic setting where $\mathcal{G}_t$ is a  {\it complete} graph every fixed number of iterations, while it is {\it not connected at all} for the rest of times (which is a special case of our setting, see Assumption \ref{ass:connectivity}). When the network is connected, each agent could gather other agents' models and perform the averaging; when the network is not connected, then each agent just performs local updates. We generalize the above FL setting to MARL in \aname ~algorithm.  

 Although our partially personalized policy structure may be relatively more difficult to analyze, it has a number of potential advantages, as we list below:

	$ \bullet $ \emph{\bf A Generic Model}. We use the partial policy sharing as a generic setting, to cover the full spectrum of strategies ranging from no sharing case ($\theta^s_i=0,\;\forall~i$) to the full sharing case (($\theta^p_i=0,\;\forall~i$)). This generic model ensures that our subsequent algorithms and analysis can be directly used for all cases. 
	
	$ \bullet $
	\emph{\bf Better Models for Homogeneous Agents}. When the agents' local tasks have a high level of similarity (a.k.a. the {\it homogeneous} setting), partially sharing models' parameters could achieve better feature representation and guarantee that
	    the agents' policies are closely related to each other. Additionally, the shared parameters could leverage more data
(i.e., data drawn from all agents) compared with the personalized parameters, so the variance in the training process can be significantly reduced, potentially resulting in better training performance.
	  Such an intuition has been verified empirically in reinforcement learning systems \cite{omidshafiei2017deep,yu2020meta,zeng2020decentralized}, where sharing policies among different learners results in more stable convergence. 
	   
	 $ \bullet $ \emph{\bf Approximate Common Knowledge.} A critical assumption often made in the analysis of multiagent systems is {\em common knowledge} \cite{Aumann}. Intuitively, this implies agents have a {\em shared} awareness of the underlying interaction.
	A key difficulty in MARL  is that agents are simultaneously learning features of the underlying environment, thus common knowledge is not guaranteed.  
	Thus notions of {\em approximate} common knowledge have been proposed for MARL \cite{Schroeder}. By relying on (partial) policy sharing mechanism, we hope to have some degree of approximate common knowledge and this is what facilitates coordination.

\section{Technical Assumptions} \label{Appendix:Assumptions}

	\begin{assumption} \label{ass: Lipschitz policy}
		 Define the score function $ \psi_{\btheta}(s, \ba) := \nabla_{\btheta} \log \pi_{\btheta}(\ba \mid s) $. For any policy parameters $ \btheta$ and $\btheta^\prime$, and any state-action $(s,\ba)$, the following holds:
		\begin{subequations}
					\begin{align}\label{Lipschitz: policy}
			\| \psi_{\btheta}(s, \ba)  - \psi_{\btheta^\prime}(s, \ba)  \| & \leq L_{\psi} \cdot \| \btheta - \btheta^\prime \|,\quad  \| \psi_{\btheta}(s, \ba) \|  \leq C_{\psi}\quad  \\
			| \pi_{\btheta}(\ba \mid s) - \pi_{\btheta^\prime}(\ba \mid s) | & \leq L_{\pi} \cdot \| \btheta - \btheta^\prime \|
			\end{align}
		\end{subequations}
		where $L_{\pi}, L_{\psi}, C_{\psi}$ are some  constants.
	\end{assumption}
	Assumption \ref{ass: Lipschitz policy} has been often used in analyzing policy gradient-type algorithms, for example see \cite{zhang2020global,agarwal2020optimality}. Many policy parameterization methods such as tabular softmax policy \cite{agarwal2020optimality}, Gaussian policy \cite{doya2000reinforcement} and Boltzmann policy \cite{konda1999actor} satisfy this assumption. %{\red[comment on other assumptions]}
	
	\begin{assumption} \label{ass: Markov chain}
		For any policy parameters $ \btheta $, the markov chain under policy $ \pi_{\btheta} $ and transition kernel $ \mathcal{P}(\cdot| s, \ba) $ is irreducible and aperiodic. Then there exist constants $ \kappa > 0 $ and $ \tau \in (0, 1) $ such that 
		\begin{align}
		\sup_{s \in \mathcal{S}} ~d_{TV}\big(\mathcal{P}^{\pi_{\btheta}}(s_t \in \cdot| s_0 = s), \mu_{\btheta}(\cdot)\big) \leq \kappa \cdot \tau^t, \quad \forall t   \label{mixing rate: markov chain}
		\end{align}
		where $ d_{TV}(\cdot) $ is the total variation (TV) norm; $ \mu_{\theta} $ is the stationary state distribution under $ \pi_{\btheta} $.
	\end{assumption}
	Assumption \ref{ass: Markov chain} assumes the Markov chain mixes at a geometric rate; see also \cite{bhandari2018finite,sun2018markov}.

\section{Auxiliary Lemmas} \label{appendix:auxiliary_lemmas}
	
	\begin{lemma} \label{lemma: projection}
		(\cite[Lemma 1]{nedic2010constrained}) Let $ \mathcal{X} $ be a nonempty closed convex set in $ \mathbb{R}^K $, then the following holds: 	
		\begin{subequations}
		   \begin{align} 
		&\left( \Pi_{\mathcal{X}} [x] - x \right)^T (x - y) \leq - \| \Pi_{\mathcal{X}}[x] - x \|^2, \; \forall~  x \in \mathbb{R}^K, \; y \in \mathcal{X} \label{bound: nedic 1} \\
		&\| \Pi_{\mathcal{X}} [x] - y \|^2 \leq \| x-y \|^2 - \| \Pi_{\mathcal{X}} [x] - x \|^2, \; \forall~x \in \mathbb{R}^K, \; y \in \mathcal{X} \label{bound: nedic 2}
		\end{align}
		\end{subequations}
				where $\Pi_{\cal X}[\cdot]$ denotes the projection operator on to the convex set $\mathcal{X}$.
	\end{lemma}
	
	\begin{lemma} \label{lemma: Lipschitz Objective}
       (\cite[Lemma 3.2]{zhang2020global}). Suppose Assumption \ref{ass: Lipschitz policy}  holds. Then the following holds:
      \begin{align}
          \| \nabla J(\btheta) - \nabla J(\btheta^\prime) \| \leq L_J \cdot \| \btheta - \btheta^\prime \|, \quad \forall~\btheta, \btheta^\prime \in \mathbb{R}^{N \times D} \label{eq: lipschitz objective}
      \end{align}
      where $J(\cdot)$ is the objective function defined in  \eqref{def: marl objective}; $L_J := \frac{R_{\max} \cdot L_{\psi}}{(1-\gamma)^2} + \frac{(1+\gamma) \cdot R_{\max} \cdot  C_{\psi}^2}{(1-\gamma)^3}$ with $L_{\psi}$ and $C_{\psi}$ defined in Assumption \ref{ass: Lipschitz policy}.
      
	\end{lemma}

	\begin{lemma} \label{lemma: Lipschitz Value Function}
	 (\cite[Lemma 4]{shen2020asynchronous}). Suppose Assumption \ref{ass: Lipschitz policy} holds. The following holds: %we have
	\begin{align}
	    \| \nabla J(\btheta) \| \leq& L_v,  \quad 
	    |J(\btheta_1) - J(\btheta_2)| \leq L_v \cdot \| \btheta_1 - \btheta_2 \|, \; \forall~ \btheta, \btheta^\prime \in \mathbb{R}^{N \times D}, \; s \in \mathcal{S}, \label{eq:Lipschitz:objective}
	\end{align}
	where the constant $ L_v := \frac{R_{\max}}{1 - \gamma}\cdot  C_{\psi} $, with $ C_{\psi} $ defined in Assumption \ref{ass: Lipschitz policy}.
	
	\end{lemma}
	
	\begin{lemma} \label{lemma:sampling_mismatch}
	(\cite[{Lemma 1}]{shen2020asynchronous}) Suppose Assumption \ref{ass: Markov chain} holds. The following holds:
	\begin{align}
	    d_{TV}(\mu_{\btheta}, d_{\btheta}) \leq 2\left( \log_{\tau} \kappa^{-1} + \frac{1}{1 - \tau} \right)(1-\gamma), \quad \forall~\btheta \in \mathbb{R}^{N \times D} \label{eq:sampling_mismatch}
	\end{align}
	where $ \mu_{\btheta}(\cdot) $ is the stationary distribution of each state $s$ under policy parameters $\btheta$ and $d_{\btheta}(\cdot)$ is the discounted visitation meausre $ d_{\btheta}(s) := (1-\gamma)\sum_{t=0}^\infty \gamma^t \cdot  \mathcal{P}^{\pi_{\btheta}} (s_t = s\mid s_0 \sim \eta)$.
	\end{lemma}

With the technical assumptions in \ref{Appendix:Assumptions}, we could bound consensus errors over the iterations. Towards this end, let us provide some basic properties for the weight matrices. Based on Assumptions \ref{ass:connectivity} - \ref{ass: mixing_matrix} which ensure the long-term connectivity and impose the underlying topology of the networked system, 
 we can obtain the following condition \cite[Lemma 9]{nedic2009distributed}:
	\begin{align}
	\| W_t \cdots W_{t + B - 1} \cdot Q \cdot \bomega \| \leq \eta \cdot \|  Q \cdot \bomega \|, \label{bound: graph connectivity}
	\end{align}
when $ \bomega := [\omega_{1}^T; \omega_{2}^T; \cdots;\omega_{N}^T] \in \mathbb{R}^{N \times K} $; and we define $ Q := I - \frac{1}{N}\bm 1 \cdot \bm 1^T \in \mathbb{R}^{N \times N}. $
Further, the constant $\eta$ in \eqref{bound: graph connectivity} is given by $ \eta: = \sqrt{1 - \frac{c}{2N^2}} \in (0,1), $
where constant $c$ is defined in Assumption \ref{ass: mixing_matrix}. 

Based on the above property, we have the following bounds on various consensus errors. Please see Appendix \ref{appendix:consensus_error} for the proof. %{\red Please see Appendix xxx for the proof.}
	\begin{lemma} \label{lemma: consensus error}
		 Based on Assumptions \ref{ass:connectivity} - \ref{ass:feature_bound}, there exist constants $L_B > 0$ and $ \ell_p > 0 $ such that the consensus errors $ \| Q \cdot \bomega_t \| + \| Q \cdot \blambda_t \| $ and $\| Q \cdot \btheta_t^s \|$ satisfy
		 \begin{subequations}
		 			\begin{align}
		 	&\| Q \cdot \bomega_t \| + \| Q \cdot \blambda_t \| \leq \rho^{t} \cdot \frac{\| \bomega_0 \| + \| \blambda_0 \|}{\eta}  + \frac{2N \cdot L_b \cdot (\beta+\zeta)}{\eta \cdot (1 - \rho)} \cdot \frac{1}{ T^{\sigma_2}} \label{lemma: consensus bound} \\
		 	&\| Q \cdot \btheta_t^s \| \leq \rho^{t} \cdot \frac{\| \btheta_{0}^s \|}{\eta} + \frac{\ell_p \cdot \alpha}{\eta \cdot (1-\rho)} \cdot \frac{1}{T^{\sigma_1}} \label{lemma: policy consensus error bound}\\
		 	&\frac{1}{T} \sum_{t = 0}^{T-1} \bigg( \| Q\cdot  \bomega_{t} \| + \| Q\cdot  \blambda_{t} \| \bigg) = \mathcal{O}\bigg( T^{-\sigma_2} \bigg), \label{bound:lower:consensus_errors} \\ 
	    &\frac{1}{T} \sum_{t = 0}^{T-1} \bigg( \| Q \cdot \blambda_{t} \| ^2 + \| Q \cdot \bomega_{t} \| ^2 \bigg)
		= \mathcal{O}(T^{-1}) + \mathcal{O}(T^{-2\sigma_2}) \label{rate:lower:critic:consensus:square} \\
		 	& \frac{1}{T} \sum_{t = 0}^{T-1} \| Q \cdot \btheta_t^s \|  = \mathcal{O} \bigg( T^{-\sigma_1} \bigg), ~ \frac{1}{T} \sum_{t = 0}^{T-1} \| Q \cdot \btheta_t^s \|^2  = \mathcal{O}(T^{ - 1}) + \mathcal{O}(T^{ - 2\sigma_1})
		 	\end{align}
		 \end{subequations}
		where $\alpha_t : = \frac{\alpha}{T^{\sigma_1}} $, $\beta_t : = \frac{\beta}{T^{\sigma_2}} $ and $\zeta_t : = \frac{\zeta}{T^{\sigma_2}} $ are three stepsizes; $ \rho := \eta^{\frac{1}{\mathcal{B}}}$.
		%Moreover, we could obtain the convergence rate for the consensus errors as $ \frac{1}{T} \sum_{t = 0}^{T-1} \| Q \omega_{t} \| = \mathcal{O}\bigg( T^{-\sigma_2} \bigg) $ and $ \frac{1}{T} \sum_{t = 0}^{T-1} \| Q \theta_t^s \| = \mathcal{O}( T^{-\sigma_1} ) $.
	\end{lemma}

		Given the fixed policy parameter $\btheta$, solving the lower level problem of \eqref{eq:bi-level}  is equivalent to solving the {\it centralized} policy evaluation problems, expressed in \eqref{eq:decentralized_reward_evaluation} and  \eqref{eq:Decentralized_TD_evaluation}.
Through the first-order optimality condition, it is easy to show that $w^*(\btheta)$ satisfies the following condition: 
% we could 
	%such that $ A \cdot \omega^*  = \frac{1}{N} \sum_{i = 1}^N b_i  $.
\begin{align}
	    A(\btheta) \cdot \omega^*(\btheta) = \frac{1}{N} \sum_{i=1}^N b_{i}(\btheta) \label{eq:TD_optimal_condition}
	\end{align}
where we have defined:
    \begin{subequations}
    \label{eq: TD fixed point}
       \begin{align}
	    A(\btheta)  :=& ~ \mathbb{E}_{s \sim \mu_{\btheta}(\cdot), s^\prime \sim \mathcal{P}^{\pi_{\btheta}}(\cdot|s) } \left[ \phi(s) \big(\phi(s) - \gamma \cdot \phi(s^\prime) \big)^T \right], \; \forall~i \in \mathcal{N}, \\
	    b_{i}(\btheta) :=& ~ \mathbb{E}_{s \sim \mu_{\btheta}(\cdot), \ba \sim \pi_{\btheta}(\cdot|s) } \left[ r_i(s,a) \cdot \phi(s) \right], \; \forall~i \in \mathcal{N}.
	\end{align} 
    \end{subequations}
Under the full-rankness and bounded assumption of the feature matrices given in Assumption \ref{ass:feature_bound},  %it ensures that feature vectors ($\phi(s)$ and $\varphi(s, a)$)are bounded and feature matrices ($ \Phi $ and $\Psi$) have full rank. By applying 
we can apply \cite[Theorem 2]{tsitsiklis1997analysis}, and show that $A(\btheta)$ is a positive definite matrix for any fixed $\btheta$. Let us define 
\begin{align}
\label{def:eigenvalue_A}
  0 < \tilde{c}_{\text{min}} \leq  c_{\min}\big(A(\btheta)\big), \quad  0 < c_{\max} \big(A(\btheta)\big) \leq \tilde{c}_{\text{max}}, ~\forall ~ \btheta \in \mathbb{R}^{N \times D}, 
\end{align}
where $c_{\min}\big(A(\btheta)\big)$ and $ c_{\max} \big(A(\btheta)\big) $ are the minimum and maximum eigenvalue of $A(\btheta)$; $\tilde{c}_{\text{min}}$ and $\tilde{c}_{\text{max}}$ are the lower bound and upper bound on the eigenvalues of $A(\btheta)$.
%whose minimum eigenvalue is lower bounded by $ \tilde{c}_{\text{min}} $ its maximum eigenvalue is upper bounded by $ \tilde{c}_{\text{max}} $; the optimal solution in \eqref{eq:value_func_approx_sol} satisfies that
% Moreover, under fixed policy parameters $ \btheta $, the unbiased critic update for any critic parameters $\omega_t$ could be expressed as below,
% 	\begin{align}
% 	    \label{eq:critic_unbiased_update}
% 	    \omega_{t+1} := \omega_t + \beta_t \cdot \big( b_{\btheta} - A_{\btheta} \cdot \omega_t \big)
% 	\end{align}
% 	until $\omega_{t}$ satisfies the optimal condition \eqref{eq:TD_optimal_condition}.
% 	}
	Then we have the following Lipschitz property of the optimal critic parameters. 
	\begin{lemma}(\cite[Proposition 2 ]{shen2020asynchronous}) \label{lemma: critic Lipschitz}
		Suppose Assumptions \ref{ass:feature_bound},\ref{ass: Lipschitz policy},\ref{ass: Markov chain} hold. Let $w^*(\btheta)$ denote the optimal solution in \eqref{eq:Decentralized_TD_evaluation} to approximate value function approximation. Then the following Lipschitz condition holds:
		\begin{align}
		\| \omega^*(\btheta_1) - \omega^*(\btheta_2) \| \leq L_\omega \cdot \| \btheta_1 - \btheta_2 \|, \quad \forall~ \btheta_1,  \btheta_2 \in \mathbb{R}^{N \times D}, \label{eq: lipschitz_critic}
		\end{align}
		where  $ L_\omega := 2\cdot R_{\max} \cdot |\mathcal{A}| \cdot L_\pi \cdot \bigg(\tilde{c}_{\min}^{-1} + \tilde{c}_{\min}^{-2} \cdot (1 + \gamma) \bigg) \bigg( 1 + \log_{\tau} \kappa^{-1} + (1-\tau)^{-1} \bigg) $. %Moreover, $c_{\min}(A)$ and $c_{\max}(A)$ denote the minimum and maximum eigenvalue of matrix $A$ in the equation \eqref{eq: TD fixed point}.
	\end{lemma}
	%\begin{proof}
		%Here, we denote $ b_{\pi_{\theta_1}} := \frac{1}{N} \sum_{i = 1}^{N} b_i(\theta_1) $ and $ b_{\pi_{\theta_2}} := \frac{1}{N} \sum_{i = 1}^{N} b_i(\theta_2) $. Then we follow the proof of \cite[Proposition 2 ]{shen2020asynchronous} and we are able to show this Lipschitz continuous property. {\red[can we just directly cite this result?]}
	%\end{proof}
	
%	{\red[need to properly define the $w^*$ in the lemma.][I think it is very confusing between 27 and 28. it is not clear for what problem (27) is defined. It is for the separable problem where the agents do individual policy evaluation (without consensus), or for the centralized problem with a single parameter $w$?]}

	\iffalse
	In Lemma \ref{lemma: consensus error}, we analyze the convergence rate of the consensus error in critic and actor. %Further, we are able to construct the convergence analysis for this decentralized actor-critic algorithm with Lemma \ref{lemma: projection} - \ref{lemma: critic Lipschitz}. 
	In Lemma \ref{lemma: critic Lipschitz}, it is shown that critic's optimal parameter $\omega^*(\theta)$ is Lipschitz continuous on policy parameter $\theta$. This result provides us an important property for the convergence analysis in each agent's critic parameters. Moreover, to provide a unified finite-time analysis for the federated multi-agent AC method, we further analyze the convergence rate for each agent's policy parameters to reach stationary points in Theorem \ref{thm: critic} - \ref{thm: actor}.
	\fi
	
	\section{Discussion: Convergence Results}
\label{Discussion:Results}
	In this section, we discuss an extension of Theorem \ref{thm:cac}.
	
	As a special case, when the agents do not share any policy parameters, that is, when  $ \btheta = \btheta^p$, the resulting algorithm reduces to the standard  Decentralized AC algorithm and we name it as Coordinated Actor-Critic with no policy sharing (\bname), whose asymptotic convergence property has been analyzed in \cite{zhang2018fully}. %In \bname ~where each agent $i \in \mathcal{N}$ independently optimizes its policy parameter $\theta_i$, it is a special case of \aname ~and the policy parameter in \bname ~is defined as $ \btheta := \{ \theta_i^p \}_{i=1}^N $. 
	The non-asymptotic convergence rate for this algorithm can be readily obtained from (a slightly modified versions of) Proposition \ref{proposition:critic} -- \ref{proposition:actor}. The following result states the convergence rate for \bname, and we refer the readers to  Appendix \ref{appendix:actor_convergence} for detailed proof steps.
	
	%Hence, we are able to optimize the stepsizes of \bname ~to improve its convergence rate through selecting $ \sigma_1 = \frac{3}{5} $ and $ \sigma_2 = \frac{2}{5} $. Then, assuming separate sampling procedures are considered in actor step and critic step to avoid the sampling error, we obtain the corollary as below:} %In AC algorithm, the estimated stochastic gradient for actor updates is biased due to inexact value function approximation. The bias introduced by the critic optimality gap and the function approximation error correspond to the last term in equation (\ref{rate: actor convergence}).
	
%	\TC{It is not clear how we can obtain two different bounds from (27)? Maybe make it more explicit.}
	
	\begin{Corollary}\label{cor:dec-ac}
		(Convergence of \bname~Algorithm) Suppose Assumptions \ref{ass:connectivity} - \ref{ass: Markov chain} hold.  Consider applying Algorithm \ref{proposed algorithm} to a problem with  fully-personalized policy parameters, that is,  $ \btheta := \cup_{i=1}^N \{ \theta^p_i \} $. Setting $ \sigma_1 = \frac{3}{5} $ and $ \sigma_2 = \frac{2}{5} $, then the following holds:
		\begin{align}
		&\frac{1}{T} \sum_{t = 0}^{T-1} \sum_{i = 1}^{N} \bigg( \mathbb{E} \bigg[ \| \omega_{i,t} - \omega^*(\btheta_t) \|^2 \bigg] + \mathbb{E} \bigg[ \| \lambda_{i,t} - \lambda^*(\btheta_t) \|^2 \bigg] \bigg) = \mathcal{O}(T^{-\frac{2}{5}}) \label{exact rate 1: critic} \\
		&\frac{1}{T} \sum_{t = 0}^{T-1} \sum_{i = 1}^{N} \mathbb{E} \bigg[ \| \nabla_{\theta_i} J(\btheta_{t}) \|^2 \bigg] = \mathcal{O}(T^{-\frac{2}{5}}) + \mathcal{O}(\epsilon_{app} + \epsilon_{sp}) \label{exact rate 1: actor}.
		\end{align}
	\end{Corollary}
	
	The critic approximation error $ \epsilon_{app} $ and the sampling error $\epsilon_{sp}$ are defined as follows  \cite{wu2020finite,shen2020asynchronous}: %where the approximation error is inevitable due to the use of linear approximation.
	\begin{align}
	    \epsilon_{\text{app}} & := \max_{\btheta} \sqrt{ \mathop{\mathbb{E}} \limits_{ \substack{s \sim \mu_{\btheta}(\cdot) }} \left[ \bigg( V_{\pi_{\btheta}}(s) - \phi(s)^T \omega^*(\btheta) \bigg)^2 \right] } + \sqrt{ \mathop{\mathbb{E}} \limits_{ \substack{s \sim \mu_{\btheta}(\cdot) \\ a \sim \pi_{\btheta}(\cdot) }} \left[ \bigg( \bar{r}(s,\ba) - \varphi(s,\ba)^T\lambda^*(\btheta) \bigg)^2 \right] } \label{error:approximation} \\
	   \epsilon_{\text{sp}} & := 4\cdot R_{\max}\cdot  C_{\psi} \cdot L_v\cdot \left( \log_{\tau} \kappa^{-1} + \frac{1}{1 - \tau} \right) \label{error:sampling}
	\end{align}
	where $\mu_{\btheta}(\cdot)$ is the stationary distribution of state under policy $\pi_{\btheta}$ and the transition kernel $\mathcal{P}(\cdot)$.

	\section{Proof of Lemma \ref{lemma: consensus error}} 
	\label{appendix:consensus_error}
	
	The proof of Lemma \ref{lemma: consensus error} is divided into two steps. In step 1, we first analyze the consensus error $\| Q \cdot \bomega_t \|$ and then extend the analysis results to $\|Q \cdot \blambda_t \|$. In step 2, we further analyze the consensus error $\| Q \cdot \btheta_t^s \|$ in the shared part of policy parameters.
	
	\begin{proof}[{\bf Step 1.}] \label{proof: Lemma consensus bound}
		Since the mixing matrix $ W_t $ is doubly stochastic so $ W_t \cdot \bm 1 = \bm 1 $, where $\bm 1$ is a column vector of all ones. We obtain that 
		\begin{align}
		W_t \cdot \bomega_{t} - \bm 1 \cdot \bar{\omega}_t^T = W_t \cdot \left( \bomega_{t} - \bm 1 \cdot \bar{\omega}_t^T \right). \nonumber
		\end{align}
		
		%As shown in \eqref{eq:TD_optimal_condition}, we have the equality that $ A \cdot \omega^* = \frac{1}{N} \sum_{i = 1}^N b_i $ where $A$ is a positive definite matrix with $ c_{\min}(A), c_{\max}(A) $ as its minimum, maximum eigenvalues and
	%\begin{align}
	%A := E_\pi[ \phi(s)(\phi(s) - \gamma \phi(s^\prime))^T ], \quad b_i: = E_\pi[r_i\cdot \phi(s)], \; \forall~i \in \mathcal{N}. \label{eq:TD_fixed_point}
	%\end{align} 
		
		By the definition of locally estimated TD error in \eqref{alg: critic updates}, it follows that 
		\begin{align}
		\delta_{i,t} & := r_{i,t} + \gamma \cdot \widehat{V}(s_{t+1} ; \omega_{i,t}) - \widehat{V}(s_t; \omega_{i,t}) \nonumber\\
		& = r_{i,t} + \gamma\cdot  \phi(s_{t+1})^T \omega_{i,t} - \phi(s_t)^T\omega_{i,t}  \label{eq:delta}   
		\end{align}
		 where $\phi(\cdot) \in \mathbb{R}^K$ is the feature mapping for any state $s \in \mathcal{S}$.
		To perform the critic step according to equation \eqref{alg: critic updates}, it holds that 
		\begin{align}
		    \omega_{i,t} = \Pi_{R_\omega} \bigg(\sum_{j \in \mathcal{N}_i(t)} W_{t-1}^{ij} \omega_{i,t-1} + \beta_t \cdot \delta_{i,t-1} \cdot \nabla_{ \omega_i} \widehat{V}(s_{t-1}; \omega_{i,t-1}) \bigg), \nonumber \\
		    =\Pi_{R_\omega} \bigg(\sum_{j \in \mathcal{N}_i(t)} W_{t-1}^{ij} \omega_{i,t-1} + \beta_t \cdot \delta_{i,t-1} \cdot \phi(s_{t-1}) \bigg), \quad  \forall i \in \mathcal{N} \label{eq:TD:sto_grad}
		\end{align}
		where $ \nabla_{ \omega_i} \widehat{V}(s_{t-1}; \omega_{i,t-1}) := \phi(s_{t-1}) $ due to linear parameterization.
		 Recall that $A(\btheta_t)$, $b_{i}(\btheta_t)$ are defined in \eqref{eq: TD fixed point}, it holds that
		\begin{align}
		    b_{i}(\btheta_t) - A(\btheta_t) \cdot \omega_{i,t} &:= \mathbb{E}_{s \sim \mu_{\btheta_t}(\cdot), \ba \sim \pi_{\btheta_t}(\cdot|s), s^\prime \sim \mathcal{P}(\cdot| s, \ba) } \bigg[ \big( r_i(s,a) + \gamma \cdot \phi(s)^T \omega_{i,t} - \phi(s^\prime)^T \omega_{i,t} \big) \cdot \phi(s) \bigg] \nonumber \\
		    &= \mathbb{E}_{s_t \sim \mu_{\btheta_t}(\cdot), \ba_t \sim \pi_{\btheta_t}(\cdot|s_t), s_{t+1} \sim \mathcal{P}(\cdot| s_t, \ba_t) } \bigg[ \delta_{i,t} \cdot \phi(s_t) \bigg]. \label{eq:critic:unbiased_gradient}
		\end{align}
		
		Hence, in \eqref{eq:TD:sto_grad} the estimated stochastic gradient at each iteration $t$ is expressed as \begin{align}
		    \delta_{i,t} \cdot \phi(s_t) 
		 &= \underbrace{\bigg(\delta_{i,t} \cdot \phi(s_t) - [b_{i}(\btheta_t) - A(\btheta_t) \cdot \omega_{i,t}]\bigg)}_{:= m_{i,t}} + \underbrace{\bigg( b_{i}(\btheta_t) - A(\btheta_t) \cdot \omega_{i,t} \bigg)}_{:=h_{i,t}}  \label{eq:delta:phi}
		\end{align}
		%\begin{align}
		    %\delta_{i,t} \cdot \phi(s_t):= h_{i,t} + m_{i,t} \label{eq:TD_estimated_stoc_grad}
		%\end{align}
		%we define $ h_{i,t} := b_{i}(\btheta_t) - A(\btheta_t) \cdot \omega_{i,t} $ and $ m_{i,t} := \delta_{i,t} \cdot \phi(s_t) - [b_{i}(\btheta_t) - A(\btheta_t) \cdot \omega_{i,t}] $; 
		 where $h_{i,t}$ is the expectation of the estimated stochastic gradient $\delta_{i,t} \cdot \phi(s_t) $; $m_{i,t}$ denotes the deviation between $\delta_{i,t} \cdot \phi(s_t) $ and its expectation $h_{i,t}$.
		
		Recall that the subroutine to update critic parameters $ \bomega_t $ in \eqref{alg: critic updates}
		is given below:
		\begin{align}
		    \widetilde{\bomega}_{t-1} = W_{t-1} \cdot \bomega_{t-1}, \quad \omega_{i,t} = \Pi_{R_\omega} \bigg(\widetilde{ \omega}_{i,t-1} + \beta_{t-1} \cdot \delta_{i,t-1} \cdot \nabla_{ \omega_i} \widehat{V}(s_{t-1}; \omega_{i,t-1}) \bigg), ~  \forall i \in \mathcal{N}.
		\end{align}
		It can be decomposed using the following steps:
		\begin{subequations} \label{subroutine:critic_update}
		   \begin{align}
	\widetilde{\omega}_{i,t-1} &= \sum_{j \in \mathcal{N}_i(t-1)} W_{t-1}^{ij} \cdot \omega_{i,t-1}  \\
	y_{i,t-1} &\stackrel{\eqref{eq:delta:phi}}= \widetilde{\omega}_{i,t-1} + \beta_{t-1} \cdot (h_{i,t-1} + m_{i,t-1})  \\
	e_{i,t-1} &= y_{i,t-1} - \Pi_{R_{\omega}}(y_{i,t-1})  \\
	\omega_{i,t} &= \Pi_{R_{\omega}}(y_{i,t-1}) = y_{i,t-1} - e_{i,t-1}. 
\end{align}
\end{subequations}
Express the above updates in matrix form, it holds that
\begin{subequations} \label{subroutine:critic_update_matrix_form}
   \begin{align}
	\widetilde{\bomega}_{t-1} &= W_{t-1} \cdot \bomega_{t-1} \\
	\bm y_{t-1} &= \widetilde{\bomega}_{t-1} + \beta_{t-1} \cdot (\bm h_{t-1} + \bm m_{t-1})  \\
	\bm e_{t-1} &= \bm y_{t-1} - \Pi_{R_\omega}(\bm y_{t-1})  \\
	\bomega_{t} &= \bm y_{t-1} - \bm e_{t-1} 
\end{align}
\end{subequations}
where $ \bm y_t, \bm h_t, \bm m_t, \bm e_t $ correspond to the collections of local vectors $ \{y_{i,t}\}, \{h_{i,t}\}, \{m_{i,t}\}, \{e_{i,t}\} $. Recall the definitions $\bomega_t = [\omega_{1,t}^T; \omega_{2,t}^T; \cdots; \omega_{N,t}^T ] \in \mathbb{R}^{N \times K}$ and $ \bar{\omega}_t^T := \frac{1}{N} \bm 1^T \bomega_t $, it follows
\begin{align}
    \bar{\omega}_t^T & = \frac{1}{N} \bm 1^T \bomega_t \nonumber \\
    & \overset{(i)}{=}  \frac{1}{N} \bm 1^T \big( W_{t-1}\cdot \bomega_{t-1} + \beta_{t-1} \cdot (\bm{h}_{t-1} + \bm{m}_{t-1}) -  \bm{e}_{t-1} \big) \nonumber\\
    & \overset{(ii)}{=} \bar{\omega}_{t-1}^{T} + \beta_{t-1} \cdot (\bar{h}_{t-1}^{T} + \bar{m}_{t-1}^{T}) -  \bar{e}_{t-1}^{T} \label{eq:average:omega}
\end{align}
where $\bar{h}_{t-1}$, $\bar{m}_{t-1}$ and $\bar{e}_{t-1}$ are the averaged vectors of $\mathbf{h}_{t-1}$, $\mathbf{m}_{t-1}$ and $\mathbf{e}_{t-1}$ (as defined similarly as $\bar{\omega}_{t-1}$); $(i)$ is from the subroutine \eqref{subroutine:critic_update_matrix_form}; $(ii)$ is from $\bm 1^T W_{t-1} = \bm 1^T$ due to double stochasticity in weight matrix $W_{t-1}$.
Recall that we have defined $ Q = I - \frac{1}{N} \bm 1 \cdot \bm 1^T $, then it is clear that $Q\cdot \bomega$ indicates the consensus error. We can express such an error as follows: %{\red[why $\bar{w}_t$ has a transpose? it is supposed to be a column vector right?]}
		\begin{align}
		Q \cdot \bomega_{t} 
		& = \bomega_{t} -  \bm 1 \bar{\omega}_{t}^T \nonumber \\
		& = W_{t-1} \cdot \bomega_{t-1} + \beta_{t-1} \cdot \left( \bm h_{t-1} + \bm m_{t-1} \right) - \bm e_{t-1} - [ \bm1 \bar{\omega}_{t-1}^T + \beta_{t-1} \cdot \bm1 [\bar{h}_{t-1} + \bar{m}_{t-1}]^T - \bm1 \bar{e}_{t-1}^T ] \nonumber \\
		& = W_{t-1} \cdot \left( \bomega_{t-1} -  \bm 1 \bar{\omega}_{t-1}^T \right) + \beta_{t-1} \cdot \left( \bm h_{t-1} - \bm1 \bar{h}_{t-1}^T \right) + \beta_{t-1} \cdot \left( \bm m_{t-1} - \bm 1 \bar{m}_{t-1}^T \right) - \left(\bm e_{t-1} - \bm 1 \bar{e}_{t-1}^T \right) \nonumber \\
		& = W_{t-1} \cdot Q  \cdot  \bomega_{t-1} + \beta_{t-1} \cdot Q \cdot (\bm h_{t-1} + \bm m_{t-1}) - Q \cdot \bm e_{t-1} \label{eq:consensus_critic_expand} \\
		& = \left(\Pi_{k = 0}^{t-1} W_k\right) \cdot Q \cdot   \bomega_0 - \sum_{k = 0}^{t-1} (\Pi_{\ell = k + 1}^{t-1} W_{\ell})  \cdot Q \cdot  \bm e_k + \sum_{k = 0}^{t-1} \beta_k \cdot  (\Pi_{\ell = k + 1}^{t-1} W_{\ell})  \cdot Q \cdot  \left( \bm h_k + \bm m_k \right). \label{eq:consensus_critic_telescope}
		\end{align}
 Then we can bound the norm of the consensus error using the following:
		\begin{align}
		\| Q \cdot \bomega_{t} \| 
		&\overset{(i)}{\leq} \|(\Pi_{k = 0}^{t-1} W_k) \cdot Q \cdot  \bomega_0\| + \sum_{k = 0}^{t-1} \| (\Pi_{\ell = k + 1}^{t-1} W_{\ell}) \cdot Q\cdot  \bm e_{k-1} \| \nonumber\\
		& \quad + \sum_{k = 0}^{t-1} \beta_k \|  (\Pi_{\ell = k + 1}^{t-1} W_{\ell}) \cdot Q \cdot  \left( \bm h_k + \bm m_k \right) \| \nonumber \\
		& \overset{(ii)}{\leq} \eta^{ \lfloor t/B \rfloor }  \cdot \| \bomega_0 \| + \sum_{k = 0}^{t-1} \eta^{ \lfloor (t - k -1) / B \rfloor } \cdot \| \bm e_k \| + \sum_{k = 0}^{t-1} \eta^{\lfloor (t - k - 1)/ B \rfloor} \cdot \beta_k \cdot \| \bm h_k + \bm m_k \| \nonumber \\
		&\overset{(iii)}{\leq} \frac{1}{\eta}\cdot \rho^{t} \cdot \| \bomega_0 \| + \frac{1}{\eta} \sum_{k = 0}^{t-1} \rho^{t - k - 1} \cdot \| \bm e_k \| + \frac{1}{\eta} \sum_{k = 0}^{t-1} \rho^{t - k - 1} \cdot \beta_k \cdot  \| \bm h_k + \bm m_k \| \label{bound: consensus_error_1}
		\end{align}
		where (ii) follows \eqref{bound: graph connectivity}; in (iii) we utilize that $ \eta^{ \lfloor t/B \rfloor } \leq \eta^{ t/B - 1} = \frac{1}{\eta} \cdot \rho^{t} $ where we define $\rho := \eta^{\frac{1}{B}}$.
		
		Next we  bound $ \| \bm h_k + \bm m_k \| $ and $ \| \bm e_k \| $. We have %recall that the critic parameter $ \omega_{i,t} $ is constrained in a fixed region with radius $R_{\omega}$. Hence, according to Assumptions \ref{ass: reward_bound} -\ref{ass:feature_bound}, it holds that there exists a positive constant $ L_b $ to satisfy that 
		\begin{align}
		    \| h_{i,t} \| + \| m_{i,t} \| 
		    & \overset{\eqref{eq:delta:phi}}{=} \bigg \| \delta_{i,t} \cdot \phi(s_t) - [b_{i}(\btheta_t) - A(\btheta_t) \cdot \omega_{i,t}]\bigg \| + \bigg\| b_{i}(\btheta_t) - A(\btheta_t) \cdot \omega_{i,t} \bigg\| \nonumber \\
		    &\leq \| \delta_{i,t} \cdot \phi(s_t) \| + 2 \cdot \| b_{i}(\btheta_t) - A(\btheta_t) \cdot \omega_{i,t}  \| \nonumber \\
		    &\overset{\eqref{eq: TD fixed point}}{=} \| \delta_{i,t} \cdot \phi(s_t) \| + 2\cdot \bigg\| \mathbb{E}_{s \sim \mu_{\btheta_t}(\cdot), \ba \sim \pi_{\btheta_t}(\cdot|s), s^\prime \sim \mathcal{P}(\cdot| s, \ba) } \bigg[ \big( r_i(s,a) + \gamma \cdot \phi(s)^T \omega_{i,t} - \phi(s^\prime)^T \omega_{i,t} \big) \cdot \phi(s) \bigg] \bigg\| \nonumber \\
		    &\overset{(i)}{\leq} \| \delta_{i,t} \cdot \phi(s_t) \| + 2\cdot \mathbb{E}_{s \sim \mu_{\btheta_t}(\cdot), \ba \sim \pi_{\btheta_t}(\cdot|s), s^\prime \sim \mathcal{P}(\cdot| s, \ba) } \bigg[ \bigg\| \big( r_i(s,a) + \gamma \cdot \phi(s)^T \omega_{i,t} - \phi(s^\prime)^T \omega_{i,t} \big) \cdot \phi(s) \bigg\| \bigg] \nonumber \\
		    & \overset{(ii)}{\leq} \| \delta_{i,t} \| + 2\cdot \mathbb{E}_{s \sim \mu_{\btheta_t}(\cdot), \ba \sim \pi_{\btheta_t}(\cdot|s), s^\prime \sim \mathcal{P}(\cdot| s, \ba) } \bigg[ \bigg\| \big( r_{i}(s,a) + \gamma \cdot \phi(s)^T \omega_{i,t} - \phi(s^\prime)^T \omega_{i,t} \big) \bigg\| \bigg] \nonumber \\
		    &\overset{\eqref{eq:delta}}{=} \bigg \| r_{i,t} + \gamma\cdot  \phi(s_{t+1})^T \omega_{i,t} - \phi(s_t)^T\omega_{i,t} \bigg \| \nonumber \\
		   & \quad + 2\cdot \mathbb{E}_{s \sim \mu_{\btheta_t}(\cdot), \ba \sim \pi_{\btheta_t}(\cdot|s), s^\prime \sim \mathcal{P}(\cdot| s, \ba) } \bigg[ \bigg\| \big( r_{i}(s,a) + \gamma \cdot \phi(s)^T \omega_{i,t} - \phi(s^\prime)^T \omega_{i,t} \big) \bigg\| \bigg]  \nonumber \\
		    & \overset{(iii)}{\leq} 3R_{\max} + 3(1+\gamma)\cdot R_{\omega} \label{bound: estimator}
		\end{align}
		where $(i)$ follows Jensen's inequality; $(ii)$ follows Assumption \ref{ass:feature_bound} that $\|\phi(s) \| \leq 1$ for any $s \in \mathcal{S}$; $(iii)$ follows the fact that $|r_i(s,a)| \leq R_{\max}$ and the critic parameter $ \omega_{i,t} $ is constrained in a fixed region $\| \omega_{i,t} \| \leq R_{\omega}$. For simplicity, in the following part we denote $L_b := 3R_{\max} + 3(1+\gamma)\cdot R_{\omega}$.
		
		Moreover, it holds that
		\begin{align}
		\| e_{i,t} \| \stackrel{\eqref{subroutine:critic_update}}= \| y_{i,t} - \Pi_{R_{\omega}}(y_{i,t}) \|
		\overset{(i)}{\leq} \| \widetilde{\omega}_{i,t} - y_{i,t} \|
		= \| \beta_t \cdot \left( h_{i,t} + m_{i,t} \right) \|
		\overset{(ii)}{\leq} \beta_t \cdot L_b \label{bound: projection error},
		\end{align}
		where $(i)$ follows from \eqref{bound: nedic 2} and $(ii)$ follows from \eqref{bound: estimator}.

		Recall that the stepsizes in critic steps are defined as $\beta_t: = \frac{\beta}{T^{\sigma_2}}, ~ \forall t $. Plugging (\ref{bound: estimator}) and (\ref{bound: projection error}) into (\ref{bound: consensus_error_1}), we get
		\begin{align}
		\| Q \cdot \bomega_{t} \| & \leq \frac{1}{\eta} \cdot \rho^{t} \cdot \| \bomega_0 \| + \frac{1}{\eta} \sum_{k = 0}^{t-1} \rho^{t-k-1} \cdot \beta_k \cdot L_b \cdot N + \frac{1}{\eta} \sum_{k = 0}^{t-1} \rho^{t-k-1}\cdot \beta_k \cdot L_b \cdot N \\
		& = \frac{1}{\eta} \cdot \rho^{t} \cdot \| \bomega_0 \| + \frac{2N \cdot L_b \cdot \beta}{\eta \cdot T^{\sigma_2}} \cdot \sum_{k = 0}^{t-1} \rho^{t - k - 1} \nonumber \\
		& \overset{(a)}{\leq} \frac{1}{\eta} \cdot \rho^{t} \cdot \| \bomega_0 \| + \frac{2N \cdot L_b \cdot \beta}{\eta \cdot T^{\sigma_2}} \cdot \frac{1}{1 - \rho}
		\label{bound:omega:consensus_error}
		\end{align}
		where $(a)$ follows the fact that $ \sum_{k = 0}^{t-1} \rho^{t - k - 1} \leq \frac{1}{1 - \rho}, ~ \forall t $.
		
		Summing \eqref{bound:omega:consensus_error} from $t = 0$ to $t = T-1$, we obtain
		   \begin{align}
		\sum_{t = 0}^{T-1} \| Q \cdot \bomega_t \| &\leq \frac{\| \bomega_0 \|}{\eta} \sum_{t=0}^{T-1} \rho^t + T \cdot \frac{2N \cdot L_b \cdot \beta}{\eta \cdot T^{\sigma_2}} \cdot \frac{1}{1 - \rho} \nonumber \\
		&\overset{(i)}{\leq} \frac{\| \bomega_0 \|}{\eta\cdot (1 - \rho)} + \frac{2N \cdot L_b \cdot \beta}{\eta \cdot (1 - \rho)} \cdot T^{1-\sigma_2}, \label{eq:critic_consensus_error_sum}
		\end{align}
		where $(i)$ is due to the fact that $ \sum_{t = 0}^{T-1}\rho^{t} \leq \frac{1}{1-\rho}$.

		In summary, we obtain the following bound on the averaged consensus violation: 
		\begin{align}
		    \frac{1}{T} \sum_{t=0}^{T-1} \| Q \cdot \bomega_t \| \leq  \frac{1}{T} \cdot \frac{\| \bomega_0 \|}{\eta\cdot (1 - \rho)} +  + \frac{2N \cdot L_b \cdot \beta}{\eta \cdot (1 - \rho)} \cdot T^{-\sigma_2} \label{rate:consensus_error:critic}.
		\end{align}

		Extending above analysis steps on deriving a bound for the consensus error  $\|Q \cdot \bomega_t\|$ in \eqref{eq:critic_consensus_error_sum} to $ \| Q \cdot \blambda_t \| $, we can show that the following holds:
		\begin{align}
		    \| Q \cdot \blambda_t \| \leq \frac{1}{\eta} \cdot \rho^{t} \cdot \| \blambda_0 \| + \frac{2N \cdot L_b \cdot \zeta}{\eta \cdot T^{\sigma_2}} \cdot \frac{1}{1 - \rho} \label{eq:reward_consensus_error}
		\end{align}
	where the stepsize $\zeta_t$ is defined as $ \zeta_t := \frac{\zeta}{T^{\sigma_2}} $. Similar as \eqref{eq:critic_consensus_error_sum}, summing up \eqref{eq:reward_consensus_error} from $t=0$ to $t = T-1$, it holds that:
	\begin{align}
	    \sum_{t=0}^{T-1} \| Q \cdot \blambda_t \| \leq \frac{\| \blambda_0 \|}{\eta\cdot (1 - \rho)} +  + \frac{2N \cdot L_b \cdot \zeta}{\eta \cdot (1 - \rho)} \cdot T^{1-\sigma_2}. \label{bound:reward_consensus_error}
	\end{align}
	In summary, we have 
	\begin{align}
	    \frac{1}{T} \sum_{t=0}^{T-1} \| Q\cdot  \blambda_t \| \leq  \frac{1}{T} \cdot \frac{\| \bomega_0 \|}{\eta\cdot (1 - \rho)} +  + \frac{2N \cdot L_b \cdot \zeta}{\eta \cdot (1 - \rho)} \cdot T^{-\sigma_2} \label{rate:consensus_error:reward}.
	\end{align}
	
	Summing up \eqref{rate:consensus_error:critic} and \eqref{rate:consensus_error:reward}, we obtain the convergence rate of the consensus errors:
	\begin{align}
	    \frac{1}{T} \sum_{t=0}^{T-1} \bigg( \| Q \cdot \bomega_t \| + \| Q \cdot \blambda_t \| \bigg) &= \frac{1}{T} \cdot \frac{\| \bomega_0 \| + \| \blambda_0 \|}{\eta\cdot (1 - \rho)} +  \frac{1}{T^{\sigma_2}} \cdot \frac{2N \cdot L_b \cdot (\beta + \zeta)}{\eta\cdot (1 - \rho)} \nonumber \\
	    &= \mathcal{O}(T^{-\sigma_2}).
	\end{align}
	
	Taking square on both side of \eqref{bound:omega:consensus_error} and applying Cauchy-Schwarz inequality, it holds that
	\begin{align}
	    \| Q \cdot \bomega_{t} \| ^2
		& \leq \frac{2}{\eta^2} \cdot \rho^{2t} \cdot \| \bomega_0 \|^2 + \frac{2}{T^{2\sigma_2}} \cdot \bigg( \frac{2N \cdot L_b \cdot \beta}{\eta \cdot (1 - \rho)} \bigg)^2 \label{bound:omega:consensus:square}
	\end{align}
	
	Summing \eqref{bound:omega:consensus:square} from $t = 0$ to $t = T-1$, it holds that 
	\begin{align}
	    \sum_{t = 0}^{T-1}  \| Q \cdot \bomega_{t} \| ^2
		& \leq \frac{2 \| \bomega_0 \|^2}{\eta^2 \cdot (1 - \rho^2)} + 2 \cdot \bigg( \frac{2N \cdot L_b \cdot \beta}{\eta \cdot (1 - \rho)} \bigg)^2 \cdot T^{1 - 2\sigma_2} \label{sum:omega:consensus:square}
	\end{align}
	
	Extending above analysis steps on deriving a bound for the consensus error  $\|Q \cdot \bomega_t\|^2$ in \eqref{sum:omega:consensus:square} to $ \| Q \cdot \blambda_t \|^2 $, we can show that the following holds:
	\begin{align}
	    \sum_{t = 0}^{T-1}  \| Q \cdot \blambda_{t} \| ^2
		& \leq \frac{2 \| \blambda_0 \|^2}{\eta^2 \cdot (1 - \rho^2)} + 2 \cdot \bigg( \frac{2N \cdot L_b \cdot \zeta}{\eta \cdot (1 - \rho)} \bigg)^2 \cdot T^{1 - 2\sigma_2} \label{sum:lambda:consensus:square}
	\end{align}
	
	Hence, adding \eqref{sum:omega:consensus:square} and \eqref{sum:lambda:consensus:square}, then divide both side by $T$, it holds that
	\begin{align}
	    \frac{1}{T} \sum_{t = 0}^{T-1} \bigg( \| Q \cdot \blambda_{t} \| ^2 + \| Q \cdot \bomega_{t} \| ^2 \bigg)
		& \leq \frac{2 (\| \blambda_0 \|^2 + \| \bomega_0 \|^2 )}{T \cdot \eta^2 \cdot (1 - \rho^2)} +  \frac{8N^2 \cdot L_b^2 \cdot (\zeta^2 + \beta^2)}{\eta^2 \cdot (1 - \rho)^2}  \cdot T^{ - 2\sigma_2} \nonumber \\
		&= \mathcal{O}(T^{-1}) + \mathcal{O}(T^{-2\sigma_2}) \label{rate:lower:consensus:square}
	\end{align}
	
	This completes the proof for the first part.
	\end{proof} 
	
	\begin{proof}[{\bf Step 2.}] \label{proof: Lemma policy consensus error}
	
	In this part, we analyze the consensus errors for the shared policy parameters $ \btheta^s $.
	
		Since the mixing matrix $ W_t $ is doubly stochastic which implies $W_t \cdot \bm 1 = \bm 1$, we obtain that 
		\begin{align}
		W_t \cdot \btheta_t^s  - \bm 1 \bar{\theta}_{t}^{s^T} = W_t \cdot \left( \bm \theta_t^s - \bm 1 \bar{\theta}_{t}^{s^T} \right) \nonumber
		\end{align}
		where we have defined $ \bar{\theta}_{t}^{s^T} := \frac{1}{N}\bm 1^T \btheta_{t}^{s} $.
		Recall that the subroutine to update shared policy parameters in \eqref{eq:actor_consensus} - \eqref{alg: actor_local updates} is given below:
		\begin{align}
	  \theta_{i,t}^s & := \sum_{j \in \mathcal{N}_i(t-1)} W_{t-1}^{ij} \theta^{s}_{j,t-1} + \alpha_{t-1} \cdot \widehat{\delta}_{i,t-1}  \cdot \nabla_{ \theta_i^s} \log \pi_i( a_{i,t-1}| s_{t-1}, \theta_{i,t-1}), ~~ \forall i \in \mathcal{N} 
	  \end{align}
	  where we have defined
	  \begin{align}
	      \widehat{\delta}_{i,t-1} & := \widehat{r}(s_{t-1}, \ba_{t-1}; \lambda_{i,t-1}) + \gamma\cdot \widehat{V}(s_{t}; \omega_{i,t-1}) - \widehat{V}(s_{t-1} ; \omega_{i,t-1})\nonumber\\
	       & = \varphi(s_{t-1}, \bm a_{t-1})^T \lambda_{i,t-1} + \gamma \cdot \phi(s_{t-1})^T \omega_{i,t-1} - \phi(s_{t})^T \omega_{i,t}.
	      \label{eq:policy_delta}
	  \end{align}
% 	  $ $. By linear parameterizations, it holds that 
% 	  \begin{align}
% 	      \widehat{\delta}_{i,t-1} :
% 	  \end{align}
	  Then we define $ g_{i,t-1} := \widehat{\delta}_{i,t-1} \cdot \nabla_{ \theta_i^s} \log \pi_i( a_{i,t-1}| s_{t-1}, \theta_{i,t-1}) $ and $\bm g := [g_1^T; g_2^T; \cdots; g_N^T]$, it holds that 
	  \begin{align}
	      \btheta_t^s := W_{t-1} \cdot \btheta_{t-1}^s + \alpha_{t-1} \cdot \bm g_{t-1}. \label{subroutine:theta:shared}
	  \end{align}
	  Recall $Q = I - \frac{1}{N}\bm 1 \cdot \bm 1^T$, we analyze the consensus error $ Q \cdot \btheta_{t}^s $ as below:
	    \begin{align}
		Q \cdot \btheta_{t}^s =& \btheta_{t}^s - \bm 1 \bar{\theta}_{t}^{s^T} \nonumber \\
		=& \bigg( W_{t-1} \cdot \btheta_{t-1}^s + \alpha_{t-1} \cdot \bm g_{t-1} \bigg) - \bigg( \bm 1 \bar{\theta}_{t-1}^{s^T} + \alpha_{t-1} \cdot \bm 1 \bar{g}_{t-1}^T \bigg)  \nonumber \\
		=& W_{t-1} \cdot Q \cdot \btheta_{t-1}^s + \alpha_{t-1} \cdot Q \cdot \bm g_{t-1}   \nonumber \\
		=& (\Pi_{k = 0}^{t-1} W_k) \cdot Q \cdot \btheta_{0}^s + \sum_{k = 0}^{t-1} \alpha_k \cdot (\Pi_{l = k + 1}^{t-1} W_l) \cdot Q \cdot \bm g_k. \label{eq:decompose_theta_consensus_error}
		\end{align}
	By Assumptions \ref{ass: reward_bound} - \ref{ass:feature_bound}, \ref{ass: Lipschitz policy}, the estimated stochastic gradient $\bm g_t = [g_{1,t}^T, g_{2,t}^T, \cdots, g_{N,t}^T]$ can be bounded as below:
		\begin{align}
		    \| \bm g_{t} \| & \overset{(i)}{\leq} \sum_{i = 1}^{N} \| g_{i,t} \| \nonumber \\
		    &= \sum_{i = 1}^{N} \| \widehat{\delta}_{i,t-1} \cdot \nabla_{ \theta_i^s} \log \pi_i( a_{i,t-1}| s_{t-1}, \theta_{i,t-1})\| \nonumber \\
		    & \overset{(ii)}{\leq} C_{\psi} \cdot  \sum_{i = 1}^{N} \bigg\| \varphi(s_{t-1}, \bm a_{t-1})^T \lambda_{i,t-1} + \gamma \cdot \phi(s_{t-1})^T \omega_{i,t-1} - \phi(s_{t})^T \omega_{i,t} \bigg\|  \nonumber \\
		    & \leq C_{\psi} \cdot  \sum_{i = 1}^{N} \bigg( \| \varphi(s_{t-1}, \bm a_{t-1})\| \cdot \| \lambda_{i,t-1}\| + \gamma \| \phi(s_{t-1})\| \cdot \|\omega_{i,t-1}\| + \| \phi(s_{t})\| \cdot \| \omega_{i,t} \| \bigg)  \nonumber \\
		    & \overset{(iii)}{\leq} N \cdot C_{\psi} \cdot \bigg(R_{\lambda} + (1+\gamma) \cdot R_{\omega}\bigg): = \ell_p \label{bound:policy_gradient}
		\end{align}
		where $(i)$ follows the definition $\bm g_t = [g_{1,t}^T, g_{2,t}^T, \cdots, g_{N,t}^T]$ and Triangle inequality; $(ii)$ follows that $ \| \nabla_{ \theta_i^s} \log \pi_i( a_{i,t-1}| s_{t-1}, \theta_{i,t-1}) \| \leq C_{\psi} $ in Assumption \ref{ass: Lipschitz policy};  $(iii)$ is due to the assumptions that $\| \varphi(s_{t-1}, \bm a_{t-1}) \| \leq 1$ and $\| \phi(s_{t-1}) \| \leq 1$, as well as that approximation parameters are restricted in fixed regions, so $\| \omega_{i,t} \| \leq R_{\omega}$ and $\| \lambda_{i,t} \| \leq R_{\lambda}$; In the last equality, we have defined $\ell_p$ as 
		\begin{align}
		    \ell_p := N \cdot C_{\psi} \cdot \bigg( R_{\lambda} + (1+\gamma)R_{\omega} \bigg). \label{eq: policy gradient bound}
		\end{align}
		Recall that the stepsizes in policy optimization are defined as $\alpha_t : = \frac{\alpha}{T^{\sigma_1}}, \forall ~ t $. Taking Frobenius norm on both side of \eqref{eq:decompose_theta_consensus_error}, we have:
		\begin{align}
		\| Q \cdot \btheta_{t}^s \| &\overset{(i)}{\leq} \| (\Pi_{k = 0}^{t-1} W_k) \cdot Q \cdot \btheta_{0}^s \| + \sum_{k = 0}^{t-1} \alpha_k \| (\Pi_{l = k + 1}^{t-1} W_l) \cdot Q \cdot \bm g_k \| \nonumber \\
		&\overset{(ii)}{\leq} \eta^{ \lfloor t/\mathcal{B} \rfloor } \cdot \| \bm \theta_{0}^s  \| + \sum_{k = 0}^{t-1} \alpha_k \cdot \eta^{ \lfloor (t-k-1)/\mathcal{B} \rfloor } \cdot \| \bm g_k \|  \nonumber \\
		&\overset{(iii)}{\leq} \frac{1}{\eta} \cdot  \rho^{t} \cdot \| \btheta_{0}^s  \| + \frac{1}{\eta}  \sum_{k = 0}^{t-1} \alpha_k \cdot \rho^{t-k-1} \cdot \| \bm g_k \| \nonumber \\
		&\overset{(iv)}{\leq}  \frac{1}{\eta} \cdot \rho^{t} \cdot \| \btheta_{0}^s \| + \frac{\ell_p}{\eta} \sum_{k = 0}^{t-1} \rho^{t - k -1} \cdot \alpha_k \nonumber \\
		& = \frac{1}{\eta} \cdot \rho^{t} \cdot \| \btheta_{0}^s \| + \frac{\ell_p \cdot \alpha}{\eta \cdot T^{\sigma_1}} \cdot \sum_{k = 0}^{t-1} \rho^{t - k -1} \nonumber \\
		& \leq \frac{1}{\eta} \cdot \rho^{t} \cdot \| \btheta_{0}^s \| + \frac{\ell_p \cdot \alpha}{\eta \cdot (1-\rho)} \cdot \frac{1}{T^{\sigma_1}}
		\label{eq:policy_consensus_error}
		\end{align}
		where $(ii)$ follows from \eqref{bound: graph connectivity}; in $(iii)$ we utilize that $ \eta^{ \lfloor t/B \rfloor } \leq \eta^{ t/B - 1} = \frac{1}{\eta} \cdot \rho^{t} $ where we have defined $\rho := \eta^{\frac{1}{B}}$; $(iv)$ follows from \eqref{bound:policy_gradient}. Summing \eqref{eq:policy_consensus_error} from $ t = 0 $ to $ t = T-1 $, it holds that:
		\begin{align}
		 \sum_{t=0}^{T-1} \| Q \cdot \btheta_{t}^s \| &\leq \frac{\| \btheta_{0}^s  \|}{\eta} \sum_{t = 0}^{T-1} \rho^t + \frac{\ell_p \cdot \alpha_0}{\eta \cdot (1-\rho)} \cdot \frac{T}{T^{\sigma_1}} \nonumber \\
		 &\leq \frac{\| \btheta_{0}^s  \|}{\eta \cdot (1 - \rho)} + \frac{\ell_p \cdot \alpha_0}{\eta \cdot (1-\rho)} \cdot T^{1-\sigma_1} \nonumber \\
		 &= \mathcal{O}(T^{1-\sigma_1}). \label{rate:policy_consensus_error}
		\end{align}
		
		%\begin{align}
		%\sum_{t = 0}^{T-1} \| Q \cdot \bm \btheta_{t}^s \| \leq& \frac{\| \btheta_{0}^s  \|}{\eta}  \sum_{t = 0}^{T-1} \rho^t + \frac{\ell_p}{\eta} \sum_{t = 0}^{T-1} \sum_{k = 0}^{t} \rho^{t-1-k}\alpha_k \nonumber \\
		%\leq& \frac{\| \btheta_{0}^s  \|}{\eta (1 - \rho)} + \frac{\ell_p}{\eta} \sum_{k = 0}^{T-1} \alpha_k \sum_{t = k}^{T-1} \rho^{t-k-1}   \nonumber \\
		%\leq& \frac{\| \btheta_{0}^s  \|}{\eta (1 - \rho)}  + \frac{\ell_p}{\eta} \frac{1}{\rho (1 - \rho)} \frac{\alpha_0 T^{1 - \sigma_1}}{1 - \sigma_1}. \nonumber
		%\end{align}
		Then dividing $T$ on both side of \eqref{rate:policy_consensus_error}, it holds that
		\begin{align}
		    \frac{1}{T}\sum_{t = 0}^{T-1} \| Q \cdot \btheta_{t}^s \| \leq \mathcal{O}(T^{- \sigma_1})
		    \label{rate:policy_consensus}
		\end{align}
		where consensus error converges to $ 0 $ as $ T $ goes to infinity. 
		
		Moreover, we can provide a bound for the averaged consensus error squared $\sum_{t = 0}^{T-1} \| Q \cdot \btheta_{t}^s \|^2$. Taking square on both sides of \eqref{eq:policy_consensus_error} and summing from $t = 0$ to $t = T-1$. We obtain:
		\begin{align}
		\sum_{t = 0}^{T-1} \| Q \cdot \btheta_{t}^s \|^2 & \overset{\eqref{eq:policy_consensus_error}}{\leq} \sum_{t = 0}^{T-1} \bigg( \frac{1}{\eta} \cdot \rho^{t} \cdot \| \btheta_{0}^s \| + \frac{\ell_p \cdot \alpha}{\eta \cdot (1-\rho)} \cdot \frac{1}{T^{\sigma_1}} \bigg)^2 \nonumber \\
		&\overset{(i)}{\leq}  \sum_{t = 0}^{T-1}  \bigg( \rho^{2t} \cdot \frac{2\| \btheta_{0}^s  \|^2 }{\eta^2} + \frac{2 \ell_p^2 \cdot \alpha^2}{\eta^2 \cdot (1 - \rho)^2} \cdot \frac{1}{T^{2\sigma_1}} \bigg)  \nonumber \\
		&\overset{(ii)}{\leq}  \frac{2 \| \btheta_{0}^s  \|^2}{\eta^2 \cdot (1 - \rho^2)}  + \frac{2 \ell_p^2 \cdot \alpha^2}{\eta^2 \cdot (1 - \rho)^2} \cdot T^{1 - 2\sigma_1} \label{eq:policy_consensus_square}
		\end{align}
		where $(i)$ follows from Cauchy–Schwarz inequality; $(ii)$ follows from $ \frac{2\| \btheta_{0}^s  \|^2 }{\eta^2} \cdot \sum_{t = 0}^{T-1}   \rho^{2t} \leq \frac{2 \| \btheta_{0}^s  \|^2}{\eta^2 \cdot (1 - \rho^2)} $.
		
		Then dividing $T$ on both side of \eqref{eq:policy_consensus_square}, it holds that
		\begin{align}
		   \frac{1}{T} \sum_{t = 0}^{T-1} \| Q \cdot \btheta_{t}^s \|^2 \leq \mathcal{O}(T^{ - 1}) + \mathcal{O}(T^{ - 2\sigma_1})
		    \label{rate:policy_consensus_square}.
		\end{align}
	This completes the proof for the second step.
	\end{proof}

%	\clearpage
	
	\section{Proof of Proposition \ref{proposition:critic}} \label{appendix:proposition_critic}
	
	\begin{proof}\label{proof: critic}
	
	    In this proof, we show the convergence results of all approximation parameters $\bomega_t$ and $\blambda_t$. We first analyze the convergence error $\| \bar{\omega}_{t} - \omega^*(\btheta_t) \|$ and then extend the results to $\| \bar{\lambda}_{t} - \lambda^*(\btheta_t) \|$. For simplicity, we write $ \omega^*(\btheta_t) $ as $ \omega^*_t $ and $ \lambda^*(\btheta_t) $ as $ \lambda^*_t $.
	
		Denoting the expectation over data sampling procedures as $\EE[\cdot]$, let us begin by bounding the error $ \mathbb{E}[ \| \bar{\omega}_{t+1} - \omega^*_{t+1} \|^2 ] $ as below:
		\begin{align}
		&\mathbb{E}\left[ \| \bar{\omega}_{t+1} - \omega^*_{t+1} \|^2 \right] \nonumber \\
		& \overset{(i)}{=}  \mathbb{E}\left[ \| \bar{\omega}_{t} + \beta_t\cdot  (\bar{h}_t + \bar{m}_t) -  \bar{e}_t - \omega^*_{t} + \omega^*_{t} - \omega^*_{t+1} \|^2 \right] \nonumber \\
		& = \mathbb{E}[ \| \bar{\omega}_{t} - \omega^*_{t}  \|^2 ] + \mathbb{E}[\| \beta_t \cdot (\bar{h}_t + \bar{m}_t) -  \bar{e}_t + \omega^*_{t} - \omega^*_{t+1} \|^2] + 2\beta_t \cdot \mathbb{E} \left[ \langle \bar{\omega}_{t} - \omega^*_{t}, \bar{h}_t + \bar{m}_t \rangle \right] \nonumber \\
		& \quad - 2\mathbb{E}[\langle \bar{\omega}_{t} - \omega^*_{t}, \bar{e}_t \rangle] + 2\mathbb{E}\left[ \langle \bar{\omega}_{t} - \omega^*_{t}, \omega^*_{t} - \omega^*_{t+1} \rangle \right] \nonumber \\
		& \overset{(ii)}{\leq} \mathbb{E}[ \| \bar{\omega}_{t} - \omega^*_{t}  \|^2 ] +  \underbrace{2\beta_t \cdot \mathbb{E} \left[ \langle \bar{\omega}_{t} - \omega^*_{t}, \bar{h}_t + \bar{m}_t \rangle \right]}_{\rm term~A} + \underbrace{2\mathbb{E}[\| \omega^*_{t} - \omega^*_{t+1} \|^2] + 2\mathbb{E}\left[ \langle \bar{\omega}_{t} - \omega^*_{t}, \omega^*_{t} - \omega^*_{t+1} \rangle \right]}_{\rm term~B} \nonumber \\
		& \quad + \underbrace{2\mathbb{E}[\| \beta_t \cdot (\bar{h}_t + \bar{m}_t) -  \bar{e}_t \|^2]}_{\rm term~C} ~ \underbrace{-2\mathbb{E}[\langle \bar{\omega}_{t} - \omega^*_{t}, \bar{e}_t \rangle]}_{\rm term~D}  \label{bound: critic descent}
		\end{align}
		where $(i)$ follows \eqref{eq:average:omega}; $(ii)$ is from Cauchy-Schwarz inequality. Recall that $ \bar{h}_t^T := \frac{1}{N} \bm 1^T \bm h_t $, $ \bar{m}_t^T := \frac{1}{N} \bm 1^T \bm m_t $ and $ \bar{e}_t^T := \frac{1}{N} \bm 1^T \bm e_t $, where $\bm h_t$, $\bm m_t$ and $\bm e_t$ are defined in  \eqref{subroutine:critic_update_matrix_form}. 
		
		In the following, let us analyze each component in \eqref{bound: critic descent}. First term A can be expressed below:
		\begin{align}
		2 \beta_t\cdot \mathbb{E}\bigg[ \langle \bar{\omega}_t - \omega^*_t, \bar{h}_t + \bar{m}_t \rangle \bigg] \overset{(a)}{=} 2 \beta_t\cdot \mathbb{E} \bigg[ \langle \bar{\omega}_t - \omega^*_t, \mathbb{E}[\bar{h}_t + \bar{m}_t| \mathcal{F}_t ] \rangle \bigg] \overset{(b)}{=} 2 \beta_t\cdot \mathbb{E} \bigg[ \langle \bar{\omega}_t - \omega^*_t, \bar{h}_t \rangle \bigg] \label{eq:martingale_noise}
		\end{align}
		where $ \mathcal{F}_t $ is the $ \sigma$-algebra generated by $ \mathcal{F}_t = \{ \bomega_{t}, \btheta_{t}, \cdots, \bomega_0, \btheta_0 \} $; $(a)$ follows Tower rule in expectations; $(b)$ is due to the fact that $ \mathbb{E}[m_{i,t} | \mathcal{F}_t] = 0, ~ \forall i \in \mathcal{N} $, which is from \eqref{eq:critic:unbiased_gradient} and \eqref{eq:delta:phi}. 
		 Recall that in \eqref{eq: TD fixed point} we have defined:
		\begin{subequations}
        \label{eq:TD:fixed_point}
       \begin{align}
	    A(\btheta)  :=& ~ \mathbb{E}_{s \sim \mu_{\btheta}(\cdot), s^\prime \sim \mathcal{P}^{\pi_{\btheta}}(\cdot|s) } \left[ \phi(s) \cdot  \big(\phi(s) - \gamma \cdot \phi(s^\prime) \big)^T \right], \; \forall~i \in \mathcal{N}, \\
	    b_{i}(\btheta) :=& ~ \mathbb{E}_{s \sim \mu_{\btheta}(\cdot), \ba \sim \pi_{\btheta}(\cdot|s) } \left[ r_i(s,a) \cdot \phi(s) \right], \; \forall~i \in \mathcal{N}.
	\end{align} 
    \end{subequations}
    Also in \eqref{eq:TD_optimal_condition}, it has shown that $\omega^*_t$ satisfies the condition \begin{align}
        A(\btheta_t) \cdot \omega_t^* = \frac{1}{N} \sum_{i=1}^N b_i(\btheta_t) = \bar{b}(\btheta_t) \label{eq:critic:optimal_condition}
    \end{align}
    where we define $\bar{b}(\btheta_t) := \frac{1}{N} \sum_{i=1}^N b_i(\btheta_t)$. Recall that we have defined 
		\begin{align}
		    \bar{h}_t := \frac{1}{N} \sum_{i=1}^N h_{i,t} = \frac{1}{N} \sum_{i=1}^N \bigg( b_i(\btheta_t) - A(\btheta_t)\cdot \omega_{i,t} \bigg) = \bar{b}(\btheta_t) - A(\btheta_t) \cdot \bar{\omega}_t, \nonumber
		\end{align}
		then it holds that  
		\begin{align}
		    \bar{h}_t &= \bar{b}(\btheta_t) - A(\btheta_t) \cdot \bar{\omega}_t  \overset{\eqref{eq:critic:optimal_condition}}{=} A(\btheta_t) \cdot (\omega^*_t - \bar{\omega}_t) \label{eq:average:h} 
		\end{align}
		Therefore, plugging \eqref{eq:average:h} into \eqref{eq:martingale_noise}, term A in \eqref{bound: critic descent} could be bounded as below:
		\begin{align}
		    2 \beta_t\cdot \mathbb{E} \bigg[ \langle \bar{\omega}_t - \omega^*_t, \bar{h}_t \rangle \bigg] = -2\beta_t \cdot \EE \bigg[ (\bar{\omega}_t - \omega^*_t)^T A(\btheta_t) \cdot (\bar{\omega}_t - \omega^*_t) \bigg] \overset{(i)}{\leq} -2\beta_t \cdot \tilde{c}_{\min} \cdot \EE \bigg[ \| \bar{\omega}_t - \omega^*_t \|^2 \bigg] \label{bound:critic:term_A}
		\end{align}
		where $(i)$ is due to the fact that $A(\btheta_t)$ is a positive definite matrix and its minimum eigenvalue $c_{\min}(A(\btheta_t)) \geq \tilde{c}_{\min}$ in \eqref{def:eigenvalue_A}.
		
		Second, term B can be bounded as below:
		\begin{align}
		& 2\mathbb{E}[\| \omega^*_{t} - \omega^*_{t+1} \|^2] + 2\mathbb{E}\left[ \langle \bar{\omega}_{t} - \omega^*_{t}, \omega^*_{t} - \omega^*_{t+1} \rangle \right] \nonumber \\
		& \overset{(a)}{\leq}  2\mathbb{E}[\| \omega^*_{t} - \omega^*_{t+1} \|^2] + 2\mathbb{E}\left[ \| \bar{\omega}_{t} - \omega^*_{t} \|  \cdot \| \omega^*_{t} - \omega^*_{t+1} \| \right] \nonumber \\
		& \overset{(b)}{\leq}  2L_{\omega}^2 \cdot \mathbb{E}[\| \btheta_{t} - \btheta_{t+1} \|^2] + 2 L_{\omega} \cdot \mathbb{E}[\| \btheta_{t} - \btheta_{t+1} \| \cdot \| \bar{\omega}_{t} - \omega^*_{t} \| ] \label{eq:term_B_derive}
		\end{align}
	where $(a)$ follows Cauchy-Schwarz inequality; $(b)$ is from the Lipschitz property (\ref{eq: lipschitz_critic}) in Lemma \ref{lemma: critic Lipschitz}.
	 Recall the definition $ \btheta_t := \{ \btheta_t^s, \btheta_t^p \} $ and $Q = I -\frac{1}{N}\bm 1 \bm 1^T$, we bound $\| \btheta_t - \btheta_{t+1} \|$ as below:
	 \begin{align}
		\| \btheta_{t} - \btheta_{t+1} \|  &\overset{(i)}{\leq}   \| \btheta^{s}_{t} - \btheta^{s}_{t+1} \| + \| \btheta^{l}_{t} - \btheta^{l}_{t+1} \|  \nonumber \\
		&\overset{(ii)}{\leq} \| \btheta^{s}_{t} - \bm1 \bar{\theta}^{s}_{t} \| +  \| \bm 1 \bar{ \theta}^{s}_{t} - \bm 1 \bar{\theta}^s_{t+1} \| +  \| \bm 1 \bar{ \theta}^{s}_{t+1} - \btheta^s_{t+1} \| + \| \btheta^{l}_{t} - \btheta^{l}_{t+1} \|  \nonumber \\
		&\overset{(iii)}{\leq} \| Q \cdot \btheta^{s}_{t} \| + \| Q\cdot \btheta^{s}_{t+1} \| +  2\alpha_t \cdot \ell_p \nonumber \\
		&\overset{\eqref{subroutine:theta:shared}}{=} \| Q \cdot \btheta^{s}_{t} \| + \| Q\cdot (W_t \cdot \btheta^{s}_{t} + \alpha_t \cdot \bm g_t) \| +  2\alpha_t \cdot \ell_p \nonumber \\
		&\overset{(iv)}{\leq}  \| Q \cdot \btheta^{s}_{t} \| + \| Q\cdot W_t \cdot \btheta^{s}_{t} \| + \alpha_t\cdot \| Q\cdot \bm g_t \| + 2\alpha_t \cdot \ell_p \nonumber \\
		&\overset{(v)}{\leq} \| Q \cdot \btheta^{s}_{t} \| + \| W_t \cdot Q \cdot \btheta^{s}_{t} \|  + 3\alpha_t \cdot \ell_p\nonumber \\
		&\overset{(vi)}{\leq} 2 \| Q \cdot \btheta^{s}_{t} \| + 3\alpha_t \cdot \ell_p. \label{bound:theta:difference}
		\end{align}
	where $(i)$ and $(ii)$ follow Triangle inequality; $ (iii) $ is due to the fact that estimated policy gradient in updating $\btheta_t$ is bounded by $\ell_p$ in \eqref{bound:policy_gradient}; $(iv)$ follows Cauchy-Schwarz inequality; in $(v)$  we used the boudnedness of the gradient   \eqref{bound:policy_gradient}, the fact that  eigenvalue value of $Q$ is upper bounded bounded by $1$, as well as  the following: %{\red[79 only has definition of $\ell_p$, you have other things in (iii), for example where the Q is coming from]}. %Since $ \| \omega_{i,t} \|, \| \lambda_{i,t} \| $ are projected into compact regions with constant radii $ R_{\omega} and R_{\lambda} $, then the gradient norm of policy parameters is bounded by the constant $ \ell_p := N C_{\psi} \bigg(R_{\lambda} + (1+\gamma)R_{\omega} \bigg) $, which is defined in equation (\ref{eq: policy gradient bound}).
		\begin{align}
		    Q\cdot W_t = \left(I - \frac{1}{N}\bm 1 \cdot \bm 1^T\right) \cdot W_t = W_t - \frac{1}{N} \bm 1 \cdot \bm 1^T
		    = W_t\cdot \left(I - \frac{1}{N}\bm 1 \cdot \bm 1^T\right) = W_t \cdot Q \label{eq:W:interchange}.
		\end{align}
	Additionally, $(vi)$ is due to the fact that the eigenvalue of weight matrix $W_t$ is bounded by $1$. Then plugging the inequality \eqref{bound:theta:difference} into \eqref{eq:term_B_derive}, term B could be bounded as follows:
	\begin{align}
		& 2\mathbb{E}[\| \omega^*_{t} - \omega^*_{t+1} \|^2] + 2\mathbb{E}\left[ \langle \bar{\omega}_{t} - \omega^*_{t}, \omega^*_{t} - \omega^*_{t+1} \rangle \right] \nonumber \\
		& \overset{\eqref{eq:term_B_derive}}{\leq}  2L_{\omega}^2 \cdot \mathbb{E}[\| \btheta_{t} - \btheta_{t+1} \|^2] + 2 L_{\omega} \cdot \mathbb{E}[\| \btheta_{t} - \btheta_{t+1} \| \cdot \| \bar{\omega}_{t} - \omega^*_{t} \| ] \nonumber \\
		&\overset{(i)}{\leq} 2L_{\omega}^2 \cdot \bigg( 8\EE\big[ \|Q \cdot \btheta_t^s \|^2 \big] + 18\alpha_t^2 \cdot \ell_p^2 \bigg) + 2L_{\omega} \cdot \EE \bigg[ \big( 2 \| Q \cdot \btheta^{s}_{t} \| + 3\alpha_t \cdot \ell_p \big) \cdot \| \bar{\omega}_t - \omega_t^* \| \bigg] \nonumber \\
		& =  16 L_\omega^2 \cdot \mathbb{E}[ \| Q \cdot \btheta_{t}^{s} \|^2 ] + 36 L_\omega^2 \cdot \ell_p^2 \cdot \alpha_t^2 + 6L_\omega \cdot \ell_p \cdot \alpha_t \cdot \mathbb{E}[\| \bar{\omega}_{t} - \omega^*_{t} \|] \nonumber\\
		& \quad + 4L_\omega \cdot \mathbb{E} \bigg[ \| Q\cdot \btheta_{t}^{s} \| \cdot \| \bar{\omega}_{t} - \omega^*_{t} \|\bigg] \label{bound:term_B}
		\end{align}
		where $(i)$ follows \eqref{bound:theta:difference} and the Cauchy-Schwarz inequality.

		Recall the definition of $ \bar{h}_t $ as $\bar{h}_t := \frac{1}{N} \sum_{i=1}^N h_{i,t} $, where $ \bar{m}_t $ and $ \bar{e}_t $ are also defined similarly. Then term C in \eqref{bound: critic descent} could be bounded as below:
		\begin{align}
		2\mathbb{E}[\| \beta_t (\bar{h}_t + \bar{m}_t) -  \bar{e}_t \|^2] 
		&\overset{(i)}{\leq} 4\beta_t^2 \cdot \mathbb{E}[\| \bar{h}_t + \bar{m}_t \|^2] + 4\mathbb{E}[\| \bar{e}_t \|^2] \nonumber \\
		&= 4\beta_t^2 \cdot \mathbb{E} \bigg[ \big \| \frac{1}{N} \sum_{i=1}^N \big(h_{i,t} + m_{i,t} \big) \big \|^2 \bigg] + 4\mathbb{E} \bigg[ \big \| \frac{1}{N} \sum_{i=1}^N e_{i,t} \big \|^2 \bigg] \nonumber \\
		&\overset{(ii)}{\leq} 4\beta_t^2 \cdot \mathbb{E} \bigg[ \frac{1}{N} \sum_{i=1}^N \big \| h_{i,t} + m_{i,t}  \big \|^2 \bigg] + 4\mathbb{E} \bigg[ \frac{1}{N} \sum_{i=1}^N \big \| e_{i,t} \big \|^2 \bigg] \nonumber \\
	&\overset{(iii)}{\leq} 8\beta_t^2 \cdot L_b^2 \label{bound:term_C}
		\end{align}
		where $(i)$ follows Cauchy-Schwarz inequality; $(ii)$ follows Jensen's inequality; $(iii)$ is from the inequalities \eqref{bound: estimator} and \eqref{bound: projection error}.
		
		Recall that $\bar{e}_t := \frac{1}{N}\sum_{i=1}^N e_{i,t} $ and in \eqref{subroutine:critic_update} we have defined  $ y_{i,t} := \widetilde{\omega}_{i,t} + \beta_{t} \cdot (h_{i,t} + m_{i,t}) $. Then term D in \eqref{bound: critic descent} can be bounded as below:
		\begin{align}
		&-2 \mathbb{E} \bigg[\langle \bar{\omega}_{t} - \omega^*_{t}, \bar{e}_t \rangle \bigg] \nonumber \\
		&= -\frac{2}{N} \sum_{i = 1}^{N} \mathbb{E} \bigg[\langle \bar{\omega}_{t} - \omega^*_{t}, e_{i,t} \rangle \bigg] \nonumber \\
		&= -\frac{2}{N} \sum_{i = 1}^{N} \mathbb{E} \bigg[\langle \bar{\omega}_{t} - y_{i,t}, e_{i,t} \rangle \bigg] - \frac{2}{N} \sum_{i = 1}^{N} \mathbb{E} \bigg[\langle y_{i,t} - \omega^*_t, e_{i,t} \rangle \bigg]  \nonumber \\
		&\overset{(i)}{\leq} \frac{2}{N} \sum_{i = 1}^{N} \mathbb{E} \bigg[ \| \bar{\omega}_{t} - y_{i,t} \| \| e_{i,t} \|\bigg] - \frac{2}{N} \sum_{i = 1}^{N} \mathbb{E} \bigg[\langle y_{i,t} - \omega^*_t, y_{i,t} - \Pi_{R_{\omega}}[y_{i,t}] \rangle \bigg]  \nonumber \\
		&\overset{(ii)}{\leq} \frac{2\beta_t \cdot L_b}{N} \sum_{i = 1}^{N} \EE \bigg[ \| \bar{\omega}_{t} - y_{i,t} \| \bigg] - \frac{2}{N} \sum_{i = 1}^{N} \mathbb{E} \bigg[ \| y_{i,t} - \Pi_{R_{\omega}}[y_{i,t}] \|^2 \bigg]  \nonumber \\
		&\overset{(iii)}{\leq} \frac{2 \beta_t \cdot L_b }{N} \sum_{i = 1}^{N} \mathbb{E} \bigg[ \| \bar{\omega}_{t} - \sum_{j = 1}^{N} W_t^{ij} \omega_{j,t} - \beta_{t} \cdot (h_{i,t} + m_{i,t}) \| \bigg] - \frac{2}{N}\sum_{i = 1}^{N} \mathbb{E} \bigg[\| e_{i,t} \|^2 \bigg] \nonumber \\
		&\overset{(iv)}{\leq} \frac{2 \beta_t \cdot L_b}{N} \sum_{i = 1}^{N} \mathbb{E} \bigg[ \| \bar{\omega}_{t} - \sum_{j = 1}^{N} W_t^{ij} \omega_{j,t} \| \bigg] + \frac{2 \beta_t \cdot L_b}{N} \sum_{i = 1}^{N} \mathbb{E} \bigg[\| \beta_t(h_{i,t} + m_{i,t}) \| \bigg] - \frac{2}{N}\sum_{i = 1}^{N} \mathbb{E} \bigg[\| e_{i,t} \|^2 \bigg] \nonumber \\
		&\overset{(v)}{\leq} \frac{2 \beta_t \cdot L_b}{N} \sum_{i = 1}^{N} \mathbb{E} \bigg[ \| \bar{\omega}_{t} - \omega_{i,t} \| \bigg] + 2 \beta_t^2 \cdot L_b^2 - \frac{2}{N}\sum_{i = 1}^{N} \mathbb{E} \bigg[\| e_{i,t} \|^2 \bigg] \nonumber \\
		&\overset{(vi)}{\leq} 2 \beta_t \cdot L_b \cdot \mathbb{E} \bigg[ \| Q \cdot \omega_{t} \| \bigg] + 2 \beta_t^2 \cdot L_b^2 - \frac{2}{N}\sum_{i = 1}^{N} \mathbb{E} \bigg[\| e_{i,t} \|^2 \bigg] \label{bound:critic:term_D}
		\end{align}
		where $(i)$ follows Cauchy-Schwarz inequality and the definition $e_{i,t}:= y_{i,t} - \Pi_{R_{\omega}}[y_{i,t}] $; $(ii)$ is from \eqref{bound: projection error} and the projection property \eqref{bound: nedic 1} in Lemma \ref{lemma: projection}; $(iii)$ follows the definition of $y_{i,t}$; $(iv)$ follows Cauchy-Schwarz inequality; $(v)$ is from the inequality \eqref{bound: estimator} and that the eigenvalues of $W_t$ is bounded by $1$; $(vi)$ is due to the fact that $ \| \bar{\omega}_{t} - \omega_{i,t} \| \leq \sqrt{ \sum_{i=1}^N \| \bar{\omega}_{t} - \omega_{i,t} \|^2 } = \| Q \cdot \omega_t \| ~ \forall ~ i  \in \mathcal{N}$. 
		
		Then we plug in the above derived bounds on terms A-D (inequalities \eqref{bound:critic:term_A}, \eqref{bound:term_B}, \eqref{bound:term_C} and \eqref{bound:critic:term_D}) into \eqref{bound: critic descent}, and obtain:
		\begin{align}
		&\mathbb{E} \bigg[ \| \bar{\omega}_{t+1} - \omega^*_{t+1} \|^2 \bigg] \nonumber \\
		&\leq (1 - 2\beta_t \cdot \tilde{c}_{\min}) \cdot \mathbb{E} \bigg[ \| \bar{\omega}_{t} - \omega^*_{t} \|^2 \bigg] +  16 L_\omega^2 \cdot \mathbb{E} \bigg[ \| Q \cdot \btheta^{s}_{t} \|^2 \bigg] + 36 L_\omega^2 \cdot \ell_p^2 \cdot \alpha_t^2 + 6L_\omega \cdot \ell_p \cdot \alpha_t \cdot \mathbb{E} \bigg[\| \bar{\omega}_{t} - \omega^*_{t} \|\bigg]  \nonumber \\
		& \quad + 4L_\omega \cdot \mathbb{E}\bigg[ \| Q \cdot \btheta^{s}_{t} \| \cdot \| \bar{\omega}_{t} - \omega^*_{t} \|\bigg] + 2\beta_t \cdot L_b \cdot \mathbb{E} \bigg[ \| Q \cdot \bomega_{t} \| \bigg] + 10\beta_t^2\cdot L_b^2 - \frac{2}{N}\sum_{i = 1}^{N} \mathbb{E} \bigg[\| e_{i,t} \|^2 \bigg] \label{bound:descent:critic}.
		\end{align}

	%	{\red[next what are you planning to do?][if you trace back, you have done your first step, what's your second step? ]}
		
		In \eqref{bound:descent:critic}, we have already obtain a bound of the distance between averaged parameter $ \bar{\omega}_{t} $ and its optimal solution $\omega_{t}^*$. Then we could utilize the inequality \eqref{bound:descent:critic} to derive the convergence rate for the averaged parameters $ \bar{\omega}_t $.
		Towards this end, we first rearrange the inequality \eqref{bound:descent:critic}, divide both sides by $ 2\beta_t \cdot \tilde{c}_{\min} $ and then sum it from $ t = 0 \text{ to } T-1$, and obtain: 
		\begin{align}
		&\sum_{t = 0}^{T-1} \mathbb{E}\bigg[ \| \bar{\omega}_{t} - \omega^*_t \|^2 \bigg] \nonumber \\
		\leq& \underbrace{\sum_{t = 0}^{T-1} \frac{1}{ 2\beta_t \cdot \tilde{c}_{\min}} \bigg( \mathbb{E}\bigg[\| \omega^*_t - \bar{\omega}_t \|^2 \bigg] - \mathbb{E}\bigg[\| \omega^*_{t+1} - \bar{\omega}_{t+1} \|^2 \bigg]  \bigg)}_{\rm term ~ I_1} + \underbrace{\frac{8L_\omega^2 }{\tilde{c}_{\min}} \cdot \sum_{t = 0}^{T-1} \frac{1}{\beta_t} \cdot \mathbb{E} \bigg[ \| Q \cdot \btheta^{s}_{t} \|^2 \bigg] }_{\rm term ~ I_2}  \nonumber \\
		&+ \underbrace{ \sum_{t = 0}^{T-1} \frac{1}{\tilde{c}_{\min}}  \bigg( 18L_w^2 \cdot \ell_p^2 \cdot \frac{\alpha_t^2}{\beta_t} + 5L_b^2 \cdot \beta_t  \bigg) }_{\rm term ~ I_3} + \underbrace{ \frac{3L_{\omega} \cdot \ell_p}{\tilde{c}_{\min}} \sum_{t = 0}^{T-1} \frac{\alpha_t}{\beta_t} \cdot \mathbb{E}\bigg[\| \bar{\omega}_t - \omega_t^* \|\bigg] }_{\rm term ~ I_4} \nonumber \\
		&  +  \underbrace{\frac{2L_\omega}{\tilde{c}_{\min}} \cdot \sum_{t = 0}^{T-1} \frac{1}{\beta_t} \cdot \mathbb{E} \bigg[ \| Q \cdot \btheta^{s}_{t} \| \cdot \| \bar{\omega}_{t} - \omega^*_{t} \| \bigg]}_{\rm term ~ I_5} + \underbrace{  \frac{L_b}{\tilde{c}_{\min}} \cdot \sum_{t = 0}^{T-1} \EE \bigg[ \|Q \cdot \bomega_t \| \bigg]}_{\rm term ~ I_6}   
		 \label{sum:bound:critic_descent}.
		\end{align}
		
		 Recall the fixed stepsizes  $ \alpha_t := \frac{\alpha}{T^{\sigma_1}} $ and $  \beta_t := \frac{\beta}{T^{\sigma_2}} ~ \forall t$.  We can bound each term in \eqref{sum:bound:critic_descent} as below.
		First, term $I_1$ can be bounded as:
		\begin{align}
		I_1 &:=  \sum_{t = 0}^{T-1} \frac{T^{\sigma_2}}{ 2\beta \cdot \tilde{c}_{\min}} \cdot \bigg( \mathbb{E}\bigg[\| \omega^*_t - \bar{\omega}_t \|^2 \bigg] - \mathbb{E}\bigg[\| \omega^*_{t+1} - \bar{\omega}_{t+1} \|^2 \bigg]  \bigg) \nonumber \\
		&= \frac{T^{\sigma_2}}{ 2\beta \cdot \tilde{c}_{\min}} \cdot \bigg( \mathbb{E}\bigg[\| \omega^*_0 - \bar{\omega}_0 \|^2 \bigg] - \mathbb{E}\bigg[\| \omega^*_{T} - \bar{\omega}_{T} \|^2 \bigg]  \bigg) \nonumber \\
		& \overset{(a)}{\leq} \frac{T^{\sigma_2} \cdot 4R_{\omega}^2}{ 2\beta \cdot \tilde{c}_{\min}}  = \mathcal{O}(T^{\sigma_2})  \label{bound:critic:term_1}
		\end{align}
		where the inequality $(a)$ follows $ \mathbb{E}\big[\| \omega^*_t - \bar{\omega}_t \|^2\big] \le 2\mathbb{E}\big[\| \omega^*_t \|^2 + \| \bar{\omega}_t \|^2\big] = 4R_{\omega}^2$, since both $\omega^*_t$ and $ \bar{\omega}_t $ are in a fixed region with radius $R_{\omega}$. 
		
		Second, term $I_2$ in \eqref{sum:bound:critic_descent} can be bounded as:
		\begin{align}
			I_2 & := \frac{8L_\omega^2 }{\tilde{c}_{\min}} \cdot \sum_{t = 0}^{T-1} \frac{1}{\beta_t}\cdot  \mathbb{E} \bigg[ \| Q \cdot \btheta^{s}_{t} \|^2 \bigg] \overset{(i)}{=} \frac{8L_\omega^2 \cdot T^{\sigma_2} }{\tilde{c}_{\min} \cdot \beta} \cdot \sum_{t = 0}^{T-1}  \mathbb{E}\bigg[ \| Q \cdot \btheta^{s}_t \|^2 \bigg] \overset{\eqref{eq:policy_consensus_square}}{=} \mathcal{O}(T^{ 1 - 2\sigma_1 + \sigma_2}) \label{bound:critic:term_2}
		\end{align}
		where $(i)$ follows the definition $\beta_t = \frac{\beta}{T^{\sigma_2}}, ~ \forall t$.
		
		Third, term $I_3$ in  \eqref{sum:bound:critic_descent} can be bounded as:
		\begin{align}
		I_3 &:= \sum_{t = 0}^{T-1} \frac{1}{\tilde{c}_{\min}} \cdot  \bigg( 18L_w^2 \cdot \ell_p^2 \cdot \frac{\alpha_t^2}{\beta_t} + 5L_b^2 \cdot \beta_t  \bigg) \nonumber \\
		& \overset{(a)}{=} \frac{18L_w^2 \cdot \ell_p^2}{\tilde{c}_{\min}}  \cdot \frac{\alpha^2}{\beta} \cdot T \cdot T^{\sigma_2 - 2\sigma_1} + \frac{5L_b^2 \cdot \beta}{\tilde{c}_{\min}} \cdot T \cdot T^{-\sigma_2} \nonumber \\
		&= \frac{18L_w^2 \cdot \ell_p^2}{\tilde{c}_{\min}}  \cdot \frac{\alpha^2}{\beta} \cdot T^{1 + \sigma_2 - 2\sigma_1} + \frac{5L_b^2 \cdot \beta}{\tilde{c}_{\min}} \cdot T^{1 - \sigma_2} \nonumber \\
		&=  \mathcal{O}(T^{1 - \sigma_2}) + \mathcal{O} \left( T^{1 + \sigma_2 - 2\sigma_1} \right) \label{bound:critic:term_3}
		\end{align}
		where $(a)$ follows the definitions $ \alpha_t := \frac{\alpha}{T^{\sigma_1}} $ and $  \beta_t := \frac{\beta}{T^{\sigma_2}}$ for any iteration $t$.
		
		Next, we can bound the term $I_4$ in  \eqref{sum:bound:critic_descent} as below: 
		\begin{align}
		I_4 &= \frac{3L_{\omega} \cdot \ell_p}{\tilde{c}_{\min}} \cdot \sum_{t = 0}^{T-1} \frac{\alpha_t}{\beta_t} \cdot \mathbb{E}\bigg[\| \bar{\omega}_t - \omega_t^* \|\bigg] \nonumber \\
	%	&= \frac{3L_{\omega} \cdot \ell_p}{\tilde{c}_{\min}} \cdot \sum_{t = 0}^{T-1} \sqrt{\frac{\alpha_t^2}{\beta_t^2}} \cdot \sqrt{\left( \mathbb{E} \bigg[\| \bar{\omega}_t - \omega_t^* \| \bigg] \right)^2} \nonumber \\
		&\overset{(i)}{\leq} \frac{3L_{\omega} \cdot \ell_p}{\tilde{c}_{\min}} \cdot \sum_{t = 0}^{T-1} \bigg( \sqrt{\frac{\alpha_t^2}{\beta_t^2}} \cdot \sqrt{ \mathbb{E} \bigg[\| \bar{\omega}_t - \omega_t^* \|^2 \bigg]} \bigg) \nonumber \\
		&\overset{(ii)}{\leq} \frac{3L_{\omega} \cdot \ell_p}{\tilde{c}_{\min}} \cdot \sqrt{\sum_{t = 0}^{T-1} \frac{\alpha_t^2}{\beta_t^2}} \cdot \sqrt{\sum_{t = 0}^{T-1} \mathbb{E} \bigg[\| \bar{\omega}_t - \omega_t^* \|^2 \bigg]} \nonumber \\
		&\overset{(iii)}{=} \frac{3L_{\omega} \cdot \ell_p \cdot \alpha^2_0}{\tilde{c}_{\min} \cdot \beta^2_0} \cdot  \sqrt{T \cdot T^{2\sigma_2 - 2\sigma_1} } \cdot \sqrt{\sum_{t = 0}^{T-1} \mathbb{E} \bigg[\| \bar{\omega}_t - \omega_t^* \|^2 \bigg]} \nonumber \\
		&=  \frac{3L_{\omega} \cdot \ell_p \cdot \alpha^2_0}{\tilde{c}_{\min} \cdot \beta^2_0} \cdot  \sqrt{ T^{1 + 2\sigma_2 - 2\sigma_1} } \cdot \sqrt{\sum_{t = 0}^{T-1} \mathbb{E} \bigg[\| \bar{\omega}_t - \omega_t^* \|^2 \bigg]} \nonumber \\
		&\overset{(iv)}{=} \sqrt{C_4 \cdot T^{1 + 2\sigma_2 - 2\sigma_1}} \cdot \sqrt{\sum_{t = 0}^{T-1} \mathbb{E} \bigg[\| \bar{\omega}_t - \omega_t^* \|^2 \bigg]} \label{bound:critic:term_4}
		\end{align}
		where $(i)$ is by Jensen's inequality; $(ii)$ follows Cauchy-Schwarz inequality; $(iii)$ follows the definition of stepsizes $\alpha_t$ and $\beta_t$; in equality $(iv)$ we define $C_4 := \bigg( \frac{3L_{\omega} \cdot \ell_p \cdot \alpha^2_0}{\tilde{c}_{\min} \cdot \beta^2_0} \bigg)^2 $.
		
		Next, the term $I_5$ in \eqref{sum:bound:critic_descent} can be bounded as below:
		\begin{align}
			I_5 &:=  \frac{2L_\omega}{\tilde{c}_{\min}} \cdot \sum_{t = 0}^{T-1} \frac{1}{\beta_t} \cdot \mathbb{E} \bigg[ \| Q \cdot \btheta^{s}_{t} \| \cdot \| \bar{\omega}_{t} - \omega^*_{t} \| \bigg]  \nonumber \\
			&\overset{(i)}{\leq} \frac{2L_\omega}{\tilde{c}_{\min}} \cdot \sum_{t = 0}^{T-1} \sqrt{ \frac{1}{\beta_t^2} \mathbb{E}\bigg[  \| Q \cdot \btheta^{s}_t \|^2\bigg] \cdot \mathbb{E}\bigg[\| \bar{\omega}_{t} - \omega^*_{t} \|^2\bigg]}  \nonumber \\
			&\overset{(ii)}{\leq} \frac{2L_\omega}{\tilde{c}_{\min}} \cdot \sqrt{\sum_{t = 0}^{T-1} \frac{1}{\beta_t^2} \mathbb{E}\bigg[  \| Q \cdot \btheta^{s}_t \|^2\bigg]} \cdot \sqrt{\sum_{t = 0}^{T-1} \mathbb{E} \bigg[\| \bar{\omega}_{t} - \omega^*_{t} \|^2 \bigg]} \nonumber \\
			&\overset{(iii)}{=} \sqrt{ \frac{C_5 \cdot T^{2\sigma_2}}{\beta^2} \sum_{t = 0}^{T-1} \mathbb{E}\bigg[ \| Q\cdot \btheta^{s}_t \|^2\bigg] } \cdot \sqrt{\sum_{t = 0}^{T-1} \mathbb{E}[\| \bar{\omega}_{t} - \omega^*_{t} \|^2]}  \label{bound:critic:term_5}
			\end{align}
		where $(i)$ and $(ii)$ follows Cauchy-Schwarz inequality; $(iii)$ follows the stepsize $\beta_t := \frac{\beta}{T^{\sigma_{2}}} $ and we define the constant $ C_5 := \big(\frac{2L_{\omega}}{\tilde{c}_{\min}}\big)^2 $.
		
		Finally, we can bound the last term $I_6$ in \eqref{sum:bound:critic_descent} as below:
		\begin{align}
		I_6 :=& \frac{L_b}{\tilde{c}_{\min}} \cdot \sum_{t = 0}^{T-1} \EE \bigg[ \|Q \cdot \bomega_t \| \bigg] \overset{\eqref{rate:consensus_error:critic}}{=} \mathcal{O}\bigg( T^{1 - \sigma_2} \bigg). \label{bound:critic:term_6}
		\end{align}
		Then we can revisit \eqref{sum:bound:critic_descent} to obtain the exact convergence rate. Let us rearrange \eqref{sum:bound:critic_descent} as below:
		\begin{align}
		\sum_{t = 0}^{T-1} \mathbb{E}\bigg[ \| \bar{\omega}_{t} - \omega^*_t \|^2 \bigg] &\leq \underbrace{(I_1 + I_2 + I_3 + I_6)}_{\rm := term ~ K_1} + (I_4 + I_5). \label{bound:critic_gap:rearrange}
		\end{align}
		For the terms $I_4 + I_5$, we utilize \eqref{bound:critic:term_4} and \eqref{bound:critic:term_5} to obtain that
		\begin{align}
		    I_4 + I_5 &\leq \bigg( \sqrt{C_4 \cdot T^{1 + 2\sigma_2 - 2\sigma_1}} + \sqrt{ C_5 \cdot \frac{T^{2\sigma_{2}}}{\beta^2} \sum_{t = 0}^{T-1} \mathbb{E}\bigg[ \| Q\cdot \btheta^{s}_t \|^2\bigg] } \bigg) \cdot \sqrt{\sum_{t = 0}^{T-1} \mathbb{E}[\| \bar{\omega}_{t} - \omega^*_{t} \|^2]} \nonumber \\
		    & \overset{(a)}{\leq} \sqrt{ \underbrace{2C_4 \cdot T^{1 + 2\sigma_2 - 2\sigma_1} + \frac{2C_5 \cdot T^{2\sigma_2}}{\beta^2} \sum_{t = 0}^{T-1} \mathbb{E}\bigg[ \| Q\cdot \btheta^{s}_t \|^2\bigg]}_{\rm :=term ~ K_2}  }  \cdot \sqrt{ \underbrace{\sum_{t = 0}^{T-1} \mathbb{E}[\| \bar{\omega}_{t} - \omega^*_{t} \|^2]}_{\rm :=term ~ K_3}} \label{bound:double_sqrt}
		\end{align}
		where $(a)$ follows Cauchy-Schwarz inequality.
		
		%Finally, we are able to analyze the convergence rate of $ \frac{1}{T} \sum_{t = 0}^{T-1} \mathbb{E}\big[\| \bar{\omega}_{t} - \omega^*_{t} \|^2\big] $. 
		With terms $K_1, K_2$ and $K_3$ defined as above, we can plug \eqref{bound:double_sqrt} into \eqref{bound:critic_gap:rearrange}, and obtain:
		\begin{align}
		&K_3 \leq K_1 + \sqrt{K_2\cdot K_3} \implies \big( \sqrt{K_3} - \frac{1}{2} \sqrt{K_2} \big)^2 \leq K_1 + \frac{1}{4}K_2 \implies \sqrt{K_3} \leq \frac{1}{2} \sqrt{K_2} + \sqrt{K_1 + \frac{1}{4}K_2} \nonumber \\
		& K_3 \leq \bigg( \frac{1}{2} \sqrt{K_2} + \sqrt{K_1 + \frac{1}{4}K_2} \bigg)^2 \overset{(a)}{\leq} \frac{1}{2}K_2 + 2K_1 + \frac{1}{2}K_2 = 2K_1 + K_2  \label{rate:compute:critic}
		\end{align}
		where $(a)$ is due to Cauchy-Schwarz inequality.
		
		 Combining the inequalities \eqref{bound:critic:term_1}, \eqref{bound:critic:term_2}, \eqref{bound:critic:term_3} and \eqref{bound:critic:term_6}, the convergence rate of term $K_1$ in \eqref{rate:compute:critic} could be expressed as below:
		 \begin{align}
		     K_1 &:= I_1 + I_2 + I_3 + I_6 \nonumber \\
		     &= \mathcal{O}\big(T^{\sigma_2}\big) + \mathcal{O}\big(T^{1 - 2\sigma_1 + \sigma_2}\big) + \mathcal{O}\big( T^{1-\sigma_2} \big) + \mathcal{O}\big(T^{1 + \sigma_2 - 2\sigma_1 }\big) + \mathcal{O}\big(T^{1 - \sigma_2}\big). \label{rate:term:K_1}
		 \end{align}
		 
		 Moreover, the convergence rate of term $K_2$ could be bounded as below:
		 \begin{align}
		     K_2 &:= 2C_4 \cdot T^{1 + 2\sigma_2 - 2\sigma_1} + \frac{2C_5 \cdot T^{2\sigma_2}}{\beta^2} \sum_{t = 0}^{T-1} \mathbb{E}\bigg[ \| Q\cdot \btheta^{s}_t \|^2\bigg] \nonumber \\
		     & \overset{(a)}{=} \mathcal{O}\big( T^{1 + 2\sigma_2 - 2\sigma_1} \big) + \mathcal{O}\big( T^{-1 + 2\sigma_2 } \big) \label{rate:term:K_2}
		 \end{align}
		 where $(a)$ follows the inequality \eqref{eq:policy_consensus_square}.
		
		Plugging \eqref{bound:double_sqrt} - \eqref{rate:term:K_2} into \eqref{bound:critic_gap:rearrange}, dividing $T$ on both sides, we can obtain the convergence rate of $\frac{1}{T} \sum_{t = 0}^{T-1} \mathbb{E}\left[ \| \bar{\omega}_{t} - \omega^*_t \|^2 \right]$ as below:
		\begin{align}
		\frac{1}{T} \sum_{t = 0}^{T-1} \mathbb{E}\bigg[\| \bar{\omega}_{t} - \omega^*_{t} \|^2\bigg] = \mathcal{O}(T^{-1+\sigma_2}) + \mathcal{O}(T^{ - \sigma_2}) + \mathcal{O} \left( T^{\sigma_2 - 2\sigma_1} \right) + \mathcal{O}(T^{-2\sigma_1 + 2\sigma_2}) + \mathcal{O}(T^{-2 + 2\sigma_2}). \label{order:critic_gap}
		\end{align}
		%where the last two terms above are related to the consensus error in the shared part of policy parameters. {\red[it seems that the above equation is not cited anywhere; if not, then we can remove its equation number.]}
		
		When the fixed stepsizes $ \zeta_t $ and $ \beta_t $ are in the same order ($\zeta_t := \frac{\zeta}{T^{\sigma_2}}$ and $\beta_t := \frac{\beta}{T^{\sigma_2}}$ for any iteration $t$), the analysis of parameters $ \bomega_{t} $ can directly extend to establish the convergence results of parameters $ \blambda_{t} $. 	Then, it holds that $ \frac{1}{T} \sum_{t = 0}^{T-1} \mathbb{E}\big[\| \bar{\lambda}_{t} - \lambda^*_{t} \|^2\big] $ has the same convergence rate as $ \frac{1}{T} \sum_{t = 0}^{T-1} \mathbb{E}\big[\| \bar{\omega}_{t} - \omega^*_{t} \|^2\big] $. 
		
		 Finally, combining the convergence results for consensus errors in \eqref{bound:lower:consensus_errors}, we obtain the convergence of approximation parameters as below:
		\begin{align}
		& \frac{1}{T} \sum_{t = 0}^{T-1} \sum_{i=1}^N \bigg( \mathbb{E}\bigg[\| \omega_{i,t} - \omega^*_{t} \|^2\bigg] + \mathbb{E}\bigg[\| \lambda_{i,t} - \lambda^*_{t} \|^2\bigg] \bigg) \nonumber \\
		& \overset{(a)}{\leq} \frac{1}{T} \sum_{t = 0}^{T-1} \sum_{i=1}^N \bigg( 2\mathbb{E}\bigg[\| \omega_{i,t} - \bar{\omega}_{t} \|^2\bigg] + 2\mathbb{E}\bigg[\| \bar{\omega}_{t} - \omega^*_{t} \|^2\bigg] + 2\mathbb{E}\bigg[\| \lambda_{i,t} - \bar{\lambda}_{t} \|^2\bigg] + 2\mathbb{E}\bigg[\| \bar{\lambda}_{t} - \lambda^*_{t} \|^2\bigg] \bigg) \nonumber \\
		&= \frac{2}{T} \sum_{t = 0}^{T-1} \sum_{i=1}^N \bigg( \mathbb{E}\bigg[\|  \bar{\omega}_{t} - \omega^*_t \|^2\bigg] + \mathbb{E}\bigg[\| \bar{\lambda}_t - \lambda^*_t \|^2\bigg] \bigg) + \frac{2}{T} \sum_{t = 0}^{T-1} \bigg( \EE \bigg[ \| Q \cdot \bomega_t \|^2 \bigg] + \EE \bigg[ \| Q \cdot \blambda_t \|^2 \bigg] \bigg) \nonumber \\
		&\overset{(b)}{=} \mathcal{O}(T^{-1+\sigma_2}) + \mathcal{O}(T^{ - \sigma_2}) + \mathcal{O} \left( T^{\sigma_2 - 2\sigma_1} \right) + \mathcal{O}(T^{-2\sigma_1 + 2\sigma_2}) + \mathcal{O}(T^{-2 + 2\sigma_2}) + \mathcal{O}(T^{-2\sigma_2}). \label{rate:lower:parameters}
		\end{align}
		where $(a)$ follows Cauchy-Schwarz inequality; $(b)$ follows \eqref{rate:lower:consensus:square} and \eqref{order:critic_gap}. % and the fact that $ \frac{1}{T} \sum_{t = 0}^{T-1} \mathbb{E}\big[\| \bar{\lambda}_{t} - \lambda^*_{t} \|^2\big] $ has the same convergence rate as $ \frac{1}{T} \sum_{t = 0}^{T-1} \mathbb{E}\big[\| \bar{\omega}_{t} - \omega^*_{t} \|^2\big] $. 
	\end{proof}
	
	\section{Proof of Proposition \ref{proposition:actor}} \label{appendix:actor_convergence}
	%In Theorem \ref{thm: actor}, we first present the convergence analysis of actor parameters with single sampling procedure in part A. Then in part B, we present the analysis results of proposed algorithm with double sampling procedures.

%	{\red[local policy is $p$, right?]}
	\begin{proof} In this proposition, we will analyze the convergence of actor in algorithm \aname.
	
	With linear approximations, recall that in \eqref{eq:local:td} we have defined: 
	$$\widehat{\delta}_{i,t} = \varphi(s_t, \bm a_t)^T \lambda_{i,t} + \gamma\cdot \phi(s_{t+1})^T \omega_{i,t} - \phi(s_t)^T \omega_{i,t}. $$ 
	Due to the facts that feature vectors are assumed to be bounded in Assumption \ref{ass:feature_bound} and the approximation parameters $\lambda_{i,t}$ and $\omega_{i,t}$ are restricted in fixed regions, we have
	\begin{align}
	  \| \widehat{\delta}_{i,t} \|\le    R_\lambda + (1+\gamma)R_{\omega}, \; \forall~i\in \mathcal{N}
	\end{align}
	For simplicity, let us define $C_{\delta}:=R_\lambda + (1+\gamma)R_{\omega}$.
% 	define a constant $ :=  \geq $ for all $i \in \mathcal{N}$ and any iteration $t$. 
	
	Recall that for each agent $ i $, we have denoted its local policy parameters as $ \theta_i := \{ \theta_i^s, \theta_i^p \} $ where $ \theta_i^s $ is the shared policy parameter and $ \theta_i^p $ is the personalized policy parameter. Moreover, for each policy optimization step, the update in shared policy parameters $ \theta_i^s $ is given by:
	\begin{align}
	    \theta_{i,t+1}^{s} := \sum_{j = 1}^{N} W_{ij}\cdot \theta^s_{j,t} + \alpha_t \cdot \widehat{\delta}_{i,t} \cdot \psi_{\theta^{s}_i}(s_t, a_{i,t}) \label{update:theta:shared}
	\end{align}
	where we have defined the score function $ \psi_{\theta^{s}_i}(s_t, a_{i,t}) := \nabla_{\theta^s} \log \pi(a_{i,t} \mid s_t,  \theta_{i,t}) $. Therefore, the update of the {\it average} of the shared policy parameter $ \bar{\theta}^s_t := \frac{1}{N} \sum_{i = 1}^{N} \theta^s_{i,t} $ is given below:
	\begin{align}
	    \bar{\theta}^s_{t+1} := \bar{\theta}^s_t + \frac{\alpha_t}{N} \sum_{i = 1}^{N} \widehat{\delta}_{i,t} \cdot \psi_{\theta^{s}_i}(s_t, a_{i,t}).
	\end{align}
	We further define $ \bar{\btheta}_t := \cup_{i=1}^N \{ \bar{\theta}_{i,t} \}$ and $ \bar{\theta}_{i,t} := \{ \bar{\theta}^s_t, \theta^{p}_{i,t} \}$. Also recall we have defined $ \btheta_t := \cup_{i=1}^N \{ \theta_{i,t} \}$ and $ \theta_{i,t} := \{ \theta^s_{i,t}, \theta^{p}_{i,t} \}$. In the analysis, we start from analyzing objective value $J(\bar{\btheta}_t)$ at each iteration $t$.
	According to $ L_J $-Lipschitz of policy gradient in Lemma \ref{lemma: Lipschitz Objective}, it holds that 
		\begin{align}
		& \quad J(\bar{\btheta}_{t+1}) \nonumber \\ &\overset{\eqref{eq: lipschitz objective}}{\geq} J(\bar{ \btheta}_{t}) + \langle \nabla J(\bar{ \btheta}_{t}), \bar{ \btheta}_{t+1} - \bar{ \btheta}_{t} \rangle - \frac{L_J}{2} \| \bar{ \btheta}_{t+1} - \bar{ \btheta}_{t} \|^2 \nonumber \\
		&= J(\bar{ \btheta}_{t}) + \langle \nabla J(\bar{ \btheta}_{t}) - \nabla J( \btheta_{t}), \bar{ \btheta}_{t+1} - \bar{ \btheta}_{t} \rangle + \langle \nabla J( \btheta_{t}), \bar{ \btheta}_{t+1} - \bar{ \btheta}_{t} \rangle - \frac{L_J}{2} \| \bar{ \btheta}_{t+1} - \bar{ \btheta}_{t} \|^2 \nonumber \\
		&= J(\bar{ \btheta}_{t}) + \sum_{i=1}^N \langle \nabla_{\theta_i} J(\bar{ \btheta}_{t}) - \nabla_{\theta_i} J( \btheta_{t}), \bar{ \theta}_{i,t+1} - \bar{ \theta}_{i,t} \rangle + \sum_{i=1}^N \langle \nabla_{\theta_i} J( \btheta_{t}), \bar{ \theta}_{i,t+1} - \bar{ \theta}_{i,t} \rangle - \frac{L_J}{2}\sum_{i = 1}^{N}\| \bar{\theta}_{i,t+1} - \bar{\theta}_{i,t} \|^2  \nonumber \\
		&\geq J(\bar{ \btheta}_{t}) - \sum_{i=1}^N \| \nabla_{\theta_i} J(\bar{ \btheta}_{t}) - \nabla_{\theta_i} J( \btheta_{t}) \| \cdot \| \bar{ \theta}_{i,t+1} - \bar{ \theta}_{i,t} \| + \sum_{i=1}^N \langle \nabla_{\theta_i} J( \btheta_{t}), \bar{ \theta}_{i,t+1} - \bar{ \theta}_{i,t} \rangle - \frac{L_J}{2}\sum_{i = 1}^{N}\| \bar{\theta}_{i,t+1} - \bar{\theta}_{i,t} \|^2 \nonumber \\
		& \overset{(i)}{\geq} J(\bar{ \btheta}_{t}) - N \cdot \alpha_t \cdot \ell_p \cdot L_J \cdot \| \bar{ \btheta}_{t} -  \btheta_{t} \| + \sum_{i=1}^N \langle \nabla_{\theta_i} J( \btheta_{t}), \bar{ \theta}_{i,t+1} - \bar{ \theta}_{i,t} \rangle - \frac{\alpha_t^2 \cdot \ell_p^2 \cdot L_J \cdot N}{2} \nonumber \\
		& \overset{(ii)}{=} J(\bar{ \btheta}_{t}) - N \cdot \alpha_t \cdot \ell_p \cdot L_J \cdot \| \btheta_t^s - \bm 1 \cdot \bar{ \theta}^{s^T}_t \| + \sum_{i = 1}^{N} \langle \nabla_{\theta_i} J(\btheta_{t}), \bar{\theta}_{i,t+1} - \bar{\theta}_{i,t} \rangle - \frac{\alpha_t^2 \cdot \ell_p^2 \cdot L_J \cdot N}{2} \label{obj:bound:bar_theta}
		\end{align}
		where $(i)$ follows \eqref{eq: lipschitz objective} and $ \| \bar{ \theta}_{i,t+1} - \bar{ \theta}_{i,t} \| \leq \alpha_t \cdot \ell_p $  where the constant $\ell_p$ is defined in \eqref{bound:policy_gradient}; $(ii)$ follows the fact that $\| \bar{\btheta}_t - \btheta_t \| = \| \btheta_t^s - \bm 1 \cdot \bar{ \theta}^{s^T}_t \|$ since we have defined $ \bar{\btheta}_t := \cup_{i=1}^N \{ \bar{\theta}_{i,t} \} = \cup_{i=1}^N \{ \bar{\theta}_{t}^{s}, \theta^{p}_{i,t} \} $.
		
		Therefore, it follows that
		\begin{align}
		& J(\bar{\btheta}_{t+1}) \nonumber \\
		& \overset{\eqref{obj:bound:bar_theta}}{\geq} J(\bar{ \btheta}_{t}) - N \cdot \alpha_t \cdot \ell_p \cdot L_J \cdot \| \btheta_t^s - \bm 1 \cdot \bar{ \theta}^{s^T}_t \| + \sum_{i = 1}^{N} \langle \nabla_{\theta_i} J(\btheta_{t}), \bar{\theta}_{i,t+1} - \bar{\theta}_{i,t} \rangle - \frac{\alpha_t^2 \cdot \ell_p^2 \cdot L_J \cdot N}{2} \nonumber \\
		& \overset{(a)}{=} J(\bar{ \btheta}_{t}) - N \cdot \alpha_t \cdot \ell_p \cdot L_J \cdot \| Q \cdot \btheta^{s}_t \| + \alpha_t \sum_{i = 1}^{N} \bigg \langle \nabla_{\theta_i^s} J( \btheta_{t}), \frac{1}{N} \sum_{j = 1}^{N}  \widehat{\delta}_{j,t} \cdot \psi_{\theta^{s}_j}(s_t, a_{j,t}; \theta_{j,t}) \bigg \rangle \nonumber \\
		& \quad + \alpha_t \sum_{i = 1}^{N} \bigg \langle \nabla_{\theta_i^p} J(\btheta_{t}), \widehat{\delta}_{i,t} \cdot \psi_{\theta^{p}_i}(s_t, a_{i,t};  \theta_{i,t}) \bigg \rangle - \frac{\alpha_t^2 \cdot \ell_p^2 \cdot L_J \cdot N}{2} \nonumber \\
		&= J(\bar{ \btheta}_{t})  + \frac{\alpha_t}{N} \sum_{i, j = 1}^{N} \bigg \langle \nabla_{\theta_i^s} J( \btheta_{t}), \widehat{\delta}_{j,t} \cdot \psi_{\theta^{s}_j}(s_t, a_{j,t}; \theta_{j,t}) \bigg \rangle  + \alpha_t \sum_{i = 1}^{N} \bigg \langle \nabla_{\theta_i^p} J(\theta_{t}), \widehat{\delta}_{i,t} \cdot \psi_{\theta^{p}_i}(s_t, a_{i,t}; \theta_{i,t}) \bigg \rangle \nonumber \\
		& \quad - N \cdot \alpha_t \cdot \ell_p \cdot L_J \cdot \| Q \cdot \btheta^{s}_t \| -  \frac{N \cdot L_J \cdot \alpha_t^2 \cdot \ell_p^2}{2} 
		\label{bound: objective}
		\end{align}
		where $(a)$ follows the fact that the update of $\bar{\theta}_{i,t} := \{ \bar{\theta}_{t}^s, \theta^p_{i,t} \}$ could be decomposed as below:
		\begin{subequations}
		    \begin{align}
		        &\bar{\theta}^s_{t+1} - \bar{\theta}^s_{t} = \frac{\alpha_t}{N} \sum_{j = 1}^{N} \widehat{\delta}_{j,t} \cdot \psi_{\theta^{s}_j}(s_t, a_{j,t}; \theta_{j,t}), \nonumber \\
		        &\theta^p_{i,t+1} - \theta^p_{i,t} = \alpha_t \cdot \widehat{\delta}_{i,t} \cdot \psi_{\theta^{p}_i}(s_t, a_{i,t}; \theta_{i,t}). \nonumber
		    \end{align}
		\end{subequations}
		 Taking expectation on both sides of inequality \eqref{bound: objective}, we 
		have:
        \begin{align}\label{eq:expectation:J}
            \mathbb{E}[J(\bar{\btheta}_{t+1})] & \ge \mathbb{E}[J(\bar{\btheta}_{t})] + I_{1,1} + I_{1,2} + I_{2,1} + I_{2,2} \nonumber\\
            & \quad - N \cdot \alpha_t \cdot \ell_p \cdot L_J \cdot \EE \big[ \| Q \cdot \btheta^{s}_t \| \big] -  \frac{N \cdot L_J \cdot \alpha_t^2 \cdot \ell_p^2}{2}
        \end{align}
		where we have defined:
		\begin{align}
		I_{1,1} & := \frac{\alpha_t}{N} \sum_{i, j = 1}^{N} \mathbb{E} \bigg[ \bigg \langle \nabla_{\theta_i^s} J( \btheta_{t}), \bigg( \widehat{\delta}_{j,t} - \widehat{\delta}_{t}^* \bigg) \cdot \psi_{\theta^{s}_j}(s_t, a_{j,t} ; \theta_{j,t}) \bigg \rangle \bigg]  \nonumber \\
		I_{1,2} & :=  \frac{\alpha_t}{N} \sum_{i, j = 1}^{N} \mathbb{E} \bigg[ \bigg \langle \nabla_{\theta_i^s} J( \btheta_{t}),  \widehat{\delta}_{t}^* \cdot \psi_{\theta^{s}_j}(s_t, a_{j,t} ; \theta_{j,t}) \bigg \rangle \bigg]  \nonumber \\
		I_{2,1} & :=  \alpha_t \sum_{i = 1}^{N}  \mathbb{E} \bigg[ \bigg \langle \nabla_{\theta_i^p} J( \btheta_{t}), \bigg( \widehat{\delta}_{i,t} - \widehat{\delta}_t^* \bigg) \psi_{\theta^{p}_i}(s_t, a_{i,t} ; \theta_{i,t}) \bigg \rangle \bigg] \nonumber \\
		I_{2,2} & :=  \alpha_t \sum_{i = 1}^{N}  \mathbb{E} \bigg[ \bigg \langle \nabla_{\theta_i^p} J( \btheta_{t}), \widehat{\delta}_t^* \cdot \psi_{\theta^{p}_i}(s_t, a_{i,t} ; \theta_{i,t}) \bigg \rangle \bigg] \nonumber \\
		\widehat{\delta}_{t}^* & : = \varphi(s_t,\ba_t)^T \lambda_t^* + \gamma\cdot \phi(s_{t+1})^T \omega_{t}^* - \phi(s_t)^T \omega_{t}^*, \label{def:TD:optimal_estimated}
		\end{align}
		and these terms satisfy the following relations:
		\begin{align}
		I_{1,1} + I_{1,2} & = \frac{\alpha_t}{N} \sum_{i, j = 1}^{N} \mathbb{E} \bigg[ \bigg \langle \nabla_{\theta_i^s} J(\btheta_{t}), \widehat{\delta}_{j,t} \cdot \psi_{\theta^{s}_j}(s_t, a_{j,t} ; \theta_{j,t}) \bigg \rangle \bigg] \nonumber \\
		I_{2,1} + I_{2,2} &= \alpha_t \sum_{i = 1}^{N}  \mathbb{E} \bigg[ \bigg \langle \nabla_{\theta_i^p} J(\btheta_{t}), \widehat{\delta}_{i,t} \cdot \psi_{\theta^{p}_i}(s_t, a_{i,t} ; \theta_{i,t} ) \bigg \rangle \bigg] \nonumber.
		\end{align}
		
		Below we analyze each term on the rhs of \eqref{eq:expectation:J}. Towards this end, let us define $ \overline{\nabla_{\btheta^s} J( \btheta_{t})} := \frac{1}{N} \sum_{i = 1}^{N} \nabla_{\theta_i^s} J(\btheta_{t}) $. We have the following: %{\red[the second euquality has problem.]}
% 		Through plugging $ \lambda^*_t $ and $ \omega^*_t $ into $ \Delta_{i,t} $, we could define $ \widehat{\delta}_{t}^* = \varphi(s_t,\ba_t)^T \lambda_t^* + \gamma \phi(s_{t+1})^T \omega_{t}^* - \phi(s_t)^T \omega_{t}^* $. Moreover, we decompose terms $I_1$ and $I_2$ as $ I_1 := I_{1,1} + I_{1,2} $ and $ I_2 := I_{2,1} + I_{2,2} $. Therefore, we have the definitions as below: 
		\begin{align}
		I_{1,1} &= \frac{\alpha_t}{N} \sum_{i, j = 1}^{N} \mathbb{E} \bigg[ \bigg \langle \nabla_{\theta_i^s} J( \btheta_{t}), \bigg( \widehat{\delta}_{j,t} - \widehat{\delta}_{t}^* \bigg) \cdot \psi_{\theta^{s}_j}(s_t, a_{j,t} ; \theta_{j,t}) \bigg \rangle \bigg]   \nonumber \\
		&=  - \alpha_t \cdot \sum_{j = 1}^{N} \cdot \mathbb{E} \bigg[ \bigg \langle \frac{1}{N} \sum_{i = 1}^{N} \nabla_{\theta_i^s} J(\btheta_{t}), \varphi(s_t, \ba_t)^T (\lambda_t^* - \lambda_{j,t}) \cdot \psi_{\theta^{s}_j}(s_t, a_{j,t} ; \theta_{j,t}) \bigg \rangle \bigg] \nonumber \\
		&\quad - \alpha_t \cdot \sum_{j = 1}^{N} \cdot \mathbb{E} \bigg[ \bigg \langle \frac{1}{N} \sum_{i = 1}^{N} \nabla_{\theta_i^s} J(\btheta_{t}),  (\gamma \cdot \phi(s_{t+1}) - \phi(s_t))^T (\omega_{t}^* - \omega_{j,t}) \cdot \psi_{\theta^{s}_j}(s_t, a_{j,t} ; \theta_{j,t}) \bigg \rangle \bigg]  \nonumber \\
		&\overset{(i)}{\geq} -\alpha_t \cdot C_{\psi} \cdot \sum_{j = 1}^{N} \mathbb{E} \bigg[ \| \overline{\nabla_{\btheta^s} J( \btheta_{t})} \| \cdot \| \lambda_{t}^* - \lambda_{i,t} \|  \bigg]  - (1 + \gamma) \cdot \alpha_t \cdot C_{\psi} \cdot \sum_{j = 1}^{N} \mathbb{E} \bigg[ \| \overline{\nabla_{\btheta^s} J( \btheta_{t})} \| \cdot \| \omega_{t}^* - \omega_{i,t} \|  \bigg]  \nonumber \\
		&\geq - 2 \alpha_t \cdot C_{\psi} \cdot \sum_{j = 1}^{N} \mathbb{E} \bigg[ \| \overline{\nabla_{\btheta^s} J( \btheta_{t})} \| \cdot \Big( \| \omega_{t}^* - \omega_{i,t} \| + \| \lambda_{t}^* - \lambda_{i,t} \| \Big) \bigg]   \nonumber \\
		&=  - 2 \alpha_t \cdot C_{\psi} \cdot \sum_{j = 1}^{N} \sqrt{ \bigg( \mathbb{E} \bigg[ \| \overline{\nabla_{\btheta^s} J( \btheta_{t})} \| \cdot \Big( \| \omega_{t}^* - \omega_{i,t} \| + \| \lambda_{t}^* - \lambda_{i,t} \| \Big) \bigg] \bigg)^2 }  \nonumber \\
		&\overset{(ii)}{\geq} - 2 \alpha_t \cdot C_{\psi} \cdot \sum_{j = 1}^{N} \sqrt{  \mathbb{E} \bigg[ \| \overline{\nabla_{\btheta^s} J( \btheta_{t})} \|^2 \bigg] \cdot \mathbb{E} \bigg[ \Big( \| \omega_{t}^* - \omega_{i,t} \| + \| \lambda_{t}^* - \lambda_{i,t} \| \Big)^2  \bigg]  }  \nonumber \\
		&\overset{(iii)}{\geq} - 4 \alpha_t \cdot C_{\psi} \cdot \sqrt{ N \cdot \mathbb{E} \bigg[ \| \overline{\nabla_{\btheta^s} J( \btheta_{t})} \|^2 \bigg]} \cdot \sqrt{ \sum_{j = 1}^{N} \bigg( \mathbb{E} \bigg[ \| \omega_{t}^* - \omega_{i,t} \|^2 \bigg] + \mathbb{E} \bigg[ \| \lambda_{t}^* - \lambda_{i,t} \|^2 \bigg] \bigg) }  \label{bound:actor:I_11}
		\end{align}
		where $(i)$ is due to the facts that all feature vectors are bounded (cf. Assumption \ref{ass:feature_bound}),  and the score functions are bounded as $ \| \psi_{\theta^{s}_j}(s_t, a_{j,t} ; \theta_{j,t}) \| \leq C_{\psi}$ (cf.  \eqref{Lipschitz: policy}); $(ii)$ and $(iii)$ follow from the Cauchy-Schwarz inequality.
	     
	     Here, we define the TD error 
	     \begin{align}
	         \delta_t := Q_{\pi_{\theta_t}}(s_t, \ba_t) - V_{\pi_{\btheta_t}}(s_t) =  \bar{r}_t + \gamma V_{\pi_{\btheta_t}}(s_{t+1}) - V_{\pi_{\btheta_t}}(s_t) \label{def:global:TD}
	     \end{align}
	     and $ \overline{\nabla_{\btheta^s} J( \btheta_{t})} := \frac{1}{N} \sum_{i = 1}^{N} \nabla_{\theta_i^s} J(\btheta_{t}) $, then the term $ I_{1,2} $ can be decomposed as below:
		\begin{align}
		I_{1,2} &:= \frac{\alpha_t}{N} \cdot \sum_{i, j = 1}^{N} \mathbb{E} \bigg[ \bigg \langle \nabla_{\theta_i^s} J( \btheta_{t}),  \widehat{\delta}_t^* \cdot \psi_{\theta^{s}_j}(s_t, a_{j,t} ; \theta_{j,t}) \bigg \rangle \bigg] \nonumber \\
		&= \alpha_t  \cdot \sum_{j=1}^N \mathbb{E} \bigg[ \bigg \langle \frac{1}{N} \sum_{i=1}^N \nabla_{\theta_i^s} J( \btheta_{t}), \widehat{\delta}_t^* \cdot \psi_{\theta^{s}_j}(s_t, a_{j,t} ; \theta_{j,t}) \bigg \rangle \bigg] \nonumber \\
		&= \alpha_t \cdot (1-\gamma) \cdot N \cdot \mathbb{E}\bigg[ \| \overline{\nabla_{\btheta^s} J(\btheta_{t})} \|^2 \bigg] + \alpha_t \cdot \sum_{j=1}^N \mathbb{E} \bigg[ \bigg \langle  \overline{\nabla_{\btheta^s} J(\btheta_{t})}, (\widehat{\delta}_t^* - \delta_t) \cdot \psi_{\theta^{s}_j}(s_t, a_{j,t} ; \theta_{j,t}) \bigg \rangle \bigg] \nonumber \\
		& \quad + \alpha_t \cdot \sum_{j=1}^N \mathbb{E} \bigg[ \bigg \langle  \overline{\nabla_{\btheta^s} J(\btheta_{t})}, \delta_t \cdot  \psi_{\theta^{s}_j}(s_t, a_{j,t} ; \theta_{j,t}) - (1-\gamma) \cdot \overline{\nabla_{\btheta^s} J(\btheta_{t})} \bigg \rangle \bigg] \label{derive:actor:I_12}.
		\end{align}

		Then we are able to bound each term above:
		\begin{align}
		& \alpha_t \cdot \sum_{j=1}^N \mathbb{E} \bigg[ \bigg \langle  \overline{\nabla_{\btheta^s} J(\btheta_{t})}, (\widehat{\delta}_t^* - \delta_t) \cdot \psi_{\theta^{s}_j}(s_t, a_{j,t} ; \theta_{j,t}) \bigg \rangle \bigg] \nonumber \\
		&\geq - \alpha_t \cdot \sum_{j = 1}^{N} \mathbb{E} \bigg[  \|  \overline{\nabla_{\btheta^s} J(\btheta_{t})}  \| \cdot \| \psi_{\theta^{s}_j}(s_t, a_{j,t} ; \theta_{j,t})  \| \cdot \big| \widehat{\delta}_t^* - \delta_t \big| \bigg] \nonumber \\
		& \overset{(i)}{\geq} - \alpha_t \cdot L_v \cdot C_{\psi} \cdot N \cdot \mathbb{E} \bigg[ \bigg| \bigg( \varphi(s_t, \ba_t)^T \lambda_t^* - \bar{r}_t \bigg) + \gamma\bigg( \phi(s_{t+1})^T \omega_{t}^* - V_{\pi_{\btheta_{t}}}(s_{t+1}) \bigg) - \bigg( \phi(s_t)^T \omega_{t}^* - V_{\pi_{\btheta_{t}}}(s_t) \bigg) \bigg| \bigg ] \nonumber \\
		& \overset{(ii)}{\geq} - \alpha_t \cdot L_v \cdot C_{\psi} \cdot N \cdot \left( \mathbb{E} \bigg[| \varphi(s_t,a_t)^T \lambda_t^* - \bar{r}_t | \bigg] + \gamma \mathbb{E} \bigg[ | \phi(s_{t+1})^T \omega_{t}^* - V_{\pi_{\btheta_{t}}}(s_{t+1}) | \bigg] + \mathbb{E}\bigg[ | V_{\pi_{\btheta_{t}}}(s_t) - \phi(s_t)^T \omega_{t}^* | \bigg] \right) \nonumber \\
		& \overset{(iii)}{\geq} - \alpha_t \cdot L_v \cdot C_{\psi} \cdot N \cdot \bigg( \sqrt{\mathbb{E} \bigg[ \left( \varphi(s_t,\ba_t)^T \lambda_t^* - \bar{r}_t \right)^2 \bigg]} + \gamma \sqrt{\mathbb{E} \bigg[ \left( \phi(s_{t+1})^T \omega_{t}^* - V_{\pi_{\btheta_{t}}}(s_{t+1}) \right)^2 \bigg]}  \nonumber \\
		& \quad + \sqrt{\mathbb{E} \bigg[ \left( V_{\pi_{\btheta_{t}}}(s_t) - \phi(s_t)^T \omega_{t}^* \right)^2 \bigg]} \bigg) \nonumber \\
		& \overset{(iv)}{\geq} - 2 N \cdot \alpha_t \cdot L_v \cdot C_{\psi} \cdot \epsilon_{app} \label{bound:I_121:approximation}
		\end{align}
		where $(i)$ follows \eqref{Lipschitz: policy}, \eqref{eq:Lipschitz:objective}, definition of $ \widehat{\delta}_t^* $ in \eqref{def:TD:optimal_estimated} and definition of $\delta_t$ in \eqref{def:global:TD}; $(ii)$ follows the triangle inequality; $(iii)$ follows Jensen's inequality; $(iv)$ follows the definition of approximation error in \eqref{error:approximation}. 
		
		Recall the definition $Q_{\pi_{\btheta}}(s,\ba) := \mathbb{E}\big[ \sum_{t=0}^\infty \gamma^t r(s_t,\ba_t) | s_0 = s, \ba_0 = \ba \big]$. According to \cite{agarwal2019reinforcement}, policy gradient could be expressed as below
		\begin{align}
		    \nabla J(\btheta) := \frac{1}{1-\gamma} \mathbb{E}_{s \sim d_{\btheta}(\cdot), a \sim \pi_{\btheta}(\cdot| s) } \bigg[ \big(Q_{\pi_{\btheta}}(s,\ba) - V_{\pi_{\btheta}}(s)\big) \cdot \psi_{\btheta}(s,\ba) \bigg] \label{eq:pg:advantange}
		\end{align}
	     where $d_{\btheta}(\cdot)$ denotes the discounted visitation measure $d_{\btheta}(s) := \mathbb{E}_{s_0 \sim \eta} \big[(1-\gamma)\sum_{t=0}^\infty \gamma^t \mathcal{P}^{\pi_{\btheta}}(s_t = s| s_0) \big]$. According to the policy gradient as shown in \eqref{eq:pg:advantange}, it holds that: %{\red[can we merge this with (124)?]}
		\begin{align}
		    \overline{\nabla_{\btheta^s} J(\btheta_{t})} := \frac{1}{1-\gamma} \mathbb{E}_{ s_t \sim d_{\btheta_t}(\cdot), \ba_t \sim \pi_{\btheta_t}(\cdot| s_t) } \bigg[ \big( Q_{\pi_{\btheta_{t}}} (s_t, \ba_t) - V_{\pi_{\btheta_{t}}} (s_t) \big) \cdot \big(\frac{1}{N} \sum_{j = 1}^{N} \psi_{\theta^{s}_j}(s_t, a_{j,t} ; \theta_{j,t})\big) \bigg]. \label{eq:pg:shared}
		\end{align}
		Therefore, the third term in \eqref{derive:actor:I_12} could be bounded as
		\begin{align}
		& \alpha_t \cdot \sum_{j=1}^N \mathbb{E} \bigg[ \bigg \langle  \overline{\nabla_{\btheta^s} J(\btheta_{t})}, \delta_t \cdot  \psi_{\theta^{s}_j}(s_t, a_{j,t} ; \theta_{j,t}) - (1-\gamma) \cdot \overline{\nabla_{\btheta^s} J(\btheta_{t})} \bigg \rangle \bigg] \nonumber \\
		&= \alpha_t \cdot N \cdot \mathbb{E} \bigg[ \Big \langle  \overline{\nabla_{\btheta^s} J(\btheta_{t})}, \delta_t \cdot \big( \frac{1}{N} \sum_{j=1}^N \psi_{\theta^{s}_j}(s_t, a_{j,t} ; \theta_{j,t}) \big) - (1-\gamma) \cdot \overline{\nabla_{\btheta^s} J(\btheta_{t})} \Big \rangle \bigg] \nonumber \\
		&\overset{(i)}{=} \alpha_t \cdot N \cdot \mathbb{E} \bigg[ \EE \Big[ \Big \langle  \overline{\nabla_{\btheta^s} J(\btheta_{t})}, \delta_t \cdot \big( \frac{1}{N} \sum_{j=1}^N \psi_{\theta^{s}_j}(s_t, a_{j,t} ; \theta_{j,t}) \big) - (1-\gamma) \cdot \overline{\nabla_{\btheta^s} J(\btheta_{t})} \Big \rangle \Big| \btheta_t \Big] \bigg] \nonumber \\
		& = \alpha_t \cdot N \cdot \mathbb{E} \bigg[ \bigg \langle \overline{\nabla_{\btheta^s} J(\btheta_{t})}, \bigg( \mathbb{E}_{ s_t \sim \mu_{\btheta_t}(\cdot), \ba_t \sim \pi_{\btheta_t}(\cdot| s_t) } \bigg[ \big( Q_{\pi_{\btheta_{t}}} (s_t, \ba_t) - V_{\pi_{\btheta_{t}}} (s_t) \big) \cdot \big(\frac{1}{N} \sum_{j = 1}^{N} \psi_{\theta^{s}_j}(s_t, a_{j,t} ; \theta_{j,t})\big) \bigg| \btheta_t \bigg] \nonumber \\
		& \quad - (1-\gamma) \cdot \overline{\nabla_{\btheta^s} J(\btheta_{t})} \bigg) \bigg \rangle  \bigg] \nonumber \\
		&\overset{(ii)}{\geq} -\alpha_t \cdot L_v \cdot N \cdot \bigg \| \mathbb{E}_{ s_t \sim \mu_{\btheta_t}(\cdot), \ba_t \sim \pi_{\btheta_t}(\cdot| s_t) } \bigg[ \big( Q_{\pi_{\btheta_{t}}} (s_t, \ba_t) - V_{\pi_{\btheta_{t}}} (s_t) \big) \cdot \big(\frac{1}{N} \sum_{j = 1}^{N} \psi_{\theta^{s}_j}(s_t, a_{j,t} ; \theta_{j,t})\big) \bigg] \nonumber \\
		& \quad - \mathbb{E}_{ s_t \sim d_{\btheta_t}(\cdot), \ba_t \sim \pi_{\btheta_t}(\cdot| s_t) } \bigg[ \big( Q_{\pi_{\btheta_{t}}} (s_t, \ba_t) - V_{\pi_{\btheta_{t}}} (s_t) \big) \cdot \big(\frac{1}{N} \sum_{j = 1}^{N} \psi_{\theta^{s}_j}(s_t, a_{j,t} ; \theta_{j,t})\big) \bigg] \bigg \| \nonumber \\
		&\overset{(iii)}{\geq} -\alpha_t \cdot L_v \cdot N \cdot \sup_{s, \ba} \bigg \| \big( Q_{\pi_{\btheta_{t}}} (s_t, \ba_t) - V_{\pi_{\btheta_{t}}} (s_t) \big) \cdot \big(\frac{1}{N} \sum_{j = 1}^{N} \psi_{\theta^{s}_j}(s_t, a_{j,t} ; \theta_{j,t})\big) \bigg \| \cdot d_{TV}( \mu_{\btheta_{t}},  d_{\btheta_{t}} ) \nonumber \\
		&\overset{(iv)}{\geq} -4\alpha_t \cdot N \cdot  L_v \cdot C_{\psi} \cdot R_{\max} \cdot \left( \log_{\tau} \kappa^{-1} + \frac{1}{1 - \tau} \right) = -\alpha_t \cdot N \cdot \epsilon_{sp}  \label{bound:I_121:sampling}
		\end{align}
		where $(i)$ follows the tower rule; $(ii)$ follows \eqref{eq:Lipschitz:objective} and \eqref{eq:pg:shared}; $(iii)$ is due to distribution mismatch between the policy gradient in \eqref{eq:pg:shared} and its estimator; the sampling error $\epsilon_{sp}$ is defined in \eqref{error:sampling} and the inequality $(iv)$ is due to the facts that $Q_{\pi_{\btheta_t}}(s_t, \ba_t) \leq \frac{R_{\max}}{1-\gamma} $, $V_{\pi_{\btheta_t}}(s_t) \leq \frac{R_{\max}}{1-\gamma} $, $\| \psi_{\theta^{s}_j}(s_t, a_{j,t} ; \theta_{j,t}) \| \leq C_{\psi} $ and 
		the distribution mismatch inequality \eqref{eq:sampling_mismatch} in Lemma \ref{lemma:sampling_mismatch}.
		
		By plugging the inequalities \eqref{bound:I_121:approximation} - \eqref{bound:I_121:sampling} into \eqref{derive:actor:I_12}, we obtain the following:
		\begin{align}
		I_{1,2} &:=  \frac{\alpha_t}{N} \sum_{i, j = 1}^{N} \mathbb{E} \bigg[ \bigg \langle \nabla_{\theta_i^s} J( \btheta_{t}),  \widehat{\delta}_t^* \cdot \psi_{\theta^{s}_j}(s_t, a_{j,t} ; \theta_{j,t}) \bigg \rangle \bigg] \nonumber \\
		&\geq  - 2 N \cdot \alpha_t \cdot L_v \cdot C_{\psi} \cdot \epsilon_{app} -\alpha_t \cdot N \cdot \epsilon_{sp} + \alpha_t \cdot (1 - \gamma) \cdot N \cdot \mathbb{E}\bigg[ \| \overline{\nabla_{\btheta^s} J(\btheta_{t})}   \|^2 \bigg] \label{bound:actor:I_12}
		\end{align}
		Moreover, by adding \eqref{bound:actor:I_11} and \eqref{bound:actor:I_12}, we obtain
		$ I_1 :=  I_1^1 +I_1^2 $ as below:
		\begin{align}
		I_1 &\geq - 2 N \cdot \alpha_t \cdot L_v \cdot C_{\psi} \cdot \epsilon_{app} -\alpha_t \cdot N \cdot \epsilon_{sp} + \alpha_t \cdot (1 - \gamma) \cdot N \cdot \mathbb{E}\bigg[ \| \overline{\nabla_{\btheta^s} J(\btheta_{t})}   \|^2 \bigg] \nonumber \\
		& - 4 \alpha_t \cdot C_{\psi} \cdot \sqrt{ N \cdot \mathbb{E} \bigg[ \| \overline{\nabla_{\btheta^s} J( \btheta_{t})} \|^2 \bigg]} \cdot \sqrt{ \sum_{j = 1}^{N} \bigg( \mathbb{E} \bigg[ \| \omega_{t}^* - \omega_{i,t} \|^2 \bigg] + \mathbb{E} \bigg[ \| \lambda_{t}^* - \lambda_{i,t} \|^2 \bigg] \bigg) }. \label{bound:results:I_1}
		\end{align}
		
	In order to bound term $I_2$, we are able to further analyze $ I_{2,1} $ and $ I_{2,1} $ as below:
		\begin{align}
		I_{2,1} &:= \alpha_t \cdot \sum_{i = 1}^{N}  \mathbb{E} \bigg[ \bigg \langle \nabla_{\theta_i^p} J( \btheta_{t}), \bigg( \widehat{\delta}_{i,t} - \widehat{\delta}_t^* \bigg) \psi_{\theta^{p}_i}(s_t, a_{i,t} ; \theta_{i,t}) \bigg \rangle \bigg] \nonumber \\
		&=  - \alpha_t \cdot \sum_{i = 1}^{N} \mathbb{E} \bigg[ \Big \langle \nabla_{\theta_i^p} J(\btheta_{t}),  \big(\gamma \phi(s_{t+1}) - \phi(s_t)\big)^T \big(\omega_{t}^* - \omega_{i,t}\big) \cdot \psi_{\theta^{p}_i}(s_t, a_{i,t} ; \theta_{i,t}) \Big \rangle \bigg]  \nonumber \\
		& \quad - \alpha_t \cdot \sum_{i = 1}^{N} \mathbb{E} \bigg[ \Big \langle \nabla_{\theta_i^p} J(\btheta_{t}), \varphi(s_t, \ba_t)^T (\lambda_{t}^* - \lambda_{i,t}) \cdot \psi_{\theta^{p}_i}(s_t, a_{i,t} ; \theta_{i,t}) \Big \rangle \bigg]  \nonumber \\
		& \overset{(i)}{\geq}  - 2 \alpha_t \cdot C_{\psi} \cdot \sum_{i = 1}^{N} \mathbb{E} \bigg[ \| \nabla_{\theta_i^p} J(\btheta_{t}) \| \cdot \bigg( \| \omega_{t}^* - \omega_{i,t} \| + \| \lambda_{t}^* - \lambda_{i,t} \| \bigg)  \bigg]  \nonumber \\
		&=  - 2 \alpha_t \cdot C_{\psi} \cdot \sum_{i = 1}^{N} \sqrt{ \bigg( \mathbb{E} \bigg[ \| \nabla_{\theta_i^p} J(\btheta_{t}) \| \cdot \bigg( \| \omega_{t}^* - \omega_{i,t} \| + \| \lambda_{t}^* - \lambda_{i,t} \| \bigg)  \bigg] \bigg)^2 }  \nonumber \\
		& \overset{(ii)}{\geq} - 2 \alpha_t \cdot C_{\psi} \cdot \sum_{i = 1}^{N} \sqrt{  \mathbb{E} \bigg[ \| \nabla_{\theta_i^p} J(\btheta_{t}) \|^2 \bigg]} \cdot \sqrt{ \mathbb{E}\bigg[ ( \| \omega_{t}^* - \omega_{i,t} \| + \| \lambda_{t}^* - \lambda_{i,t} \|  )^2  \bigg]  } \nonumber \\
		& \overset{(iii)}{\geq} - 4 \alpha_t \cdot C_{\psi} \cdot \sqrt{ \sum_{i=1}^N \mathbb{E} \bigg[ \| \nabla_{\theta_i^p} J(\btheta_{t}) \|^2 \bigg]} \cdot \sqrt{ \sum_{i = 1}^{N} \mathbb{E}\bigg[  \| \omega_{t}^* - \omega_{i,t} \|^2 + \| \lambda_{t}^* - \lambda_{i,t} \|^2  \bigg]  }
		\label{bound:results:I_21}
		\end{align}
		where $(i)$ is due to the facts that all feature vectors are bounded by Assumption \ref{ass:feature_bound} and the score functions are bounded as : $ \| \psi_{\theta^{p}_i}(s_t, a_{j,t} ; \theta_{i,t}) \| \leq C_{\psi}$ in \eqref{Lipschitz: policy}; $(ii)$ and $(iii)$ are from  Cauchy–Schwarz inequality. Moreover, for the term $ I_{2,2} $ in \eqref{eq:expectation:J}, we can further express it as:
		\begin{align}
		I_{2,2} &:= \alpha_t \sum_{i = 1}^{N}  \mathbb{E} \bigg[ \bigg \langle \nabla_{\theta_i^p} J( \btheta_{t}), \widehat{\delta}_t^* \cdot \psi_{\theta^{p}_i}(s_t, a_{i,t} ; \theta_{i,t}) \bigg \rangle \bigg] \nonumber \\
		&= \alpha_t \cdot (1-\gamma) \sum_{i = 1}^{N} \mathbb{E} \bigg[ \| \nabla_{\theta_i^p} J( \btheta_{t}) \|^2 \bigg] + \alpha_t \sum_{i = 1}^{N} \mathbb{E} \bigg[ \bigg \langle \nabla_{\theta_i^p} J( \btheta_{t}), \bigg( \widehat{\delta}_t^* - \delta_t  \bigg) \cdot \psi_{\theta^{p}_i}(s_t, a_{i,t} ; \theta_{i,t}) \bigg \rangle \bigg]  \nonumber \\
		& \quad + \alpha_t \sum_{i = 1}^{N} \mathbb{E} \bigg[ \bigg \langle \nabla_{\theta_i^p} J( \btheta_{t}), \delta_t \cdot \psi_{\theta^{p}_i}(s_t, a_{i,t} ; \theta_{i,t}) - (1-\gamma) \nabla_{\theta_i^p} J( \btheta_{t}) \bigg \rangle \bigg] \label{bound:actor:I_22}
		\end{align}
		where we have defined $ \delta_t := \bar{r}_t + \gamma V_{\pi_{\btheta_t}}(s_{t+1}) - V_{\pi_{\btheta_t}}(s_t) $.
		
		For each term above in \eqref{bound:actor:I_22}, it holds that 
		\begin{align}
		& \alpha_t \sum_{i = 1}^{N} \mathbb{E} \bigg[ \bigg \langle \nabla_{\theta_i^p} J( \btheta_{t}), \bigg( \widehat{\delta}_t^* - \delta_t  \bigg) \cdot \psi_{\theta^{p}_i}(s_t, a_{i,t} ; \theta_{i,t}) \bigg \rangle \bigg]  \nonumber \\
		&\geq - \alpha_t \sum_{i = 1}^{N} \mathbb{E} \bigg[  \|  \nabla_{\theta_i^p} J( \btheta_{t})  \| \cdot \| \psi_{\theta^{p}_i}(s_t, a_{i,t} ; \theta_{i,t})   \| \cdot \big| \widehat{\delta}_t^* - \delta_t \big| \bigg] \nonumber \\
		& \overset{(i)}{\geq} - \alpha_t \cdot L_v \cdot C_{\psi} \cdot N \cdot \mathbb{E} \bigg[ \bigg| \bigg( \varphi(s_t, \ba_t)^T \lambda_t^* - \bar{r}_t \bigg) + \gamma\bigg( \phi(s_{t+1})^T \omega_{t}^* - V_{\pi_{\btheta_{t}}}(s_{t+1}) \bigg) - \bigg( \phi(s_t)^T \omega_{t}^* - V_{\pi_{\btheta_{t}}}(s_t) \bigg) \bigg| \bigg ] \nonumber \\
		& \overset{(ii)}{\geq} - \alpha_t \cdot L_v \cdot C_{\psi} \cdot N \cdot \left( \mathbb{E}[| \varphi(s_t,a_t)^T \lambda_t^* - r_t |] + \gamma \mathbb{E} \bigg[ | \phi(s_{t+1})^T \omega_{t}^* - V_{\pi_{\btheta_{t}}}(s_{t+1}) | \bigg] + \mathbb{E}\bigg[ | V_{\pi_{\btheta_{t}}}(s_t) - \phi(s_t)^T \omega_{t}^* | \bigg] \right) \nonumber \\
		& \overset{(iii)}{\geq} - \alpha_t \cdot L_v \cdot C_{\psi} \cdot N \cdot \bigg( \sqrt{\mathbb{E} \bigg[ \left( \varphi(s_t,\ba_t)^T \lambda_t^* - \bar{r}_t \right)^2 \bigg]} + \gamma \sqrt{\mathbb{E} \bigg[ \left( \phi(s_{t+1})^T \omega_{t}^* - V_{\pi_{\btheta_{t}}}(s_{t+1}) \right)^2 \bigg]}  \nonumber \\
		& \quad + \sqrt{\mathbb{E} \bigg[ \left( V_{\pi_{\btheta_{t}}}(s_t) - \phi(s_t)^T \omega_{t}^* \right)^2 \bigg]} \bigg) \nonumber \\
		& \overset{(iv)}{\geq} - 2 N \cdot \alpha_t \cdot L_v \cdot C_{\psi} \cdot \epsilon_{app} \label{bound:I_221:approximation}
		\end{align}
		where $(i)$ follows \eqref{eq:Lipschitz:objective} and $\| \psi_{\theta^{p}_i}(s_t, a_{i,t} ; \theta_{i,t}) \| \leq C_{\psi} $; $(ii)$ follows from the triangle inequality; $(iii)$ follows Jensen's inequality; $(iv)$ follows the definition of approximation error in \eqref{error:approximation}.
		
		Similarly as the derivation in \eqref{bound:I_121:sampling}, we can further bound the third term in \eqref{bound:actor:I_22} as below:
		\begin{align}
		\alpha_t \sum_{i = 1}^{N} \mathbb{E} \bigg[ \bigg \langle \nabla_{\theta_i^p} J( \btheta_{t}), \delta_t \cdot \psi_{\theta^{l}_i}(s_t, a_{i,t} ; \theta_{i,t}) - (1-\gamma) \nabla_{\theta_i^p} J( \btheta_{t}) \bigg \rangle \bigg] \geq -\alpha_t \cdot N \cdot \epsilon_{sp} \label{bound:I_221:sampling}.
		\end{align}
		
		Plugging the inequalities \eqref{bound:I_221:approximation} - \eqref{bound:I_221:sampling} into \eqref{bound:actor:I_22}, then it holds that: 
		\begin{align}
		I_{2,2} & \geq  \alpha_t \cdot (1 - \gamma) \cdot \sum_{i = 1}^{N} \mathbb{E} \bigg[ \| \nabla_{\theta_i^p} J( \btheta_{t}) \|^2 \bigg] - \alpha_t \cdot N \cdot \epsilon_{sp} - 2 \alpha_t \cdot N \cdot L_v \cdot C_{\psi} \cdot \epsilon_{app}.  \label{bound:results:I_22}
		\end{align}
		
		Adding \eqref{bound:results:I_21} and \eqref{bound:results:I_22}, it follows that: 
		\begin{align}
		I_{2,1} + I_{2,2}  & \geq \alpha_t \cdot (1 - \gamma) \cdot \sum_{i = 1}^{N} \mathbb{E} \bigg[ \| \nabla_{\theta_i^p} J( \btheta_{t}) \|^2 \bigg] - \alpha_t \cdot N \cdot \epsilon_{sp} - 2 \alpha_t \cdot N \cdot L_v \cdot C_{\psi} \cdot \epsilon_{app}  \nonumber \\
		& \quad - 4 \alpha_t \cdot C_{\psi} \cdot \sqrt{ \sum_{i=1}^N \mathbb{E} \bigg[ \| \nabla_{\theta_i^p} J(\btheta_{t}) \|^2 \bigg]} \cdot \sqrt{ \sum_{i = 1}^{N} \mathbb{E}\bigg[ ( \| \omega_{t}^* - \omega_{i,t} \| + \| \lambda_{t}^* - \lambda_{i,t} \|  )^2  \bigg]  } \label{bound:results:I_2}
		\end{align}
		
		Denote $ P_{t} :=  \sum_{i = 1}^{N} \mathbb{E}\big[ \| \omega_{t}^* - \omega_{i,t} \|^2 + \| \lambda_{t}^* - \lambda_{i,t} \|^2  \big]  $. Then we plug the inequalities \eqref{bound:results:I_1} and \eqref{bound:results:I_2} into \eqref{eq:expectation:J}, it holds that 
		\begin{align}
		\mathbb{E}[ J(\bar{\btheta}_{t+1}) ]
		&\geq \mathbb{E} [  J(\bar{ \btheta}_{t}) ] - N \cdot \alpha_t \cdot \ell_p \cdot L_J \cdot E \big[ \| Q \cdot \btheta^{s}_t \| \big] -  \frac{N \cdot L_J \cdot \alpha_t^2 \cdot \ell_p^2}{2} \nonumber \\
		& \quad + \alpha_t \cdot (1 - \gamma) \cdot N \cdot \mathbb{E}\bigg[ \| \overline{\nabla_{\btheta^s} J(\btheta_{t})}   \|^2 \bigg] - 2 N \cdot \alpha_t \cdot L_v \cdot C_{\psi} \cdot \epsilon_{app} -\alpha_t \cdot N \cdot \epsilon_{sp}  \nonumber \\
		& \quad + \alpha_t \cdot (1 - \gamma) \cdot \sum_{i = 1}^{N} \mathbb{E} \bigg[ \| \nabla_{\theta_i^p} J( \btheta_{t}) \|^2 \bigg] - 2 N \cdot \alpha_t \cdot L_v \cdot C_{\psi} \cdot \epsilon_{app} -\alpha_t \cdot N \cdot \epsilon_{sp}  \nonumber \\
		& \quad - 4 \alpha_t \cdot C_{\psi} \cdot \bigg( \sqrt{ N \cdot \mathbb{E} \bigg[ \| \overline{\nabla_{\btheta^s} J( \btheta_{t})} \|^2 \bigg] } + \sqrt{ \sum_{i = 1}^{N} \mathbb{E} \bigg[ \| \nabla_{\theta_i^p} J(\btheta_{t}) \|^2 \bigg] } \bigg) \cdot \sqrt{ P_t  } \label{bound:ascent:bar_theta}.
		\end{align}
		
		Denote a constant $ C_s :=  N \cdot \ell_p \cdot L_J$. Rearrange inequality \eqref{bound:ascent:bar_theta} and apply Cauchy-Schwarz inequality, it holds that 
		\begin{align}
		& N \cdot \mathbb{E}\bigg[ \| \overline{\nabla_{\btheta_s} J(\btheta_{t})}   \|^2 \bigg] + \sum_{i = 1}^{N} \mathbb{E} \bigg[ \| \nabla_{\theta_i^p} J( \btheta_{t}) \|^2 \bigg]  \nonumber \\
		&\leq \frac{1}{\alpha_t \cdot (1-\gamma)} \cdot \mathbb{E} \bigg[ J(\bar{\btheta}_{t+1}) - J(\bar{ \btheta}_{t}) \bigg]  + \frac{1}{1 - \gamma} \bigg( \frac{\alpha_t \cdot N  \cdot L_J \cdot \ell_p^2}{2} + 4 N \cdot L_v \cdot C_{\psi} \cdot \epsilon_{app} +  2N \cdot \epsilon_{sp} \bigg) \nonumber \\
		& + \frac{C_s}{1 - \gamma} \cdot \mathbb{E}\bigg[ \| Q \cdot \btheta^s_t \| \bigg] + \frac{8 C_{\psi} }{1 - \gamma} \cdot \sqrt{ N \cdot \mathbb{E}\bigg[ \| \overline{\nabla_{\btheta^s} J(\btheta_{t})}   \|^2 \bigg] + \sum_{i = 1}^{N} \mathbb{E} \bigg[ \| \nabla_{\theta_i^p} J(\btheta_{t}) \|^2 \bigg] } \cdot \sqrt{ P_{t} }  \label{bound: objective gradient}
		\end{align}
		
		Then we can denote $ G_t := N \cdot \mathbb{E}\bigg[ \| \overline{\nabla_{\btheta^s} J(\btheta_{t})}   \|^2 \bigg] + \sum_{i = 1}^{N} \mathbb{E} \bigg[ \| \nabla_{\theta_i^p} J( \btheta_{t}) \|^2 \bigg] $. Through summing the inequality \eqref{bound: objective gradient} from $ t = 0 $ to $ T-1 $ and divide $ T $ on both side, it holds that
		\begin{align}
		&\frac{1}{T} \sum_{t = 0}^{T-1} G_t \nonumber \\
		&\leq \underbrace{ \frac{1}{T} \sum_{t = 0}^{T-1} \frac{1}{\alpha_t(1 - \gamma)} \cdot \bigg( \mathbb{E} \big[ J(\bar{\btheta}_{t+1}) \big] - \EE \big[ J(\bar{\btheta}_{t}) \big] \bigg) + \frac{C_s}{T (1-\gamma)} \cdot \sum_{t = 0}^{T-1} \mathbb{E}\bigg[ \| Q \cdot \btheta^s_t \| \bigg] + \frac{N L_J \ell_p^2}{2T(1-\gamma)} \cdot \sum_{t = 0}^{T-1} \alpha_t }_{\rm term ~ M_1}  \nonumber \\
		& \quad + \underbrace{\frac{1}{1-\gamma} \bigg( 4 N \cdot L_v \cdot C_{\psi} \cdot \epsilon_{app} + 2N \cdot \epsilon_{sp} \bigg)}_{\rm term ~ M_2}  + \frac{8 C_{\psi}}{T(1-\gamma)} \cdot \sum_{t = 0}^{T-1} \sqrt{ G_t } \cdot \sqrt{ P_{t} } \label{bound:gradient:norm_square}
		\end{align}
		
		Then \eqref{bound:gradient:norm_square} could be expressed as below: 
		\begin{align}
		\frac{1}{T} \sum_{t = 0}^{T-1} G_t & \overset{\eqref{bound:gradient:norm_square}}{\leq} M_1 + M_2 + \frac{8C_{\psi}}{T \cdot (1-\gamma)} \cdot \sum_{t = 0}^{T-1} \sqrt{ G_t } \cdot \sqrt{ P_t } \nonumber \\
		&\overset{(a)}{\leq} M_1 + M_2 + \frac{8C_{\psi} }{1-\gamma} \cdot \sqrt{ \frac{1}{T} \sum_{t = 0}^{T-1} G_t} \cdot \sqrt{\frac{1}{T}\sum_{t = 0}^{T-1} P_t } \label{bound:average_grad:norm_square}
		\end{align}
		where $(a)$ follows Cauchy-Schwarz inequality. Then we define $ C_g := \frac{8 C_{\psi}}{1 - \gamma} $, $ B_1 := \frac{1}{T} \sum_{t = 0}^{T-1} G_t $, $ B_2 := M_1 + M_2 $ and $ B_3 := \frac{1}{T}\sum_{t = 0}^{T-1} P_t $. The inequality \eqref{bound:average_grad:norm_square} could be expressed as below:
		\begin{align}
		&B_1 \leq B_2 + C_g \cdot \sqrt{B_1 \cdot B_3} \implies \left( \sqrt{B_1} - \frac{C_g}{2} \cdot \sqrt{B_3} \right)^2 \leq B_2 + \frac{C_g^2}{4} \cdot B_3 \nonumber \\
		&\sqrt{B_1} \leq \sqrt{ B_2 + \frac{C_g^2}{4} \cdot B_3 } + \frac{C_g}{2} \cdot \sqrt{B_3} \implies B_1 \leq 2 B_2 + C_{g}^2 \cdot B_3 \label{order inequality: actor}
		\end{align}
		
		Then we are able to analyze the convergence rate of each term in \eqref{bound:gradient:norm_square}. Recall the fixed stepsize $\alpha_t = \frac{\alpha}{T^{\sigma_1}}$. We bound each component in $M_1$ defined above.
		\begin{align}
	    & \frac{1}{T} \sum_{t = 0}^{T-1} \frac{1}{\alpha_t (1-\gamma)} \cdot \bigg( \mathbb{E} \big[ J(\bar{\btheta}_{t+1}) \big] - \mathbb{E} \big[ J(\bar{\btheta}_{t}) \big] \bigg) \nonumber \\
		& \overset{(a)}{\leq} \frac{T^{\sigma_1}}{\alpha \cdot T \cdot (1-\gamma)} \cdot  \mathbb{E} \big[ J(\bar{\btheta}_{T}) \big] \nonumber \\
		& \overset{(b)}{\leq} \frac{T^{\sigma_1}}{\alpha_{T-1} \cdot T \cdot (1-\gamma)} \cdot \frac{R_{\max}}{1-\gamma} = \mathcal{O}\bigg(T^{\sigma_1 - 1} \bigg) \label{rate:M_11}
		\end{align}
		where $(a)$ follows $\alpha_t = \frac{\alpha}{T^{\sigma_1}}$; $(b)$ is due to the fact that each reward is bounded by $R_{\max}$ and $J(\btheta) \leq \sum_{t=0}^\infty \gamma^t \cdot R_{\max} = \frac{R_{\max}}{1-\gamma}$ for any $\btheta \in R^{N \times D}$. 
		
		For the second and third terms in  $M_1$, it holds that
		\begin{align}
		&\frac{C_s}{(1-\gamma) \cdot T} \cdot \sum_{t = 0}^{T-1} \mathbb{E}\bigg[ \| Q \cdot \btheta^s_t \| \bigg] \overset{\eqref{rate:policy_consensus}}{=} \mathcal{O}\bigg( T^{-\sigma_1} \bigg) \label{rate:M_12} \\
		&\frac{N \cdot L_J \cdot \ell_p^2}{(1-\gamma) \cdot T} \cdot \sum_{t = 0}^{T-1} \alpha_t = \frac{N \cdot L_J \cdot \ell_p^2}{(1-\gamma) } \cdot \frac{\alpha}{T^{\sigma_1}} = \mathcal{O}\bigg( T^{-\sigma_1} \bigg) \label{rate:M_13}
		\end{align}
		
		Combining \eqref{rate:M_11} - \eqref{rate:M_13}, we obtain that the convergence rate of term $M_1$ in \eqref{bound:gradient:norm_square} could be expressed as below:
		\begin{align}
		    M_1 = \mathcal{O}\bigg( T^{\sigma_1 - 1} \bigg) + \mathcal{O}\bigg(T^{-\sigma_1} \bigg) \label{rate:bound:M_1}
		\end{align}
		
		%{\blue Then we bound the right hand side of the above inequality}
		
		Then according to \eqref{bound:gradient:norm_square} and \eqref{order inequality: actor}, it holds that
		\begin{align}
		&\frac{1}{T} \sum_{t = 0}^{T-1} \bigg( N \cdot \mathbb{E}\bigg[ \| \overline{\nabla_{\btheta^s} J(\btheta_{t})}   \|^2 \bigg] + \sum_{i = 1}^{N} \mathbb{E} \bigg[ \| \nabla_{\theta_i^p} J( \btheta_{t}) \|^2 \bigg] \bigg) \nonumber \\
		\leq& 2M_1 + 2M_2 + \frac{C_g^2}{T} \sum_{i=1}^T \sum_{i = 1}^{N} \mathbb{E}\bigg[  \| \omega_{t}^* - \omega_{i,t} \|^2 + \| \lambda_{t}^* - \lambda_{i,t} \|^2   \bigg] \nonumber.
		\end{align}
		
		By applying in the convergence results of approximation parameters in \eqref{rate:lower:parameters}, we obtain that 
		\begin{align}
		& \frac{1}{T} \sum_{t = 0}^{T-1} \bigg( N \cdot \mathbb{E} \bigg[ \| \overline{\nabla_{\btheta^s} J(\btheta_{t})}   \|^2 \bigg] + \sum_{i = 1}^{N} \mathbb{E} \bigg[ \| \nabla_{\theta_i^p} J( \btheta_{t}) \|^2 \bigg] \bigg) \nonumber \\
		& \overset{\eqref{rate:lower:parameters}}{\leq} 2M_1 + 2M_2 + \frac{2}{T} \sum_{t = 0}^{T-1} \sum_{i=1}^N \bigg( \mathbb{E}\bigg[\|  \bar{\omega}_{t} - \omega^*_t \|^2\bigg] + \mathbb{E}\bigg[\| \bar{\lambda}_t - \lambda^*_t \|^2\bigg] \bigg) + \frac{2}{T} \sum_{t = 0}^{T-1} \bigg( \EE \bigg[ \| Q \cdot \bomega_t \|^2 \bigg] + \EE \bigg[ \| Q \cdot \blambda_t \|^2 \bigg] \bigg) \nonumber \\
		& = \mathcal{O}(\epsilon_{app} + \epsilon_{sp}) + \mathcal{O}(T^{\sigma_1 - 1}) + \mathcal{O}(T^{-\sigma_1}) + \mathcal{O}(T^{-1+\sigma_2}) + \mathcal{O}(T^{ - \sigma_2}) + \mathcal{O} \left( T^{\sigma_2 - 2\sigma_1} \right) \nonumber \\
		& \quad + \mathcal{O}(T^{-2\sigma_1 + 2\sigma_2}) + \mathcal{O}(T^{-2 + 2\sigma_2}) + \mathcal{O}(T^{-2\sigma_2}).  \label{rate: actor}
		\end{align}
		
		To optimize the convergence rate, we choose $ \alpha_t = \mathcal{O}(\frac{1}{T^{0.6}}), \beta_t = \mathcal{O}(\frac{1}{T^{0.4}}) $ so that $ \sigma_1 = \frac{3}{5} $ and $ \sigma_2 = \frac{2}{5} $, then plug them into \eqref{rate: actor}. Therefore, the convergence rate of the actor is:
		\begin{align}
		\frac{1}{T} \sum_{t = 0}^{T-1} \bigg( N \cdot \mathbb{E} \bigg[ \| \overline{\nabla_{\btheta^s} J(\btheta_{t})}   \|^2 \bigg] + \sum_{i = 1}^{N} \mathbb{E} \bigg[ \| \nabla_{\theta_i^p} J( \btheta_{t}) \|^2 \bigg] \bigg) = \mathcal{O}(T^{-\frac{2}{5}}) + \mathcal{O}(\epsilon_{app} + \epsilon_{sp})  \label{rate: policy}.
		\end{align}
		This completes the proof.
	\end{proof}
	
	\section{Detailed Analysis for Double Sampling Procedures} \label{double sampling analysis}
	
	For the \aname ~algorithm with double sampling procedures in Algorithm \ref{alg: double sampling}, we generate two different tuples $x_t := (s_t, \ba_t, s_{t+1})$ and $\tilde{x}_t := (\tilde{s}_t, \tilde{\ba}_t, \tilde{s}_{t+1})$ in each iteration $t$ to perform critic step and actor step. In $x_t$, the state $s_t$ is sampled from the stationary distribution $\mu_{\btheta_t}(\cdot)$. In $\tilde{x}_t$, the state $\tilde{s}_t$ is sampled from the discounted visitation measure $d_{\btheta_t}(\cdot)$ where $ d_{\btheta}(\tilde{s}) := (1-\gamma)\sum_{t=0}^\infty \gamma^t \cdot  \mathcal{P}^{\pi_{\btheta}} (s_t = \tilde{s}\mid s_0 \sim \eta)$.
	
	%we substitute all tuples $x_t := (s_t, a_t, s_{t+1})$ by $\tilde{x}_t := (\tilde{s}_t, \tilde{a}_t, \tilde{s}_{t+1})$ in the analysis steps of Proposition \ref{proposition:actor}.
	
	Then the difference between double sampling procedures and single sampling procedure comes from bounding the term $I_{1,2}$ in \eqref{derive:actor:I_12} and $I_{2,2}$ in \eqref{bound:actor:I_22}. With double samples $x_t$ and $\tilde{x}_t$ at each iteration $t$, the sampling error $\epsilon_{sp}$ could be avoided in analyzing $I_{1,2}$ and $I_{2,2}$

	With tuple $\tilde{x}_t$ to perform actor step, the component $ I_{1,2} $ in \eqref{derive:actor:I_12} could be expressed as  
	
	\begin{align}
		I_{1,2} &:= \alpha_t \cdot (1-\gamma) \cdot N \cdot \mathbb{E}\bigg[ \| \overline{\nabla_{\btheta^s} J(\btheta_{t})} \|^2 \bigg] + \alpha_t \cdot \sum_{j=1}^N \mathbb{E} \bigg[ \bigg \langle  \overline{\nabla_{\btheta^s} J(\btheta_{t})}, (\widehat{\delta}_t^* - \delta_t) \cdot \psi_{\theta^{s}_j}(\tilde{s}_t, \tilde{a}_{j,t} ; \theta_{j,t}) \bigg \rangle \bigg] \nonumber \\
		& \quad + \alpha_t \cdot \sum_{j=1}^N \mathbb{E} \bigg[ \bigg \langle  \overline{\nabla_{\btheta^s} J(\btheta_{t})}, \delta_t \cdot  \psi_{\theta^{s}_j}(\tilde{s}_t, \tilde{a}_{j,t} ; \theta_{j,t}) - (1-\gamma) \cdot \overline{\nabla_{\btheta^s} J(\btheta_{t})} \bigg \rangle \bigg] \nonumber
		\end{align}
	 Following the inequality \eqref{bound:I_121:approximation}, the second term in $I_{1,2}$ above could be bounded. Then it holds that 
	 \begin{align}
	     I_{1,2} \geq& -2N \cdot \alpha_t \cdot L_v \cdot C_{\psi} \cdot \epsilon_{app} + \alpha_t \cdot (1-\gamma) \cdot N \cdot \mathbb{E}\bigg[ \| \overline{\nabla_{\btheta^s} J(\btheta_{t})} \|^2 \bigg] \nonumber \\
	     & + \alpha_t \cdot \sum_{j=1}^N \mathbb{E} \bigg[ \bigg \langle  \overline{\nabla_{\btheta^s} J(\btheta_{t})}, \delta_t \cdot  \psi_{\theta^{s}_j}(\tilde{s}_t, \tilde{a}_{j,t} ; \theta_{j,t}) - (1-\gamma) \cdot \overline{\nabla_{\btheta^s} J(\btheta_{t})} \bigg \rangle \bigg] \nonumber
	 \end{align}
	 For the third term in $I_{1,2}$, it holds that
	 	\begin{align}
		& \alpha_t \cdot \sum_{j=1}^N \mathbb{E} \bigg[ \bigg \langle  \overline{\nabla_{\btheta^s} J(\btheta_{t})}, \delta_t \cdot  \psi_{\theta^{s}_j}(\tilde{s}_t, \tilde{a}_{j,t} ; \theta_{j,t}) - (1-\gamma) \cdot \overline{\nabla_{\btheta^s} J(\btheta_{t})} \bigg \rangle \bigg] \nonumber \\
		&= \alpha_t \cdot N \cdot \mathbb{E} \bigg[ \Big \langle  \overline{\nabla_{\btheta^s} J(\btheta_{t})}, \delta_t \cdot \big( \frac{1}{N} \sum_{j=1}^N \psi_{\theta^{s}_j}(\tilde{s}_t, \tilde{a}_{j,t} ; \theta_{j,t}) \big) - (1-\gamma) \cdot \overline{\nabla_{\btheta^s} J(\btheta_{t})} \Big \rangle \bigg] \nonumber \\
		&\overset{(i)}{=} \alpha_t \cdot N \cdot \mathbb{E} \bigg[ \EE \Big[ \Big \langle  \overline{\nabla_{\btheta^s} J(\btheta_{t})}, \delta_t \cdot \big( \frac{1}{N} \sum_{j=1}^N \psi_{\theta^{s}_j}(\tilde{s}_t, \tilde{a}_{j,t} ; \theta_{j,t}) \big) - (1-\gamma) \cdot \overline{\nabla_{\btheta^s} J(\btheta_{t})} \Big \rangle \Big| \btheta_t \Big] \bigg] \nonumber \\
		& \overset{(ii)}{=} \alpha_t \cdot N \cdot \mathbb{E} \bigg[ \bigg \langle \overline{\nabla_{\btheta^s} J(\btheta_{t})}, \bigg( \mathbb{E}_{ \tilde{s}_t \sim d_{\btheta_t}(\cdot), \tilde{\ba}_t \sim \pi_{\btheta_t}(\cdot| \tilde{s}_t) } \bigg[ \big( Q_{\pi_{\btheta_{t}}} (\tilde{s}_t, \tilde{\ba}_t) - V_{\pi_{\btheta_{t}}} (\tilde{s}_t) \big) \cdot \big(\frac{1}{N} \sum_{j = 1}^{N} \psi_{\theta^{s}_j}(\tilde{s}_t, \tilde{a}_{j,t} ; \theta_{j,t})\big) \bigg] \nonumber \\
		& \quad - (1-\gamma) \cdot \overline{\nabla_{\btheta^s} J(\btheta_{t})} \bigg) \bigg \rangle  \bigg] \nonumber \\
		& = 0 \nonumber
		\end{align}
		where $(i)$ follows Tower rule; $(ii)$ follows the policy gradient expression in \eqref{eq:pg:shared}.
		
	Therefore, we bound term $ I_{1,2} $ as below:
	\begin{align}
	     I_{1,2} \geq& -2N \cdot \alpha_t \cdot L_v \cdot C_{\psi} \cdot \epsilon_{app} + \alpha_t \cdot (1-\gamma) \cdot N \cdot \mathbb{E}\bigg[ \| \overline{\nabla_{\btheta^s} J(\btheta_{t})} \|^2 \bigg] \nonumber
	 \end{align}
	 
	 Moreover, the component $I_{2,2}$ in \eqref{bound:actor:I_22} is expressed as below:
	 \begin{align}
		I_{2,2} :=& \alpha_t (1-\gamma) \sum_{i = 1}^{N} \mathbb{E} \bigg[ \| \nabla_{\theta_i^p} J( \btheta_{t}) \|^2 \bigg] + \alpha_t \sum_{i = 1}^{N} \mathbb{E} \bigg[ \bigg \langle \nabla_{\theta_i^p} J( \btheta_{t}), \bigg( \widehat{\delta}_t^* - \delta_t  \bigg) \cdot \psi_{\theta^{p}_i}(\tilde{s}_t, \tilde{a}_{i,t} ; \theta_{i,t}) \bigg \rangle \bigg]  \nonumber \\
		& \quad + \alpha_t \sum_{i = 1}^{N} \mathbb{E} \bigg[ \bigg \langle \nabla_{\theta_i^p} J( \btheta_{t}), \delta_t \cdot \psi_{\theta^{p}_i}(\tilde{s}_t, \tilde{a}_{i,t} ; \theta_{i,t}) - (1-\gamma) \nabla_{\theta_i^p} J( \btheta_{t}) \bigg \rangle \bigg] \label{double:bound:I_22}
		\end{align}
		
	Similarly following the same steps in analyzing term $I_{1,2}$, the sampling error $\epsilon_{sp}$ could be avoided due to double samples in each iteration $t$. Hence, it holds that
	\begin{align}
	     I_{2,2} \geq& \alpha_t \cdot (1 - \gamma) \cdot \sum_{i = 1}^{N} \mathbb{E} \bigg[ \| \nabla_{\theta_i^p} J( \btheta_{t}) \|^2 \bigg] - 2 \alpha_t \cdot N \cdot L_v \cdot C_{\psi} \cdot \epsilon_{app}  \nonumber
	 \end{align} 
	
	Following remaining analysis steps in Proposition \ref{proposition:actor}, we obtain the convergence analysis for \aname ~with double sampling procedures in Algorithm \ref{alg: double sampling}. We are able to avoid the sampling error $\epsilon_{sp}$ at the cost of utilizing one more sample at each iteration. Therefore, we are able to present the results in Corollary \ref{cor:double_sample_cac}.

\end{document}